\RequirePackage[svgnames]{xcolor}

\documentclass[11pt,letterpaper]{mystyle}

\usepackage[all]{hypcap}
\usepackage[svgnames]{xcolor}
\usepackage[numbers]{natbib}
\usepackage{hyperref}[citecolor=cyan]

\hypersetup{
    colorlinks = true,
    citecolor = {cyan},
    linkcolor=magenta,
}
\tcbuselibrary{skins,breakable}

\newtcolorbox{carrierdefinition}{
enhanced,
breakable,
frame hidden,
colback=black!2,
borderline west={1.2pt}{0pt}{black!55},
boxsep=0pt,
left=8pt,
right=7pt,
top=6pt,
bottom=6pt,
before skip=7pt,
after skip=9pt
}

\newcommand{\greenmark}{\textcolor{ForestGreen}{\ding{51}}}
\newcommand{\redmark}{\textcolor{red}{\ding{55}}}
\newcommand{\lowmark}{\raisebox{-1.5pt}{\includegraphics[height=0.8em]{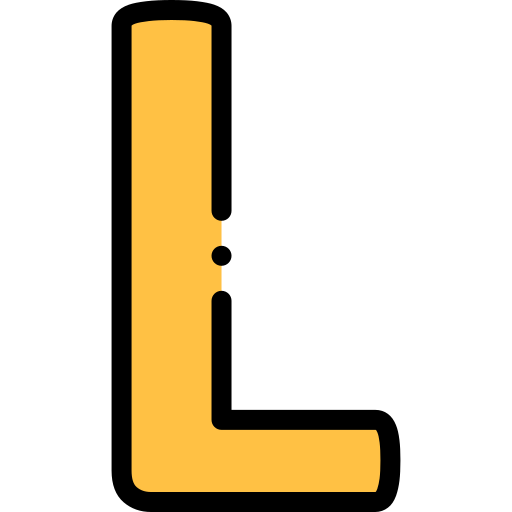}}}
\newcommand{\midmark}{\raisebox{-1.5pt}{\includegraphics[height=0.8em]{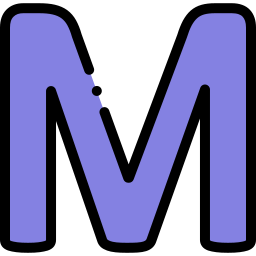}}}
\newcommand{\highmark}{\raisebox{-1.5pt}{\includegraphics[height=0.9em]{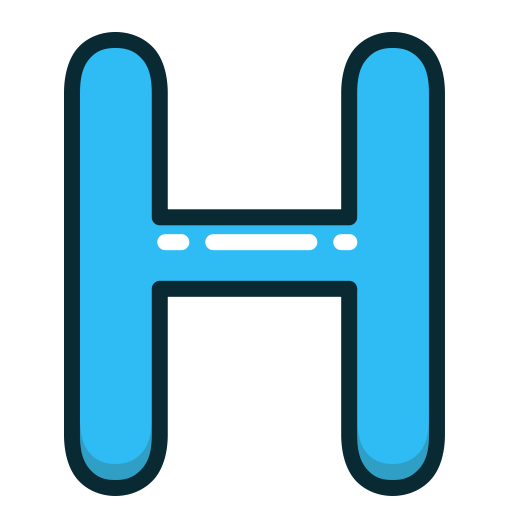}}}
\newcommand{\visualicon}{\raisebox{-1.5pt}{\includegraphics[height=0.9em]{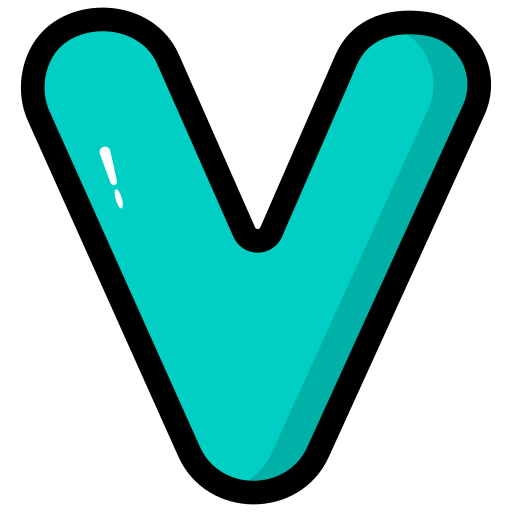}}}
\newcommand{\impliciticon}{\raisebox{-1.5pt}{\includegraphics[height=0.9em]{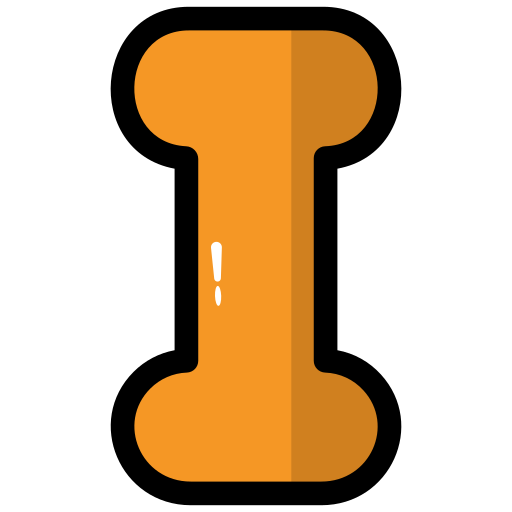}}}
\newcommand{\expliciticon}{\raisebox{-1.5pt}{\includegraphics[height=0.9em]{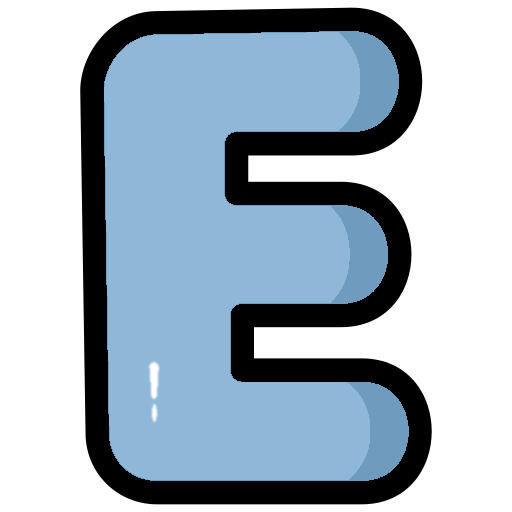}}}
\newcommand{\paraicon}{\raisebox{-1.5pt}{\includegraphics[height=0.9em]{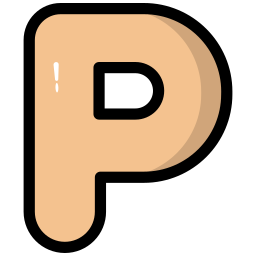}}}

\newcommand{\ruleicon}{\raisebox{-1.5pt}{\includegraphics[height=0.9em]{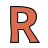}}}
\newcommand{\learnedicon}{\raisebox{-1.5pt}{\includegraphics[height=0.9em]{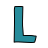}}}
\newcommand{\hybridicon}{\raisebox{-1.5pt}{\includegraphics[height=0.9em]{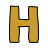}}}

\usepackage{algorithm}
\usepackage{algorithmicx}
\usepackage{algpseudocode}
\usepackage{microtype}
\usepackage{graphicx}
\expandafter\def\csname ver@subfig.sty\endcsname{}
\usepackage{booktabs} %
\usepackage{float}
\usepackage{bigstrut}

\usepackage{amsmath}
\usepackage{amssymb}
\usepackage{mathtools}
\usepackage{amsthm}
\usepackage{mathrsfs}
\usepackage{nicefrac}
\usepackage{dsfont}
\usepackage{enumitem}
\usepackage{subcaption}
\usepackage{graphicx,subfig}
\usepackage{cleveref}
\usepackage{bxcoloremoji}
\usepackage{placeins}
\usepackage{float}

\usepackage[utf8]{inputenc} %
\usepackage[T1]{fontenc}    %
\usepackage{hyperref}       %
\usepackage{url}            %
\usepackage{booktabs}       %
\usepackage{amsfonts}       %
\usepackage{nicefrac}       %
\usepackage{microtype}      %
\usepackage{graphicx}
\usepackage{subcaption} 
\usepackage{amssymb}
\usepackage{fdsymbol}
\usepackage{wrapfig}
\usepackage{lipsum}
\usepackage{enumitem}
\usepackage{stackengine}
\usepackage[font=small,labelfont=bf]{caption}
\usepackage{color}
\usepackage{adjustbox}

\usepackage{rotating}
\usepackage{makecell}

\usepackage[dvipsnames]{xcolor} 
\usepackage{boldline}

\newcommand{\schedbadge}[3]{%
  \tikz[baseline=-0.55ex]{%
    \node[
      rounded corners=0.32ex,
      fill=#1,
      text=#2,
      text width=1.10em,
      text height=1.00em,
      text depth=0.20em,
      inner xsep=0.18em,
      inner ysep=0.06em,
      align=center,
      font=\bfseries
    ] {#3};%
  }%
}
\newcommand{\schedcadence}[1]{\schedbadge{CadetBlue!75}{white}{#1}}
\newcommand{\schedcondition}[1]{\schedbadge{Goldenrod!55}{black!70}{#1}}

\newcommand{\schedpair}[2]{\schedcadence{#1}\kern0.12em\schedcondition{#2}}
\newcommand{\schedicon}[1]{%
  \raisebox{-0.18em}{\includegraphics[height=0.90em]{figures/t_ev_op/#1.pdf}}%
}
\newcommand{\cadenceframe}{\schedicon{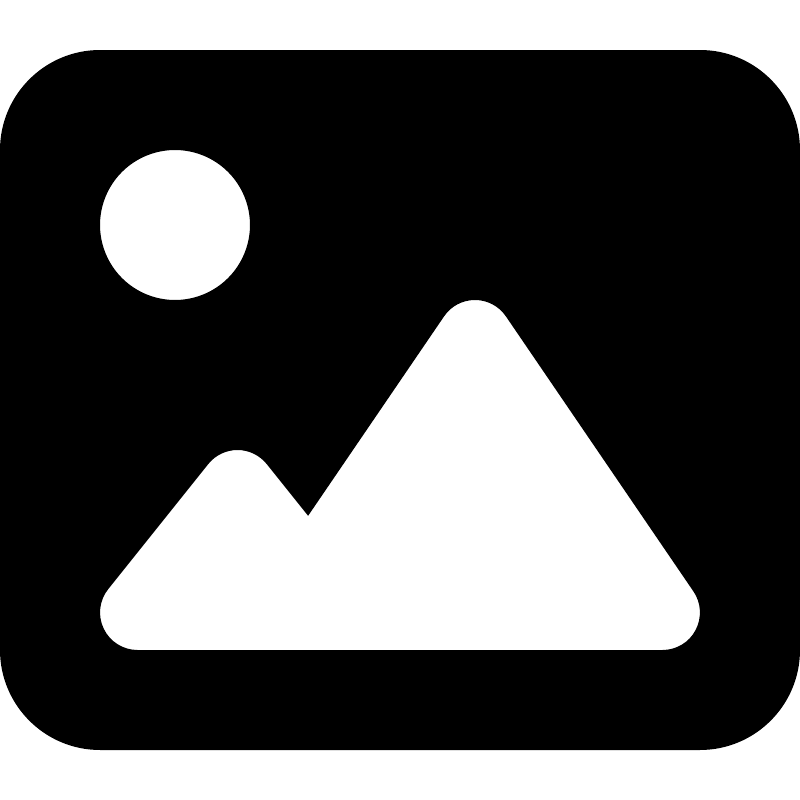}}
\newcommand{\cadenceblock}{\schedicon{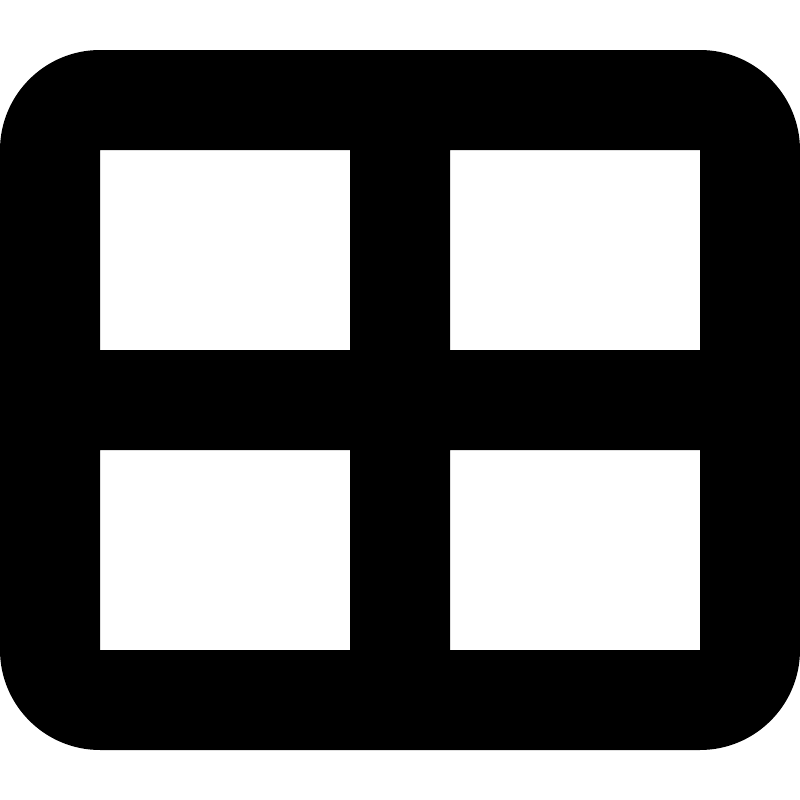}}
\newcommand{\cadencechunk}{\schedicon{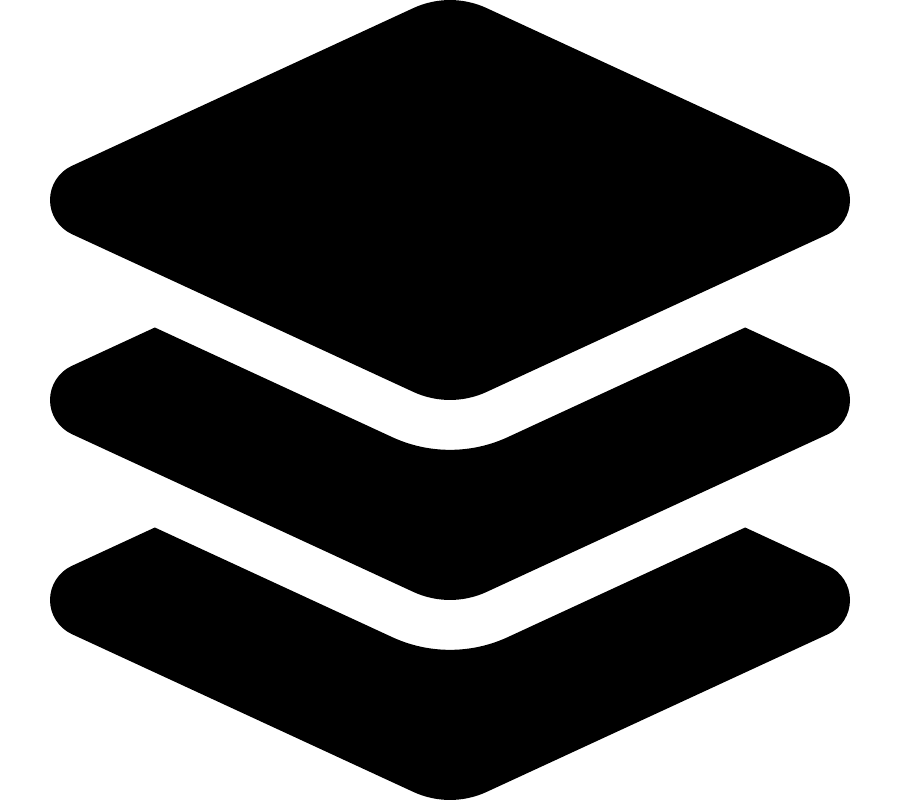}}
\newcommand{\cadenceshot}{\schedicon{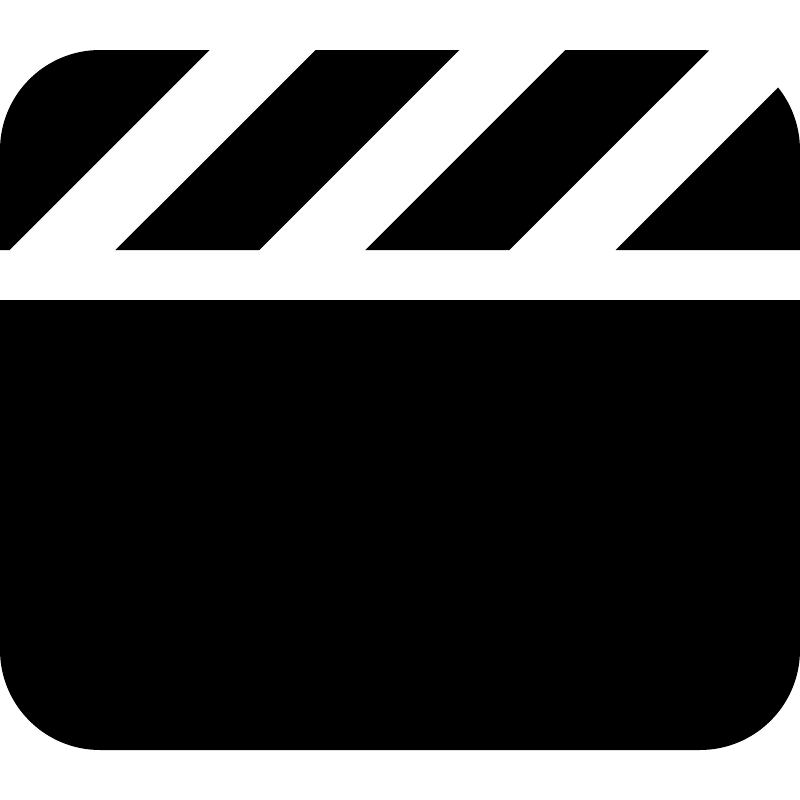}}

\newcommand{\conditionquery}{\schedicon{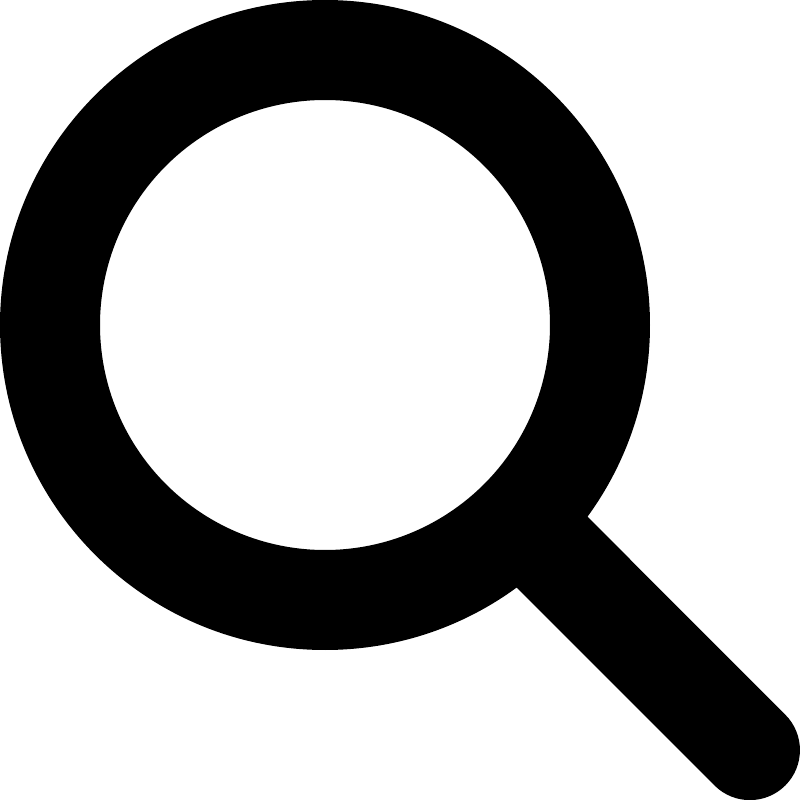}}
\newcommand{\conditionevent}{\schedicon{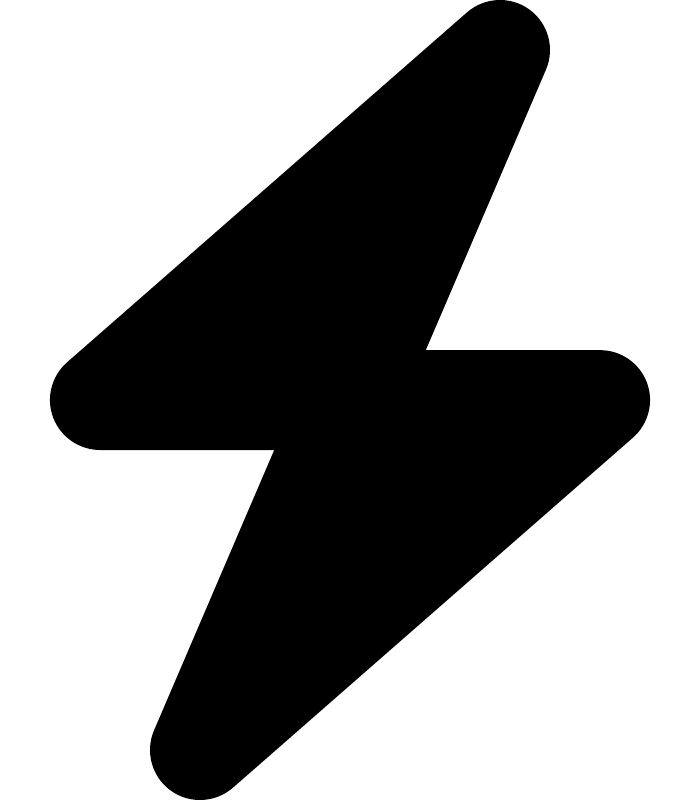}}
\newcommand{\conditionprompt}{\schedicon{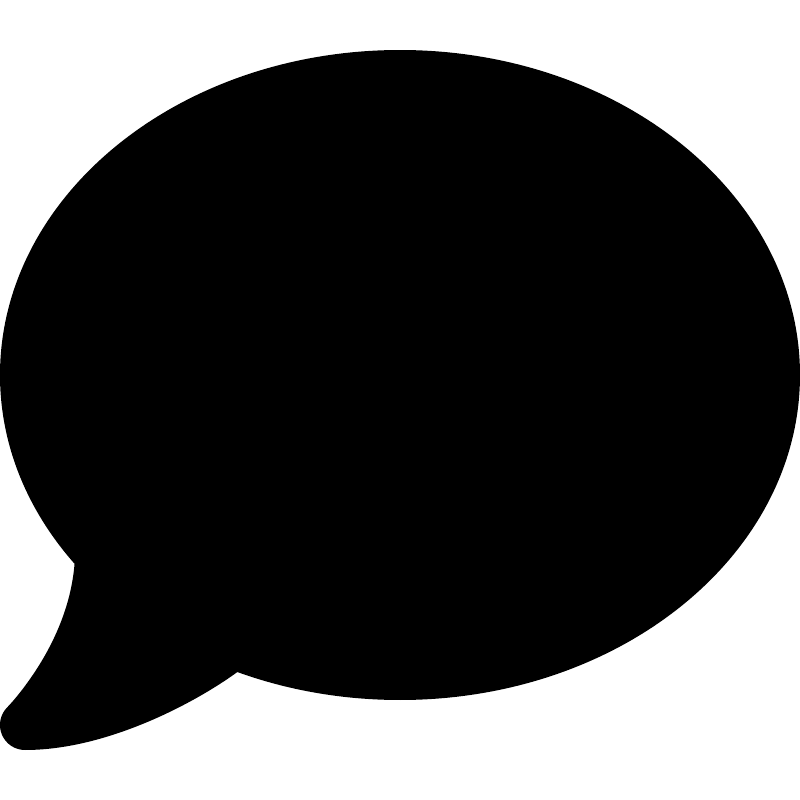}}
\newcommand{\conditionadaptive}{\schedicon{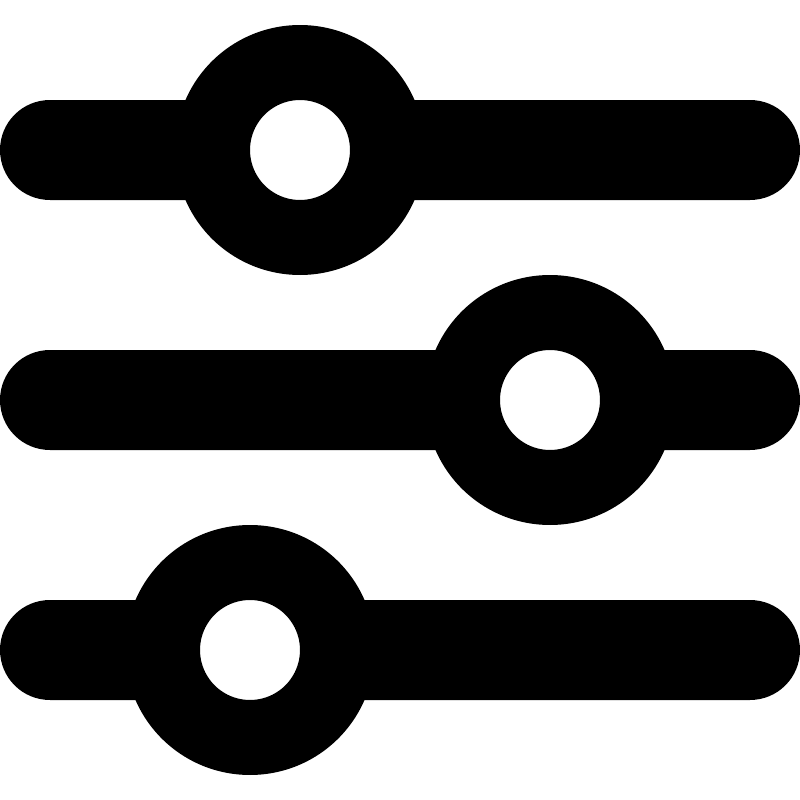}}
\newcommand{\scheduleicons}[2]{#1\kern0.18em#2}
\newcommand{\gapyes}{\schedicon{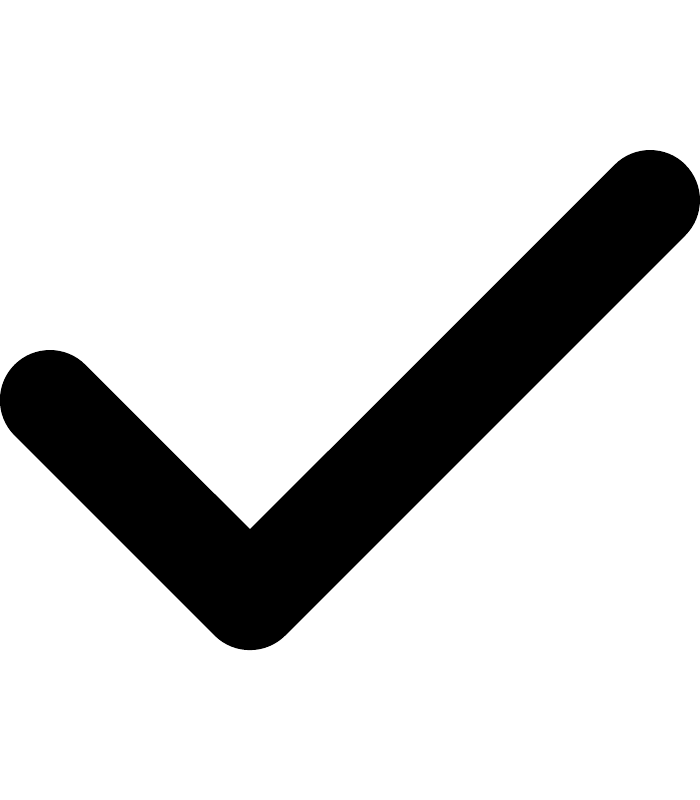}}
\newcommand{\gapno}{\schedicon{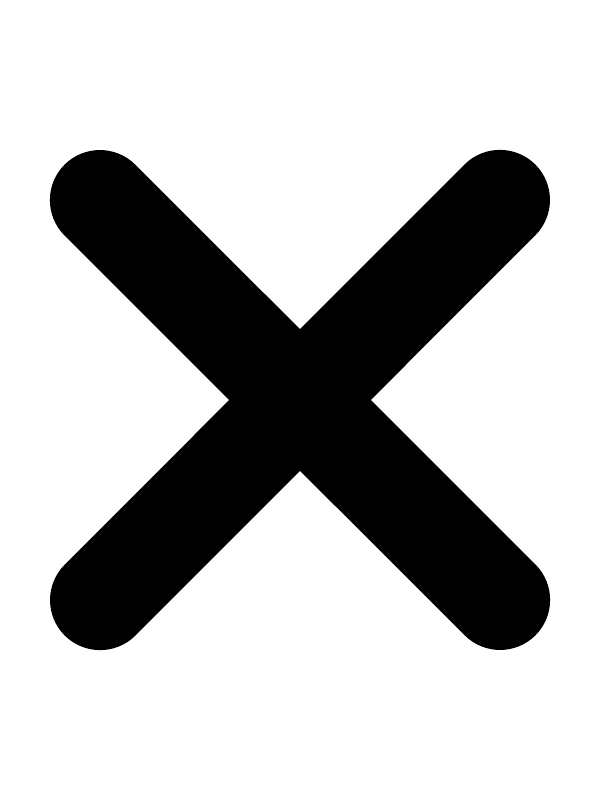}}
\newcommand{\gapsubset}{\schedicon{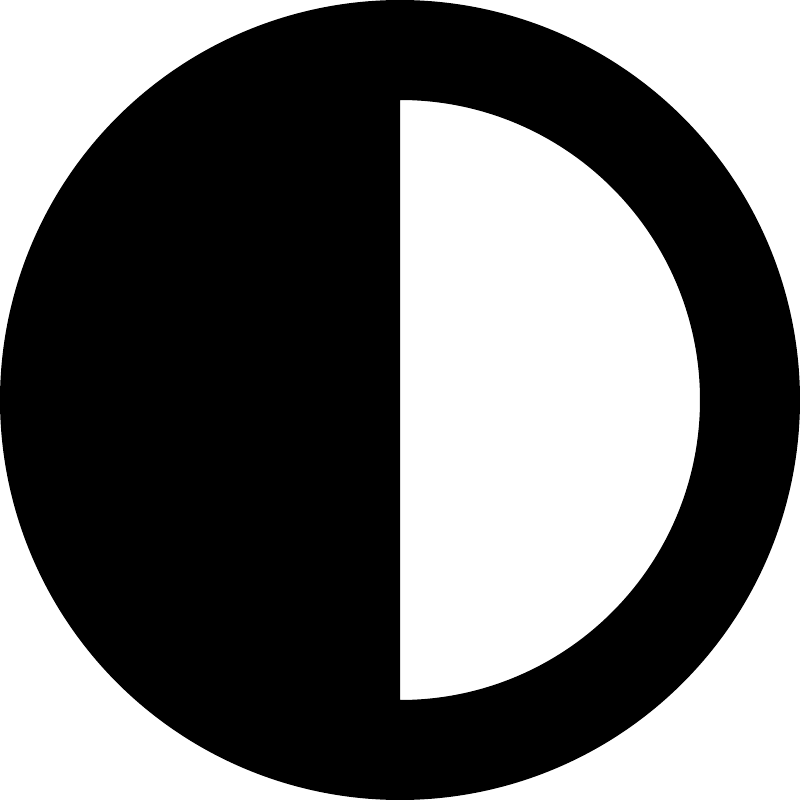}}

\newcommand{\opbox}[2]{%
  \begingroup
  \setlength{\fboxsep}{0.8pt}%
  \setlength{\fboxrule}{0.35pt}%
  \raisebox{0.06em}{%
    \fcolorbox{#1!38!black!10}{#1!7}{%
      \strut\scriptsize\ttfamily #2%
    }%
  }%
  \endgroup
}

\newcommand{\opwrite}[1]{\opbox{cyan}{#1}}
\newcommand{\opread}[1]{\opbox{green}{#1}}
\newcommand{\opupdate}[1]{\opbox{orange}{#1}}
\newcommand{\opmanage}[1]{\opbox{violet}{#1}}
\newcommand{\opinject}[1]{\opbox{teal}{#1}}

\newcommand{\x}{\mathbf{x}}

\definecolor{blanchedalmond}{rgb}{1.0, 0.92, 0.8}
\definecolor{carmine}{rgb}{0.59, 0.0, 0.09}
\definecolor{lightblue}{rgb}{0.22,0.45,0.70}%

\renewcommand{\mathbf}{\boldsymbol}

\makeatletter
\def\Ddots{\mathinner{\mkern1mu\raise\p@
\vbox{\kern7\p@\hbox{.}}\mkern2mu
\raise4\p@\hbox{.}\mkern2mu\raise7\p@\hbox{.}\mkern1mu}}
\makeatother

\definecolor{amaranth}{rgb}{0.9, 0.17, 0.31}
\definecolor{antiquebrass}{rgb}{0.8, 0.58, 0.46}
\definecolor{antiquefuchsia}{rgb}{0.57, 0.36, 0.51}
\definecolor{chromeyellow}{rgb}{0.31, 0.47, 0.26}

\newcommand{\github}{\raisebox{-1.5pt}{\includegraphics[height=1.05em]{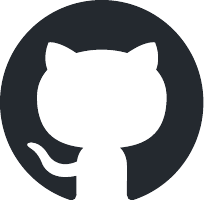}}}

\newtcolorbox{AIbox}[2][]{aibox,title=#2,#1}
\definecolor{lightblue}{rgb}{0.22,0.45,0.70}%
\definecolor{Gray}{gray}{0.95}
\definecolor{Cornsilk}{rgb}{1.0, 0.97, 0.86}

\newcommand{\obsbox}[1]{%
    \begin{tcolorbox}[
        colframe=black!70, 
        colback=blue!4, 
        boxrule=1pt, 
        arc=2mm, 
        top=3pt, bottom=3pt, left=3pt, right=3pt, boxsep=1pt,
        width=0.94\linewidth, 
        center               
    ]
        #1
    \end{tcolorbox}
}

\newtcolorbox{prompt}[3][]{
    colback=blue!4,
    colbacktitle=TinaCrimson!50,
    colframe=TinaCrimson!50,
    coltitle=white,
    fonttitle=\bfseries,
    boxsep=2pt,
    left=5pt,
    right=3pt,
    top=3pt,
    bottom=3pt,
    boxrule=1.4pt,
    enhanced,
    breakable,
    #1,
}

\newcommand{\probox}[1]{%
    \begin{tcolorbox}[
        colframe=TinaCrimson!60, 
        colback=blue!4,    
        toprule=0pt,       
        bottomrule=0pt,    
        rightrule=0pt,     
        leftrule=2.4pt,      
        arc=0pt,           
        outer arc=0pt,     
        boxsep=1pt,
        top=3pt,
        bottom=3pt,
        left=8pt,          
        right=3pt
    ]
        #1
    \end{tcolorbox}
}

\usepackage{amsmath}

\usepackage[all]{hypcap}

\title{~~The Past Frames the Future:\\{\fontsize{16}{12}\selectfont\textit{~~~~Memory for Autoregressive Video Generation --- A Survey}}}

\runningtitle{The Past Frames the Future: Memory for Autoregressive Video Generation}

\author{
~~~~~~Harold Haodong Chen$^{\ast}$, Rongjin Guo$^{\ast}$, Disen Lan$^{\ast}$, Wen-Jie Shu$^{\ast}$, Hongfei Zhang$^{\ast}$,\\
\vspace{-1.8mm}
\textbf{~~~~~~Hanzhe Hu, Shengtao Yao, Zixin Zhang, Guibin Zhang, Zhefan Rao, Jinxiu Liu, Yexin Liu,}\\
\vspace{-1.6mm}
\textbf{~~~~~~Rui Peng, Yuhao Liu, Bin Ren, Shuai Yang, Yukang Chen, Salman Khan, Ying-Cong Chen$^{\dagger}$,}\\
\vspace{-1.6mm}
\textbf{~~~~~~Ser-Nam Lim$^{\dagger}$, Rynson W.H. Lau$^{\dagger}$, Nicu Sebe, Yu Cheng$^{\dagger}$, Ming-Hsuan Yang, Qifeng Chen$^{\dagger}$}\\
~~~~~~HKUST, CityUHK, FDU, ZODA, CMU, NYU, HKUST(GZ), NUS, Georgia Tech,\\
\vspace{-1.6mm}
~~~~~~PKU, MBZUAI, NVIDIA, UCF, UNITN, NTU, UC Merced\\
{~~~~~~\small $^{\ast}$\textit{Core Contributors (Alphabetically).} $^{\dagger}$\textit{Core Supervisors.}}
}

\correspondingauthor{Note: Given the rapidly evolving field, we welcome suggestions on works that may have been overlooked; please contact us via email or GitHub.}

\begin{document}

\begin{abstract}

\vspace{-2mm}
\noindent
\makebox[\linewidth][c]{%
    \includegraphics[width=1\linewidth]{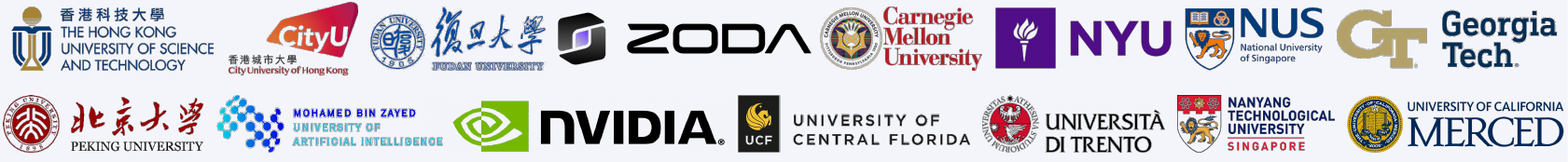}%
}

\vspace{3.4mm}
Advances in generative models have substantially improved the fidelity of video generation, propelling the field toward long-horizon generation, interactive world modeling, and evolving visual environments.
Autoregressive (AR) video generation offers a natural paradigm for these tasks by sequentially extending visual sequences through a causal step-wise rollout. However, a fundamental bottleneck emerges: as the generated sequence expands, practical models must operate under strictly bounded context windows, storage, and computational limits. Consequently, critical historical information, \textit{e.g.}, entity identities, spatial layouts, dynamic states, and intervention-induced causal changes, often leaves the active context long before its relevance diminishes. Overcoming this limitation and maintaining temporal persistence constitutes a fundamental memory problem.
This survey presents a systematic and comprehensive review of \textbf{\textit{memory mechanisms}} in AR video generation. We formulate memory operationally as persistent historical information maintained across outer AR steps, capable of influencing future generation even after the originating evidence is no longer locally accessible. Building upon this unified framework, we organize the literature through five complementary perspectives: \textbf{(I)~Forms}, the representational carriers of history; \textbf{(II)~Functions}, the specific semantic and physical information requiring preservation; \textbf{(III)~Operations}, the lifecycle of writing, reading, updating, managing, and integrating memory; \textbf{(IV)~Learning}, the optimization of memory behaviors under closed-loop rollouts; and \textbf{(V)~Evaluation}, the paradigms for diagnosing genuine memory capabilities.
Through these perspectives, we emphasize a core insight: effective memory transcends mere capacity. Retained states must remain accurate, accessible, and causally influential to subsequent generation. We conclude by synthesizing open challenges, including composable and resource-aware memory architectures, trustworthy state updating, self-rollout learning, and standardized evaluation. By bridging representations, mechanisms, and learning paradigms, this survey establishes a structured foundation for developing reliable, memory-conditioned video generation systems.

\vspace{2mm}


\vspace{5mm}

\coloremojicode{1F4C5} \textbf{Date}: September 2026

\coloremojicode{1F4E7} \textbf{Contact}: {\scriptsize\texttt{\{haroldchen328, guorong3529, disenlan1002, wenjieshu2003, fayehongfeizhang\}@gmail.com}}

\github{} \textbf{Repository}: \href{https://github.com/HaroldChen19/Awesome-AR-Video-Memory}{\textcolor{magenta}{\texttt{https://github.com/HaroldChen19/Awesome-AR-Video-Memory}}}

\end{abstract}

\maketitle
\clearpage
\begingroup
\hypersetup{linkcolor=TinaCrimson}
\linespread{0.88}\selectfont
\tableofcontents
\endgroup
\clearpage

\begin{figure*}[!b]
\centering
\vspace{-1.2em}
\includegraphics[width=1\linewidth]{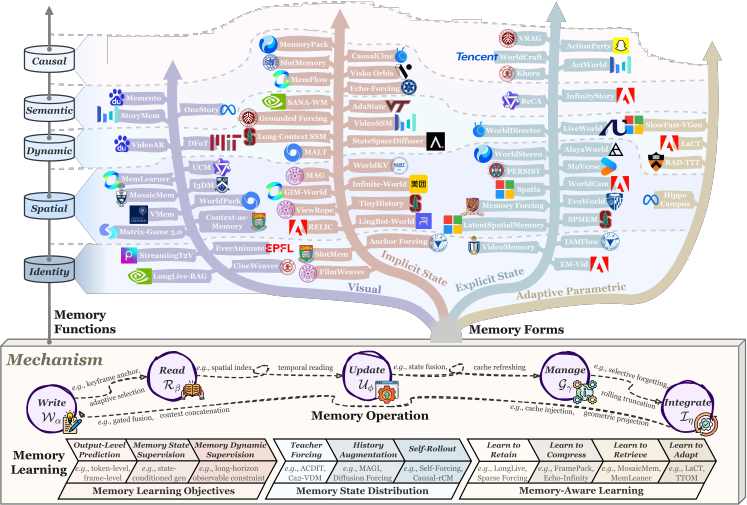}
\vspace{-1.8em}
\caption{The landscape of memory mechanisms in autoregressive video generation. The upper tree positions representative methods along two core dimensions: representational carriers (\textbf{Forms}~\S\ref{sec:representation}) and preservation goals (\textbf{Functions}~\S\ref{sec:functions}). The bottom panels outline the operational lifecycle (\textbf{Operations}~\S\ref{sec:operations}) and training paradigms (\textbf{Learning}~\S\ref{sec:learning}).}
\label{fig:figure_teaser}
\vspace{-1.2em}
\end{figure*}

\vspace{-4mm}
\section{Introduction}
\label{sec:intro}
\vspace{-1mm}

The rapid evolution of generative modeling, driven by scalable Transformer-based architectures, particularly diffusion transformers \citep{vaswani2017attention, peebles2023scalable}, has enabled substantial progress in high-fidelity image and short-video generation \citep{esser2024scaling, labs2025flux, cai2025z, wan2025wan, hacohen2026ltx}. The field is now advancing toward longer-horizon visual generation, including open-ended video generation \citep{yang2025longlive, huang2026self} and interactive video world modeling \citep{team2026advancing, mao2025yume}, where models must continuously extend visual sequences while preserving entities, layouts, motions, events, and long-range temporal commitments. A natural paradigm for this objective is \textit{autoregressive (AR) video generation}, which sequentially predicts future tokens, frames, or chunks conditioned on previously generated or observed visual units. Evolving from early discrete-token frameworks (\textit{e.g.}, VideoGPT \citep{yan2021videogpt}) and inference-time AR adaptations (\textit{e.g.}, FreeNoise \citep{qiu2024freenoise}), to recent AR diffusion frameworks (\textit{e.g.}, Causal Forcing \citep{zhu2026causal}), AR generation provides a scalable interface for streaming, continuation, and controllable long-form generation.

Despite this scalability, extending AR generation to long horizons raises a central challenge: maintaining \textit{temporal persistence under bounded access to history}. High-fidelity local generation alone is insufficient, since information required at a later step may no longer be available in the immediate context. Long-term consistency therefore depends on \textit{retention}: the ability to selectively preserve, update, retrieve, and integrate historical information according to its future relevance. Without effective retention, recursive generation may progressively forget or distort previously established entities, appearances, spatial layouts, motion states, or events \citep{huang2026self, zhu2026causal, yang2025longlive}. Existing strategies range from rolling visual windows and KV-cache retention \citep{kim2024fifo, yin2025slow, qiu2024freenoise} to more structured designs based on 3D priors \citep{wang2026latent}, compressed latent states \citep{zhang2025tinyhistory}, and retrieval banks \citep{hu2026longlive}. Although developed under different terminologies and architectural assumptions, these techniques address a shared functional problem: \textit{preserving and exposing useful historical information to subsequent generation}. In this survey, we study them through the unifying lens of \textbf{memory mechanisms}.

\probox{
\ding{228} \textbf{Memory Definition.}\quad
\textit{Memory is a persistent representation of past observations, generated states, or established world states and semantic commitments that is maintained across autoregressive steps and systematically conditions subsequent generation. Operationally, a state functions as memory when its removal or modification can affect later outputs after the corresponding evidence is no longer directly available in the immediate context.}
}

This definition does not draw a rigid boundary between memory and long context. Recent tokens, frames, or chunks that are densely consumed at every step constitute the active local context. As historical information is selectively retained, compressed, retrieved, consolidated, or revised beyond this local window, it moves from active context toward persistent memory. The distinction therefore depends on how historical information is maintained and used, rather than on a particular carrier, module name, or context-window size.

Despite growing attention, memory in AR video generation remains insufficiently systematized in two respects. \textbf{(I) Lack of a Paradigm-Specific Taxonomy:} Existing surveys on video diffusion \citep{xing2024survey, melnik2024video, li2024survey} and world models \citep{zhu2024sora, ding2025understanding, liu2026towards} cover broader generative paradigms but do not treat memory as a central design problem of causal AR rollouts, where memory is often discussed narrowly in terms of retrieval or long-context conditioning. \textbf{(II) Fragmented Conceptual Standards:} Recent studies use the term ``memory'' for mechanisms that differ substantially in their objectives, assumptions, representations, and deployment settings, while functionally related mechanisms may instead be described as context, cache, history, or state. This fragmentation obscures shared design principles and trade-offs, making systematic comparison difficult.

To bridge these gaps, this paper presents a comprehensive survey and unified framework for memory mechanisms in autoregressive video generation. We first establish the problem setting and formalize the role of memory under bounded access to history in Section~\S\ref{sec:background}. Building on this foundation, we organize the literature along five core questions:
\obsbox{
\begin{itemize}[leftmargin=1.6em]
    \item[\ding{182}] \textbf{Form:} \textit{What representations can memory take in AR video generation?}
    \vspace{0.6em}
    \item[\ding{183}] \textbf{Function:} \textit{What does memory need to preserve for future generation?}
    \vspace{0.6em}
    \item[\ding{184}] \textbf{Operations:} \textit{How is memory written, read, updated, managed, and integrated?}
    \vspace{0.6em}
    \item[\ding{185}] \textbf{Learning:} \textit{How are memory states and operations learned under AR rollout?}
    \vspace{0.6em}
    \item[\ding{186}] \textbf{Evaluation:} \textit{How can memory capability and its effect on generation be evaluated?}
\end{itemize}
}

To answer these questions, the remainder of this survey systematically deconstructs the memory pipeline. Section~\S\ref{sec:representation} addresses question~\ding{182} by categorizing the carriers of historical information into visual, implicit, explicit, and parametric forms. Section~\S\ref{sec:functions} addresses question~\ding{183} by organizing the historical properties that memory must preserve into identity, spatial, dynamic, semantic, and causal responsibilities. Section~\S\ref{sec:operations} then describes the operational lifecycle through which historical evidence is written, read, updated, managed, and integrated during sequential rollout (question~\ding{184}).

Moving from architecture to optimization, Section~\S\ref{sec:learning} for question~\ding{185} reviews how memory is learned through training objectives, memory-state distributions, and memory-aware optimization strategies. 
For question~\ding{186}, Section~\S\ref{sec:evaluation} examines existing evaluation practices, with particular attention to the distinction between memory-revealing protocols and general measures of video quality or temporal consistency.
Finally, Section~\S\ref{sec:future} synthesizes these structural and empirical insights to identify unresolved challenges and future directions for memory-conditioned AR video generation.
Together, these perspectives span three levels: system characterization (Forms, Functions, and Operations), system construction (Learning), and evidential validation (Evaluation).

\vspace{-1.2em}
\paragraph{Contribution.} In summary, the primary contributions of this survey are fourfold: \textbf{(\textit{i})} we provide a unified formulation of memory in autoregressive video generation, clarifying its role in sustaining temporal persistence across long-horizon rollouts; \textbf{(\textit{ii})} we develop a structured taxonomy that characterizes memory mechanisms by their representational carriers (Forms), generative responsibilities (Functions), and operational lifecycles (Operations); \textbf{(\textit{iii})} we systematically review how memory is learned and evaluated, covering training objectives, memory-state distributions, memory-aware optimization, and the distinction between general video consistency and memory-revealing evidence; and \textbf{(\textit{iv})} we synthesize existing progress and shared trade-offs to identify key challenges and future directions for memory-conditioned video generation.

\vspace{-1.2em}
\paragraph{Scope of This Survey.}
This survey focuses on memory mechanisms for \emph{autoregressive} video generation, where visual units, \textit{e.g.}, tokens, frames, or chunks, are causally generated from previously observed or generated history. We cover both discrete-token and continuous AR generators, including diffusion-, flow-, and Transformer-based formulations, when they explicitly preserve, manage, or transform historical state to condition subsequent visual generation. This includes long-horizon and interactive world generation under causal visual rollout, while excluding short-video generation without persistent cross-step memory. Closely related world-action models (WAMs) are covered selectively: recent systems increasingly adopt AR video-action modeling, yet their primary objective is typically action or policy prediction, with memory often realized implicitly through persistent KV states rather than treated as a dedicated design problem (\textit{e.g.}, LingBot-VA \citep{li2026causal}). We therefore discuss such models only when their memory mechanisms directly inform AR visual generation, rather than attempting comprehensive coverage of the WAM literature.

\section{Background: Why is Memory Needed?}
\label{sec:background}

Memory in autoregressive (AR) video generation is realized through diverse mechanisms, ranging from retained visual context and neural states to retrieval stores and structured world representations. To establish a unified foundation, we first formulate AR video generation as a causal rollout over generic visual units, including tokens, frames, latents, clips, and chunks, and distinguish the principal paradigms used to generate each unit. By separating the outer temporal rollout from the inner generation process, this formulation covers discrete predictors as well as diffusion- and flow-based generators under a common notation. We next examine bounded historical access in practical AR systems and clarify the relationship between active local context and persistent memory. Finally, we introduce a memory-conditioned formulation and use it to characterize the failure modes caused by limited access to history, providing the formal basis for the representational forms, functional responsibilities, operations, learning strategies, and evaluation practices discussed in subsequent sections.

\subsection{Autoregressive Video Generation}
\label{sec:2.1}

Let a raw video of \(T\) frames be denoted by \(\mathbf{x}_{1:T} = \{\mathbf{x}_1, \dots, \mathbf{x}_T\}\), where \(\mathbf{x}_t\) represents a spatial frame. To accommodate different generation granularities and representation spaces, we abstract the video into an ordered sequence of \(N\) autoregressive visual units, \(\mathbf{y}_{1:N} = \{\mathbf{y}_1, \dots, \mathbf{y}_N\}\). Depending on the model, each unit \(\mathbf{y}_n\) may correspond to a discrete token block \citep{yan2021videogpt, kondratyuk2023videopoet, wang2024emu3}, a continuous latent frame \citep{voleti2022mcvd, yang2023diffusion, valevski2025diffusion, alonso2024diffusion}, a short clip \citep{henschel2025streamingt2v, qiu2024freenoise}, or a multi-frame video chunk \citep{teng2025magi, ji2026videoar}.

AR video generation factorizes the conditional distribution of these units into causal prediction steps:
\begin{equation}
p_\theta(\mathbf{y}_{1:N}\mid \mathbf{c})
=
\prod_{n=1}^{N}
p_\theta(\mathbf{y}_n \mid \mathbf{y}_{<n}, \mathbf{c}),
\end{equation}
where \(\theta\) denotes the model parameters, \(\mathbf{y}_{<n} = \{\mathbf{y}_1, \dots, \mathbf{y}_{n-1}\}\) denotes the preceding visual units, and \(\mathbf{c}\) denotes external conditioning signals, such as text prompts \citep{kondratyuk2023videopoet, wan2025wan}, reference images \citep{ren2024consisti2v, liu2026infinitystar}, camera trajectories \citep{gao2026memcam, team2026advancing}, or control commands \citep{valevski2025diffusion, alonso2024diffusion}.

Throughout this survey, \textit{\textbf{autoregression} refers to the outer causal rollout over visual units rather than to a particular architecture used to generate each unit}. The step-wise conditional distribution \(p_\theta(\mathbf{y}_n \mid \mathbf{y}_{<n}, \mathbf{c})\) can be parameterized by different generative paradigms. Thus, early discrete token-based AR models and recent AR diffusion or flow-based systems share the same temporal conditioning logic, while differing in how the local unit \(\mathbf{y}_n\) is synthesized.

\begin{figure*}[!t]
\centering
\vspace{-0.4em}
\includegraphics[width=1\linewidth]{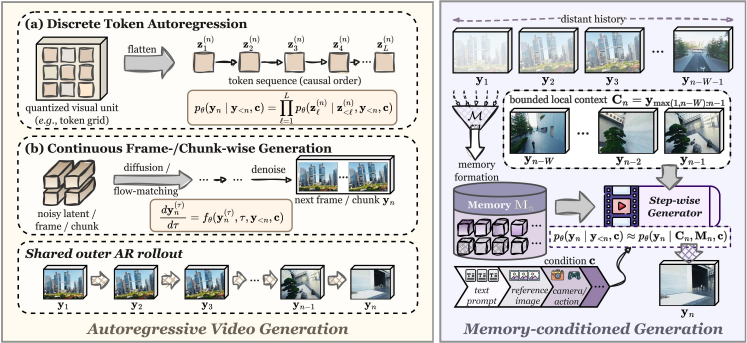}
\vspace{-1.6em}
\caption{Overview of autoregressive video generation and memory-conditioned generation. (\textbf{\textit{Left}}) AR video generation encompasses discrete token autoregression and continuous frame-/chunk-wise generation, which differ in the inner generation process but share the same causal outer rollout over visual units. (\textbf{\textit{Right}}) Under bounded local context, a persistent memory state retains information from distant history and conditions the step-wise generator together with the current context and external conditions, enabling an approximation to full-history generation.}
\label{fig:figure_background}
\vspace{-0.6em}
\end{figure*}

\subsection{Paradigms of Step-wise Generation}

While the temporal factorization establishes causality, the step-wise conditional model $p_\theta(\mathbf{y}_n \mid \mathbf{y}_{<n}, \mathbf{c})$ remains highly flexible. Based on the representation and synthesis process of $\mathbf{y}_n$, existing methods follow two primary paradigms (Figure~\ref{fig:figure_background} (\textit{Left})): \textit{discrete token autoregression} and \textit{continuous frame- or chunk-wise generation}.

\vspace{-1.2em}
\paragraph{Discrete Token Autoregression.}
Early frameworks (\textit{e.g.}, VideoGPT \citep{yan2021videogpt}, VideoPoet \citep{kondratyuk2023videopoet}, Emu3 \citep{wang2024emu3}) map visual inputs into discrete token vocabularies via learned quantizers, such as VQ-VAE \citep{van2017neural}. In this paradigm, each visual unit \(\mathbf{y}_n\) is flattened into a sequence of \(L\) tokens, denoted as \(\mathbf{z}^{(n)}_{1:L} = \{\mathbf{z}^{(n)}_1, \dots, \mathbf{z}^{(n)}_L\}\). The inner generation step further decomposes into next-token prediction:
\begin{equation}
p_\theta(\mathbf{y}_n \mid \mathbf{y}_{<n}, \mathbf{c}) = \prod_{\ell=1}^{L} p_\theta(\mathbf{z}^{(n)}_{\ell} \mid \mathbf{z}^{(n)}_{<\ell}, \mathbf{y}_{<n}, \mathbf{c}).
\end{equation}
Transformer architectures \citep{vaswani2017attention, peebles2023scalable} naturally parameterize this distribution with causal masking and key-value (KV) caches. However, high-fidelity video often requires long token sequences, increasing inference latency and exposing generation to error accumulation both within and across visual units \citep{xing2024survey}.

\vspace{-1.2em}
\paragraph{Continuous Frame- or Chunk-wise Generation.}
Recent architectures often operate in continuous pixel or latent spaces \citep{voleti2022mcvd, yang2023diffusion}, where \(\mathbf{y}_n\) represents a frame \citep{alonso2024diffusion, valevski2025diffusion} or a multi-frame chunk \citep{ji2026videoar, teng2025magi}. Instead of generating tokens one by one, the inner generator generates the entire unit through a conditional generation process. For diffusion- \citep{song2020score} and flow-based \citep{lipman2022flow} models, this process can be viewed as an inner denoising or flow trajectory conditioned on the available history:
\begin{equation}
\frac{d\mathbf{y}_n^{(\tau)}}{d\tau} = f_\theta(\mathbf{y}_n^{(\tau)}, \tau, \mathbf{y}_{<n}, \mathbf{c}),
\end{equation}
where \(\tau\) denotes the internal diffusion or flow time and is distinct from the external AR step \(n\). The state \(\mathbf{y}_n^{(1)}\) may be initialized from noise, while \(f_\theta\) denotes a learned vector field instantiated according to the underlying score- or velocity-based formulation.

Continuous chunk-wise generation leverages the perceptual strength of modern diffusion and flow models and can shorten the outer AR horizon by generating multiple frames per step. Nevertheless, each generated unit still becomes part of the conditioning history for subsequent steps. The two paradigms therefore differ in their inner synthesis processes but face the same long-horizon requirement: preserving useful historical information as direct access to the expanding history becomes bounded.

\subsection{Bounded Historical Access}

The AR factorization provides a scalable way to extend generation in time, but it does not by itself guarantee long-term consistency. In principle, the conditioning history \(\mathbf{y}_{<n}\) grows monotonically with the rollout. In practice, standard self-attention incurs quadratic sequence-length complexity. Sparse-attention architectures \citep{child2019generating} and IO-aware kernels such as FlashAttention \citep{dao2022flashattention} substantially improve practical efficiency, but continually growing visual histories remain constrained by accelerator memory, KV-cache storage, and latency.
Practical step-wise generators therefore typically operate on a bounded local context:
\begin{equation}
\mathbf{C}_n = \mathbf{y}_{\max(1,n-W):n-1},
\end{equation}
where \(W\) denotes the maximum number of recent visual units retained in the active context. The full-history conditional distribution \(p_\theta(\mathbf{y}_n \mid \mathbf{y}_{<n}, \mathbf{c})\) is consequently approximated by \(p_\theta(\mathbf{y}_n \mid \mathbf{C}_n, \mathbf{c})\).

\vspace{-1.2em}
\paragraph{Open-ended Long Video Generation.}
For open-ended long video generation, bounded context controls computation and supports local continuity, but removes direct access to evidence outside the active window \citep{dai2019transformer}. Information about earlier entities, scene layouts, motion states, or visual anchors must then be propagated through recent generations, preserved in a separate historical state, or become unavailable to the generator. This loss of access compounds other sources of long-horizon degradation, including autoregressive error accumulation, stochastic variation across generated chunks, and the mismatch between teacher-forced training and self-rolled-out inference. Reconciling expanding temporal dependencies with bounded computation is therefore a central requirement for long-horizon generation.

\vspace{-1.2em}
\paragraph{Action-conditioned and Interactive Generation.}
The same constraint becomes more pronounced in action-conditioned or interactive video generation, \textit{e.g.}, Genie \citep{bruce2024genie}, where future observations depend not only on previous visual units but also on past interventions. Let \(\mathbf{a}_n\) denote the control action applied before step \(n\). The causal rollout can be written as:
\begin{equation}
p_\theta(\mathbf{y}_{1:N} \mid \mathbf{c}, \mathbf{a}_{1:N})
=
\prod_{n=1}^{N}
p_\theta(\mathbf{y}_n
\mid
\mathbf{y}_{<n}, \mathbf{a}_{\leq n}, \mathbf{c}).
\end{equation}
We distinguish \textit{observation-changing controls}, such as camera or agent-navigation commands in an otherwise static environment, from \textit{state-changing interventions} that modify entities, relations, or the environment itself. Both introduce long-range dependencies, but the former primarily stresses spatial preservation, whereas causal preservation concerns the persistent consequences of the latter. For state-changing interventions, world evolution may depend on actions that occurred far outside the active window. If only recent observations and actions remain accessible, an earlier intervention, such as moving an object or opening a door, may no longer constrain the output when its consequence must later be reflected \citep{wang2026matrix, nam2026worldcam, mao2025yume}. This is a causal instance of bounded historical access: the generator must preserve action-induced state changes after the originating evidence has left the local context.

\vspace{-1.2em}
\paragraph{From Active Context to Persistent Memory.}
Long context and memory are not separated by a fixed window length. Recent tokens, frames, or KV states that are supplied directly to every generation step constitute the active context; enlarging this context increases the amount of history immediately available to the generator. Historical conditioning serves a memory role when information is persistently maintained and its access is explicitly managed through operations such as selection, compression, indexing, retrieval, consolidation, or revision. In this survey, the distinction between $\mathbf{C}_n$ and the memory state introduced next is therefore operational: $\mathbf{C}_n$ denotes the recent history directly consumed by the step-wise generator, whereas memory denotes managed historical state maintained across AR steps. Practical systems may combine both forms of conditioning.

Both open-ended and interactive generation settings thus expose the same structural tension: historical dependencies can grow indefinitely, whereas generation operates under finite capacity. This motivates memory-conditioned AR generation \citep{zhu2025memorize, king2026echo, wu2026infinite}, in which a persistent and tractable historical state complements the bounded active context.

\subsection{Memory-Conditioned Generation}
\label{sec:2.4}

Following the operational distinction above, we introduce a unified formulation that abstracts heterogeneous memory mechanisms under a common memory-conditioned AR framework. Specifically, we augment the bounded active context with a persistent memory state that carries managed historical information across AR steps, as illustrated in Figure~\ref{fig:figure_background} (\textit{Right}). This abstraction is conceptually related to classical differentiable memory architectures, which couple persistent storage with learned mechanisms for reading, writing, and state management \citep{graves2014neural,weston2014memory,azarafrooz2022differentiable}.

\vspace{-1.2em}
\paragraph{Memory Formulation.}
Let \(\mathbf{C}_n = \mathbf{y}_{\max(1,n-W):n-1}\) denote the bounded local context available at step \(n\), and let \(\mathbf{M}_n \in \mathcal{M}\) denote the persistent memory state available before generating $\mathbf{y}_n$. The finite-context approximation is then generalized as:
\begin{equation}
p_\theta(\mathbf{y}_{1:N} \mid \mathbf{c})
=
\prod_{n=1}^{N}
p_\theta(\mathbf{y}_n \mid \mathbf{C}_n, \mathbf{M}_n, \mathbf{c}).
\end{equation}
For action-conditioned generation, the current action can be included explicitly:
\begin{equation}
p_\theta(\mathbf{y}_{1:N} \mid \mathbf{c}, \mathbf{a}_{1:N})
=
\prod_{n=1}^{N}
p_\theta(\mathbf{y}_n
\mid
\mathbf{C}_n, \mathbf{M}_n, \mathbf{a}_n, \mathbf{c}).
\end{equation}
Here, $\mathbf{a}_n$ specifies the current intervention, while the effects of earlier actions are carried through $\mathbf{C}_n$ and $\mathbf{M}_n$. Unless otherwise specified, we absorb actions and other external controls into the generalized condition \(\mathbf{c}\) for notational simplicity.

\vspace{-1.2em}
\paragraph{Memory State.}
Conceptually, \(\mathbf{M}_n\) provides a tractable representation of historical information that is not guaranteed to remain directly accessible through \(\mathbf{C}_n\). In a simple non-interactive setting, memory may aggregate the truncated prefix:
\begin{equation}
\mathbf{M}_n = \mathcal{A}(\mathbf{y}_{<n-W}, \mathbf{c}),
\end{equation}
where \(\mathcal{A}\) denotes an aggregation or compression operator. More generally, memory need not be a static summary of distant history. It may take the form of an attention cache, an indexed retrieval store, a recurrent hidden state, a structured scene representation, or an adaptive parameter state. The resulting model seeks to approximate full-history conditioning:
\begin{equation}
p_\theta(\mathbf{y}_n \mid \mathbf{y}_{<n}, \mathbf{c}) \approx p_\theta(\mathbf{y}_n \mid \mathbf{C}_n, \mathbf{M}_n, \mathbf{c}).
\end{equation}
Conceptually, $\mathbf{M}_n$ may encode historical evidence, a revisable estimate of the current world state, or both. The former preserves information from past observations or generated outputs, whereas the latter represents the state currently treated as valid and may change as new evidence arrives.

\vspace{-1.2em}
\paragraph{Memory Lifecycle.}
At each AR step, memory is read and integrated before generation, then written, updated, and managed after generation. The model first forms a query from the current context:
\begin{equation}
\mathbf{q}_n = Q_\theta(\mathbf{C}_n, \mathbf{c}),
\end{equation}
and retrieves relevant historical information:
\begin{equation}
\mathbf{r}_n = \mathcal{R}_\beta(\mathbf{q}_n, \mathbf{M}_n).
\end{equation}
The retrieved memory is then integrated with the local context and external condition:
\begin{equation}
\mathbf{h}_n = \mathcal{I}_\eta(\mathbf{C}_n, \mathbf{r}_n, \mathbf{c}),
\end{equation}
after which generation proceeds according to \(p_\theta(\mathbf{y}_n \mid \mathbf{h}_n)\). Once \(\mathbf{y}_n\) has been generated, or observed during training, a write candidate is extracted:
\begin{equation}
\mathbf{w}_n = \mathcal{W}_\alpha(\mathbf{y}_n, \mathbf{C}_n, \mathbf{c}).
\end{equation}
The persistent state is subsequently updated:
\begin{equation}
\widetilde{\mathbf{M}}_{n+1}
=
\mathcal{U}_\phi(\mathbf{M}_n, \mathbf{w}_n),
\end{equation}
and a management operator enforces the memory budget $B$ through retention, compression, or revision:
\begin{equation}
\mathbf{M}_{n+1}
=
\mathcal{G}_\gamma(\widetilde{\mathbf{M}}_{n+1}; B).
\end{equation}
This lifecycle does not assume a particular memory form. Depending on the specific implementation, its operations may be instantiated in different ways, including appending latent frames to a FIFO queue \citep{kim2024fifo}, refreshing KV states \citep{chen2026past, yang2025longlive}, updating recurrent hidden states \citep{yu2025videossm}, maintaining retrieval stores \citep{yu2025context, hu2026longlive}, or revising structured scene states \citep{xiao2025worldmem, li2025vmem}.

\vspace{-1.2em}
\paragraph{Memory Scope.}
The scope of memory is determined by persistent temporal transfer rather than by module naming or carrier type. A state functions as memory when it is maintained across AR steps to preserve historical information that is not guaranteed to remain available in the active context and systematically influences subsequent generation. Static conditions, such as an initial text prompt, are not memory by themselves unless the system derives and maintains evolving states, interactions, or action-induced changes from them during rollout.

\begingroup
\setlength{\tabcolsep}{28pt}
\begin{table*}[!t]
\renewcommand{\arraystretch}{1.14}
\caption{Summary of core notations used in this survey.}
\vspace{-0.6em}
\label{tab:notations}
\centering
\footnotesize
\begin{tabular}{c p{0.69\linewidth}}
\hlineB{2.5}
\rowcolor{CadetBlue!20}
\textbf{Notation} & \textbf{Description} \\
\hlineB{1.5}

\multicolumn{2}{c}{
\textit{Sequence \& Context (\S\ref{sec:background})}
} \\
\hline
\rowcolor{gray!10}
$\mathbf{y}_{1:N}$
& Sequence of $N$ autoregressive visual units \\

$\hat{\mathbf{y}}_n$
& Model-generated autoregressive visual unit at step $n$ \\

\rowcolor{gray!10}
$\mathbf{H}_n=\mathbf{y}_{<n}$
& Complete history available before generation step $n$ \\

$\mathbf{z}_{1:L}^{(n)}$
& Sequence of $L$ discrete tokens flattened from visual unit $\mathbf{y}_n$ \\

\rowcolor{gray!10}
$\mathbf{z}_n$
& Continuous latent representation of $\mathbf{y}_n$, when applicable \\

$\mathbf{C}_n$
& Bounded local context available at step $n$, typically
$\mathbf{y}_{\max(1,n-W):n-1}$ \\

\rowcolor{gray!10}
$W$
& Size of the bounded local context window \\

$\mathbf{c}$
& Generalized external condition (\textit{e.g.}, text prompts or references);
actions are explicit when specified \\

\rowcolor{gray!10}
$\mathbf{a}_n$
& Action or control signal applied before generation step $n$ \\

\hline
\multicolumn{2}{c}{
\textit{Memory State Variables \& Carriers
(\S\ref{sec:representation} \& \S\ref{sec:operations})}
} \\
\hline
\rowcolor{gray!10}
$\mathbf{M}_n$
& Persistent memory state available before generating $\mathbf{y}_n$ \\

$\widetilde{\mathbf{M}}_{n+1}$
& Post-update intermediate memory state before management \\

\rowcolor{gray!10}
$\mathbf{M}_n^{\mathrm{vis}}$
& Visual memory carrier preserving historical observations \\

$\mathbf{M}_n^{\mathrm{imp}}$
& Implicit memory carrier retaining history in model-native neural states \\

\rowcolor{gray!10}
$\mathbf{M}_n^{\mathrm{exp}}$
& Explicit memory carrier representing structured world-state variables \\

$\mathbf{M}_n^{\mathrm{par}}$
& Parametric memory carrier retaining history in designated adaptable parameters \\

\rowcolor{gray!10}
$\Delta\theta_u$
& Adaptable parameter-memory state after the $u$-th adaptation event \\

$\mathbf{w}_n$
& Write candidate extracted after $\mathbf{y}_n$ is generated or observed \\

\rowcolor{gray!10}
$\mathbf{q}_n$
& Query state formulated for memory retrieval \\

$\mathbf{r}_n$
& Retrieved memory context exposed by the reading operation \\

\rowcolor{gray!10}
$\mathbf{h}_n$
& Memory-conditioned generative context used by the step-wise generator \\

$B$
& Capacity budget for memory storage, computation, or retrieval \\

\hline
\multicolumn{2}{c}{
\textit{Core Operational Operators (\S\ref{sec:operations})}
} \\
\hline
\rowcolor{gray!10}
$\mathcal{Q}_\theta$
& Query-formation operator producing $\mathbf{q}_n$ from current context and conditions \\

$\mathcal{A}$
& Optional aggregation or compression operator summarizing historical evidence \\

\rowcolor{gray!10}
$\mathcal{W}_\alpha$
& Writing operator extracting evidence into memory candidates \\

$\mathcal{R}_\beta$
& Reading operator retrieving or routing memory according to $\mathbf{q}_n$ \\

\rowcolor{gray!10}
$\mathcal{U}_\phi$
& Update operator governing the evolution of the memory state \\

$\mathcal{G}_\gamma$
& Management operator enforcing capacity, compression, or revision \\

\rowcolor{gray!10}
$\mathcal{I}_\eta$
& Integration operator fusing retrieved memory into the generation process \\

$\mathcal{F}$
& Effective memory transition incorporating writing, updating, and optional management \\

\hline
\multicolumn{2}{c}{
\textit{Learning Objectives \& Distributions (\S\ref{sec:learning})}
} \\
\hline
\rowcolor{gray!10}
$\mathcal{L}_{\mathrm{pred}}$
& Output-level prediction objective
(\textit{e.g.}, conditional denoising or flow matching) \\

$\mathcal{L}_{\mathrm{mem}}$
& Memory-state objective supervising or regularizing memory content \\

\rowcolor{gray!10}
$\mathcal{L}_{\mathrm{dyn}}$
& Memory-dynamics objective constraining multi-step state evolution \\

$\mathcal{L}_{\mathrm{aware}}$
& Training objective under deployment-time memory constraints \\

\rowcolor{gray!10}
$q_{\mathrm{TF},n}$
& Memory-state distribution induced by ground-truth histories at step $n$ \\

$q_{\theta,n}$
& Memory-state distribution induced by self-generated AR rollouts at step $n$ \\

\rowcolor{gray!10}
$q_{\mathrm{aug},n}$
& Augmented memory-state distribution induced by perturbed histories at step $n$ \\

\hlineB{2.5}
\end{tabular}

\vspace{0.25em}
\begin{minipage}{0.98\linewidth}
\scriptsize
\textit{Note:}
The step index of $q_{\mathrm{TF},n}$, $q_{\theta,n}$, and
$q_{\mathrm{aug},n}$ is omitted when clear from context.
\end{minipage}
\end{table*}
\endgroup

\subsection{Failure Modes of Memoryless Rollouts}
\label{sec:2.5}

The memory-conditioned formulation treats \(\mathbf{M}_n\) as a managed state for historical information that is not guaranteed to remain accessible through $\mathbf{C}_n$. We use \textit{memoryless rollout} to refer operationally to generation in which such persistent state is absent or severely constrained, leaving the model to rely primarily on the finite-context distribution \(p_\theta(\mathbf{y}_n \mid \mathbf{C}_n, \mathbf{c})\). This does not imply an absence of all history, since $\mathbf{C}_n$ still provides recent visual units. Such bounded conditioning can preserve local continuity but offers limited support for long-range visual, semantic, and causal coherence, leading to several recurring failure modes.

\vspace{-1.2em}
\paragraph{Entity Forgetting.}
Once an object or character leaves the active context, the evidence needed to identify it may no longer condition generation. Upon reappearance, the entity may be omitted, duplicated, assigned different attributes, or replaced by a visually similar but distinct instance \citep{he2026entitybench, zhang2026mbench}. In human-centric generation, this often appears as drift in facial identity, body shape, or other instance-defining details.

\vspace{-1.2em}
\paragraph{Appearance Drift.}
Even when an entity remains visible, its appearance may progressively deviate through unintended changes in texture, color, lighting, or style \citep{team2025inferix, huang2024vbench}. Since each step conditions primarily on recent generated outputs, small visual errors can be repeatedly propagated and amplified in the absence of a stable long-range reference.

\vspace{-1.2em}
\paragraph{Spatial Inconsistency.}
Without persistent spatial state, the generator has no direct record of global scene structure after the supporting observations leave the active context. When a previously observed region re-enters the field of view, the model may produce incompatible layouts, background geometry, or object placements \citep{xu2026worldroambench, ye2026mind}. These failures reflect weak preservation of spatial topology across temporally separated observations.

\vspace{-1.2em}
\paragraph{Dynamics Degradation.}
Limited historical access also weakens the preservation of evolving motion states \citep{bansal2025videophy, jain2026mechverse}. Although individual transitions may remain locally plausible, extended rollouts can exhibit unnatural kinematics, motion freezing, phase inconsistency, or physical contact violations. These failures become more pronounced when early motion errors are repeatedly fed back through the active context and alter subsequent dynamics.

\vspace{-1.2em}
\paragraph{Semantic Drift.}
Long-form generation must preserve semantic commitments established during the rollout, including events, roles, relations, and unresolved goals. Even when the initial prompt remains available, these commitments may leave the active context, causing memoryless rollouts to contradict or repeat prior events, lose narrative progression, or fail to complete extended actions \citep{tang2025seqbench, wang2025your}. The video may therefore remain locally plausible while diverging from its global semantic intent.

\vspace{-1.2em}
\paragraph{Causal or State Inconsistency.}
For state-changing interventions, the resulting world state should persist after the action leaves the active context. Otherwise, an opened door may revert to closed, a displaced object may return to its original position, or an altered environment may reset \citep{duan2026liveworld, ma2026out}. Such failures occur when interventions are treated as transient controls rather than as updates to a persistent state whose consequences constrain later generation.

These symptoms are not uniquely caused by missing memory: similar failures may
also arise from imperfect dynamics modeling, control execution, or rendering.
Here they identify behaviors for which loss of relevant distant history can be
a contributing factor, with attribution discussed explicitly in
Section~\S\ref{sec:evaluation}.
Collectively, these failure modes show that the central challenge in long-horizon AR generation is not temporal length alone, but temporal persistence under bounded historical access. Addressing this challenge requires mechanisms that retain entity identity, anchor appearance, preserve spatial topology, track evolving dynamics, maintain semantic commitments, and record causal state transitions. The following sections examine these mechanisms through their representational forms, functional responsibilities, operational policies, learning paradigms, and evaluation practices.

\section{Forms: What Carries Memory?}
\label{sec:representation}

Before examining what memory should preserve to address the failure modes identified in Section~\S\ref{sec:2.5}, we first ask a structural question: \textit{what computational objects carry historical information across AR steps?} This motivates a carrier-centric taxonomy of memory mechanisms.

We use \textit{memory carrier} to denote the representational object in which historical information is retained and made available to subsequent generation. As the concrete basis on which memory functions and operations are implemented, the carrier provides a natural basis for organizing existing approaches. Accordingly, we distinguish four major types of memory carriers:
\obsbox{
\begin{itemize}[leftmargin=1.2em]
\item \textbf{Visual Memory} (\S\ref{sec:VM}), which preserves past frames, clips, or observation-aligned visual representations as reusable historical evidence.
\item \textbf{Implicit State Memory} (\S\ref{sec:ISM}), which retains history in model-native latent states without predefined semantic or geometric structure.
\item \textbf{Explicit State Memory} (\S\ref{sec:ESM}), which represents history through structured variables describing entities, scene layouts, dynamics, relations, events, or other interpretable world states.
\item \textbf{Adaptive Parametric Memory} (\S\ref{sec:PM}), which encodes sequence-, instance-, or experience-specific information in designated parameters or adaptable weights updated from prior evidence.
\end{itemize}
}

\begin{figure*}[!t]
\centering
\vspace{-0.4em}
\includegraphics[width=0.8\linewidth]{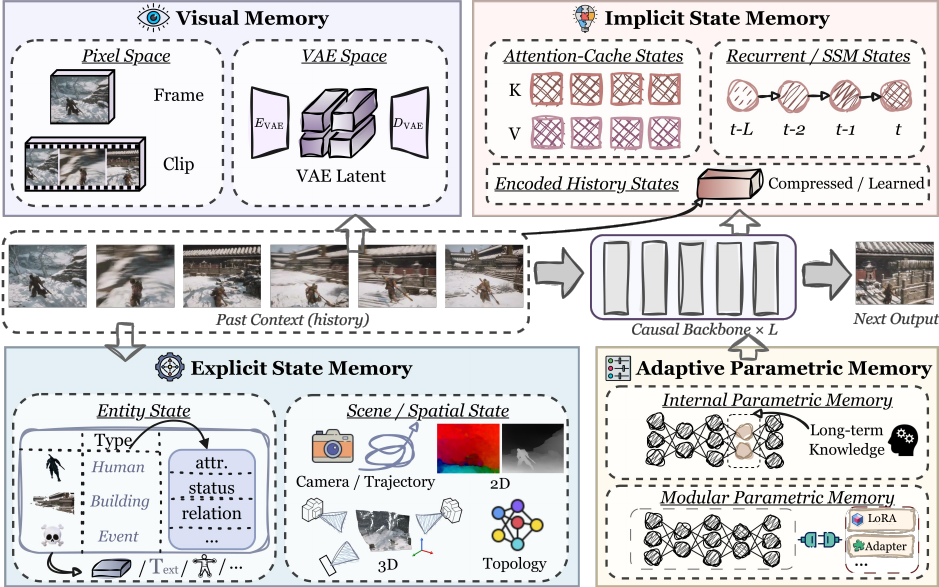}
\vspace{-0.6em}
\caption{Overview of memory carriers in autoregressive video generation. Historical context can persist as \textit{visual memory}, \textit{implicit state memory}, \textit{explicit state memory}, or \textit{adaptive parametric memory}, each comprising distinct carrier forms. These memory types can be combined to condition the causal backbone for subsequent generation.}
\label{fig:figure_forms}
\vspace{-0.6em}
\end{figure*}

Figure~\ref{fig:figure_forms} provides a schematic overview of these carrier families and their representative realizations. Tables~\ref{tab:carrier_taxonomy_a} and~\ref{tab:carrier_taxonomy_b} further summarize representative methods according to their memory-bearing objects and cross-cutting carrier properties, including storage growth, unit structure, access keys, interpretability, editability, and model coupling.

The first three categories realize memory as non-parametric runtime states or stored objects, whereas parametric memory places persistence in adapted parameter states. This taxonomy is defined by the form of the memory-bearing object rather than by the preservation goal it serves. In practice, a single AR video generation system may combine visual anchors, latent caches, structured states, and adaptive parameters within the same generation pipeline.

\subsection{Visual Memory}
\label{sec:VM}

\probox{
\textbf{Definition.}\quad Visual memory preserves historical information as observation-aligned visual evidence, either as directly renderable content or as latent representations derived from specific past frames, clips, or regions.
}
A visual memory unit may therefore be an RGB frame, a video clip, or a VAE latent, provided that it remains traceable to a specific historical observation or segment. Formally, letting $\mathbf{y}_i$ denote a historical visual unit, visual memory can be written as:
\begin{equation}
\mathbf M_n^{\mathrm{vis}}
=\{\mathbf m_i^{\mathrm{vis}}=E_{\mathrm{vis}}(\mathbf y_i)
\mid i\in\mathcal I_n^{\mathrm{vis}}\},
\qquad
\mathcal I_n^{\mathrm{vis}} \subseteq \{1,\ldots,n-1\},
\end{equation}
where $E_{\mathrm{vis}}$ may be the identity mapping for pixel-space units or an autoencoder, such as a VAE encoder, that maps observations into latent representations. The defining criterion is not whether a memory unit is directly human-readable, but whether it retains a traceable correspondence to the historical visual evidence from which it was derived.

Compared with more consolidated or structured state representations, visual memory retains high-fidelity appearance and spatial detail with relatively little abstraction. Its contents remain tied to identifiable historical observations, allowing past visual evidence to be reused when needed by subsequent generation. This fidelity, however, comes with two main costs. First, retaining observation-level evidence introduces substantial redundancy, so storage and computation can grow with rollout length unless memories are selectively retained or compressed. Second, because the stored evidence may itself be generated, visual artifacts or inconsistencies can be preserved and later reintroduced into subsequent generation.

We organize visual memory primarily by encoding space, distinguishing~\ding{192} \emph{pixel-space visual memory} from \ding{193} \emph{VAE-space visual memory}. The former retains directly renderable RGB observations, whereas the latter stores observation-aligned latents in an autoencoder or generator latent space.

\subsubsection{Pixel-Space Visual Memory}

Pixel-space visual memory retains historical observations in directly renderable RGB image or video form. In this subsection, $\mathbf{x}_t$ denotes the raw video frame at time $t$. A pixel-space memory unit may correspond to either an individual historical frame or a temporally contiguous clip:
\begin{equation}
    \mathbf{m}^{\mathrm{pix}}_i
    =
    \mathbf{x}_{s_i:u_i}
    =
    \left(
        \mathbf{x}_{s_i},
        \mathbf{x}_{s_i+1},
        \ldots,
        \mathbf{x}_{u_i}
    \right),
    \qquad
    s_i \leq u_i,
\end{equation}
where $s_i=u_i$ corresponds to frame-level memory and $s_i<u_i$ to clip-level memory. Each unit remains directly traceable to a specific historical frame interval, making it the most directly inspectable form of visual memory. It can be rendered, replaced, or re-encoded into the generator's native input representation when needed.

\vspace{-1.2em}
\paragraph{Frames.}
Frame-level memory stores complete historical frames as its basic units:
\begin{equation}
    \mathbf{M}^{\mathrm{frame}}_n
    =
    \left\{
        \mathbf{m}^{\mathrm{pix}}_i
        \mid
        i \in \mathcal{I}^{\mathrm{vis}}_n,
        \; s_i = u_i
    \right\}.
\end{equation}
Individual frames preserve fine-grained appearance, identity cues, texture, viewpoint-specific structure, and other visual details that may be difficult to reconstruct after the corresponding observation leaves the local context. 
Existing designs broadly follow three strategies. At the local end of this spectrum, recent-frame retention \citep{voleti2022mcvd,gao2024vid} (\textit{e.g.}, MCVD) provides dense short-range conditioning for local continuity. Such history primarily belongs to the active context when it is supplied densely at every step; it assumes a memory role when selected or retained beyond default local access. Sparse anchor frames \cite{ren2024consisti2v,yang2026gloria, li2026everanimate, tian2026streamchar, qiu2026ge, lai2026groundshot} (\emph{e.g.}, ConsistI2V) preserve stable long-range references; and retrieval-based approaches \cite{yu2025context,li2025vmem,li2026i3dm, guo2025end, zhu2026shareverse, li2026walking} (\emph{e.g.}, Context-as-Memory) reactivate non-local frames according to visual or geometric relevance. These choices trade coverage and accessibility against storage and reuse errors: dense retention is redundant, while sparse or retrieved references may become stale, misaligned, or inconsistent with the current state. More fundamentally, a single frame records an observed state but provides limited evidence about the temporal process that produced it.

\vspace{-1.2em}
\paragraph{Clips.}
Clip-level memory instead treats a temporally contiguous video segment as one memory unit:
\begin{equation}
    \mathbf{M}^{\mathrm{clip}}_n
    =
    \left\{
        \mathbf{m}^{\mathrm{pix}}_i
        \mid
        i \in \mathcal{I}^{\mathrm{vis}}_n,
        \; s_i < u_i
    \right\}.
\end{equation}
The retained segment may correspond to a recent generated chunk~\citep{yang2023video,gao2024ca2,weng2024art,song2025history,xie2025progressive} (\textit{e.g.}, LGC-VD, Ca2-VDM), an event-centered interval (\textit{e.g.}, LCT \citep{guo2025long}), or a retrieved historical episode (\emph{e.g.}, PlenopticDreamer~\citep{fu2026plenoptic}). Relative to frame memory, clips preserve not only appearance and scene state but also short-range motion, temporal order, and interaction dynamics, providing stronger temporal evidence across AR generation boundaries. This additional information comes at higher storage and conditioning cost, and previously observed motion may become an outdated prior when the current action, scene state, or intended dynamics have changed.

Pixel-space memory trades efficiency for direct access to high-fidelity historical evidence. Frame memory is well suited to visual anchoring and state recall, whereas clip memory additionally preserves short-range temporal evolution at greater cost. In both cases, reliable reuse requires retaining evidence that remains relevant without repeatedly propagating stale or erroneous observations.

\subsubsection{VAE-Space Visual Memory}

VAE-space visual memory retains historical observations in the latent space of a video VAE or a related visual autoencoder. Its key property is that the latent representation remains aligned with the source observation and can be approximately decoded back to visual space. For a historical segment $\mathbf{x}_{s_i:u_i}$, where $s_i=u_i$ denotes a single frame and $s_i<u_i$ a video clip, its clean VAE-space representation is:
\begin{equation}
  \mathbf{m}_i^{\mathrm{vae},0}
  =
  E_{\mathrm{VAE}}(\mathbf{x}_{s_i:u_i})
  \in
  \mathbb{R}^{T_i' \times H' \times W' \times C'},
  \qquad
  D_{\mathrm{VAE}}(\mathbf{m}_i^{\mathrm{vae},0})
  \approx
  \mathbf{x}_{s_i:u_i},
\end{equation}
where $E_{\mathrm{VAE}}$ and $D_{\mathrm{VAE}}$ denote the VAE encoder and decoder, respectively, and $T_i'$, $H'$, $W'$, and $C'$ denote the temporal, spatial, and channel dimensions of the latent representation. In particular, $T_i'$ depends on both the duration of the source segment and the temporal compression rate of the VAE. Although spatial and temporal compression may reduce resolution, the resulting latent remains tied to a bounded source interval: a temporal latent slice may correspond to a single frame under temporally uncompressed encoding or to a short group of adjacent frames under a temporally compressive video VAE. This observation alignment distinguishes VAE-space visual memory from generic hidden states whose contents need not correspond to identifiable past observations.

In diffusion-based generators, historical VAE latents may also be retained at an intermediate noise level $\tau$:
\begin{equation}
\mathbf{m}_i^{\mathrm{vae},\tau}
=
\alpha_{\tau}\mathbf{m}_i^{\mathrm{vae},0}
+
\sigma_{\tau}\boldsymbol{\epsilon}_i,
\qquad
\boldsymbol{\epsilon}_i
\sim
\mathcal{N}(\mathbf{0},\mathbf{I}),
\end{equation}
where $\alpha_{\tau}$ and $\sigma_{\tau}$ are determined by the diffusion noise schedule. The noise level does not change the carrier category: clean or noised latents function as memory only when they persist beyond the generation of their source frame or segment and are subsequently reused to condition later AR steps \citep{zhu2026astra}.

Existing systems realize VAE-space memory through several operational patterns. Sequential carry-over (\textit{e.g.}, LVDM \citep{he2022latent}) conditions a future chunk on preceding latent clips or prefixes. Windowed designs~\citep{kim2024fifo, xie2025progressive, chen2025ouroboros, mao2026yume1, yang2026decmem, huang2026learning, duan2026long} (\textit{e.g.}, FIFO-Diffusion, PA-VDM) retain a bounded queue or overlapping set of historical latents, sometimes assigning different noise levels to different parts of the retained context. Retrieval-based designs \citep{hu2026longlive, guo2026memorize, akdemir2026zero, guo2026contextmaster} (\emph{e.g.}, LongLive-RAG) instead preserve non-local latent blocks as separately accessible entries and reactivate them when distant visual evidence becomes relevant. These strategies differ in how memory is maintained and accessed, while sharing the same observation-aligned latent carrier.

Relative to decoded RGB history, VAE-space memory is more compact and more closely matched to the representation consumed by latent-space generators, reducing storage and conditioning overhead and, in some designs, avoiding repeated decoding and re-encoding. This efficiency comes at the cost of information loss and stronger model coupling: details removed by the encoder cannot be recovered from the stored latent, and representation changes across autoencoders or generator variants may limit reuse. Moreover, errors retained in generated latents can still be fed back into subsequent AR steps and accumulate over long rollouts. VAE-space memory therefore provides an efficient intermediate point between pixel-level fidelity and more strongly compressed or consolidated neural state.

\begingroup
\setlength{\tabcolsep}{3pt}
\begin{table*}[!t]
\renewcommand{\arraystretch}{1.06}
\vspace{-1em}
\caption{Carrier-centric taxonomy of selected memory mechanisms
(\textbf{Part~I}: Visual and Implicit State Memory).
\textbf{Carrier} denotes the persistent object read by a later generation step.
\textbf{SG} denotes resident-storage growth with rollout length (Fixed, Bounded, Linear, or scene-scaled);
\textbf{US} denotes unit structure (Itemized, Distributed, Structured, Entity, or Hybrid);
\textbf{Key} gives the signal or index used to access a retained historical unit; \textit{None} indicates no separately addressable historical unit;
\textbf{Int.}, \textbf{Edit.}, and \textbf{MC} denote interpretability, direct editability of instance-specific memory without retraining, and model coupling, respectively.
\lowmark~, \midmark~, and \highmark~denote low, medium, and high levels, respectively.}
\label{tab:carrier_taxonomy_a}
\vspace{-0.6em}
\centering
\footnotesize
\begin{tabular}{
    l|
    >{\raggedright\arraybackslash}p{7.2cm}
    cccccc
}
\hlineB{2.5}
\rowcolor{CadetBlue!20}
\textbf{Method} & \textbf{Carrier} & \textbf{SG} & \textbf{US} & \textbf{Key} & \textbf{Int.} & \textbf{Edit.} & \textbf{MC} \\
\hlineB{1.5}
\multicolumn{8}{c}{\cellcolor{gray!10}\textit{I. Visual Memory}} \\
\hline
WorldPack~\citep{oshima2025worldpack} & Geometrically packed historical frame tokens  & Boun. & Item. & Pose & \midmark & \midmark & \midmark \\
MemLearner~\citep{yu2026memlearner} & Historical video-latent frames  & Boun. & Item. & Learned query & \midmark & \midmark & \midmark \\
FramePack~\citep{zhang2026frame} & Progressively packed multiscale VAE-frame history  & Boun. & Item. & None & \midmark & \midmark & \midmark \\
LongLive-RAG~\citep{hu2026longlive} & Searchable historical VAE-latent blocks & Linear & Item. & Content & \midmark & \midmark & \midmark \\
MagicWorld~\citep{li2025magicworld} & Bounded generated-frame latent cache  & Boun. & Item. & Content & \midmark & \midmark & \midmark \\
WorldPlay~\citep{sun2025worldplay} & Pose/FOV-retrieved historical VAE chunks & Linear & Item. & Pose + time & \midmark & \midmark & \midmark \\
HyDRA~\citep{chen2026out} & Historical VAE latents tokenized at read time & Boun. & Item. & Learned query & \midmark & \midmark & \midmark \\
StreamingT2V~\citep{henschel2025streamingt2v} & First-chunk features + previous-frame condition & Fixed & Item. & None & \midmark & \lowmark & \midmark \\
MosaicMem~\citep{yu2026mosaicmem} & Geographically addressed historical latent patches & Linear & Item. & Spatial & \midmark & \highmark & \midmark \\
COVRAG~\citep{joo2026retrieve} & Retrieved frames with cached depth for selection  & Boun. & Item. & Pose + depth & \highmark & \midmark & \midmark \\
UCM~\citep{xu2026ucm} & Retrieved VAE frame tokens with pose/depth indices  & Boun. & Item. & Pose + depth & \highmark & \midmark & \midmark \\
VMem~\citep{li2025vmem} & Past views indexed by observed surfels & Linear & Item. & Surfel & \highmark & \midmark & \midmark \\
WorldMem~\citep{xiao2025worldmem} & Frame bank with pose/time metadata  & Boun. & Item. & Pose + time & \highmark & \midmark & \midmark \\
I3DM~\citep{li2026i3dm} & Historical frames with NVS-aligned injection & Linear & Item. & Pose + ray & \highmark & \midmark & \midmark \\
PlenopticDreamer~\citep{fu2026plenoptic} & Retrieved video--camera pairs  & Boun. & Item. & Pose & \highmark & \midmark & \midmark \\
Matrix-Game 3.0~\citep{wang2026matrix} & Camera-retrieved historical VAE frame latents & Linear & Item. & Pose & \highmark & \midmark & \midmark \\
Pathwise-TTC~\citep{Pathwise-TTC} & Persistent first-generated-frame correction anchor & Fixed & Item. & None & \highmark & \midmark & \midmark \\
ART-V~\citep{weng2024art} & First-generated-frame visual anchor & Fixed & Item. & None & \highmark & \midmark & \midmark \\
MemCam~\citep{gao2026memcam} & Compressed historical frame features & Linear & Item. & Pose & \highmark & \highmark & \midmark \\
Context-as-Memory~\citep{yu2025context} & Historical frames with camera poses & Linear & Item. & Pose & \highmark & \highmark & \lowmark \\

\hline
\multicolumn{8}{c}{\cellcolor{gray!10}\textit{II. Implicit State Memory}} \\
\hline
MAG~\citep{zhu2025memorize} & Compressed historical KV cache  & Boun. & Item. & None & \lowmark & \lowmark & \highmark \\
BIFE~\citep{InterVBench} & Semantics-sparse KV memory bank  & Boun. & Item. & Learned query & \lowmark & \lowmark & \highmark \\
EgoLCD~\citep{EgoLCD} & Sparse historical KV bank  & Boun. & Item. & Learned query & \lowmark & \lowmark & \highmark \\
SlotMemory~\citep{SlotMemory} & Slot-routed historical KV bank  & Boun. & Item. & Slot + content & \lowmark & \lowmark & \highmark \\
Closing the Loop~\citep{Closing-the-Loop} & Loop-retrieved clean KV chunks  & Boun. & Item. & Pose + depth & \lowmark & \lowmark & \highmark \\
Future Forcing~\citep{Future-Forcing} & Salience-pruned and merged KV cache  & Boun. & Item. & None & \lowmark & \lowmark & \highmark \\
OmniMem~\citep{OmniMem} & Retrieved historical KV blocks with pointer index  & Boun. & Item. & Learned query & \lowmark & \lowmark & \highmark \\
PackForcing~\citep{mao2026packforcing} & Sink, compressed-middle, and recent KV tiers  & Boun. & Item. & Time & \lowmark & \lowmark & \highmark \\
Surprise Forcing~\citep{Surprise-Forcing} & External historical value-token memory bank  & Boun. & Item. & Learned query & \lowmark & \lowmark & \highmark \\
Relax Forcing~\citep{zhao2026relax} & Sink, history, and tail KV memory  & Boun. & Item. & Time & \lowmark & \lowmark & \highmark \\
Anchor Forcing~\citep{yang2026anchorforcing} & Sink, junction, and local KV caches  & Boun. & Item. & Time & \lowmark & \lowmark & \highmark \\
LongLive~\citep{yang2025longlive} & Persistent sink KV plus rolling recent KV  & Boun. & Item. & Time & \lowmark & \lowmark & \highmark \\
Rolling Forcing~\citep{liu2026rolling} & Persistent sink KV plus rolling temporal KV  & Boun. & Item. & Time & \lowmark & \lowmark & \highmark \\
RELIC~\citep{hong2025relic} & Compressed full-history KV plus recent KV & Linear & Item. & Pose & \lowmark & \lowmark & \highmark \\
Context Forcing~\citep{chen2026contextforcing} & Sink, slow, and fast KV tiers  & Boun. & Item. & None & \lowmark & \lowmark & \highmark \\
Sparse Forcing~\citep{xu2026sparse} & Top-$C$ persistent remote KV blocks  & Boun. & Item. & None & \lowmark & \lowmark & \highmark \\
FadeMem~\citep{lu2026fademem} & Power-law consolidated hierarchical KV  & Boun. & Item. & Time & \lowmark & \lowmark & \highmark \\
Grounded Forcing~\citep{chen2026groundedforcing} & Diversity-updated global anchor KV bank  & Boun. & Item. & Time & \lowmark & \lowmark & \highmark \\
VideoSSM~\citep{yu2025videossm} & Recurrent global SSM state & Fixed & Dist. & None & \lowmark & \lowmark & \highmark \\
MALT~\citep{yu2025malt} & Recurrent hidden-state memory latents & Fixed & Dist. & None & \lowmark & \lowmark & \highmark \\
Hybrid Forcing~\citep{li2026hybridforcing} & Recurrent linear-attention KV state & Fixed & Dist. & None & \lowmark & \lowmark & \highmark \\
MemRoPE~\citep{kim2026memrope} & Dual-EMA recurrent KV state & Fixed & Dist. & None & \lowmark & \lowmark & \highmark \\
GIM-World~\citep{wei2026geometry} & Geometry-supervised memory-query tokens & Fixed & Dist. & Pose & \lowmark & \lowmark & \highmark \\
Long-Context SSM~\citep{po2025long} & Block-wise recurrent SSM state & Fixed & Dist. & None & \lowmark & \lowmark & \highmark \\
\hline
\multicolumn{8}{r}{\textit{Continued on next page.}} \\
\hlineB{2.5}
\end{tabular}%
\vspace{-2em}
\end{table*}
\endgroup

\begingroup
\setlength{\tabcolsep}{2pt}
\begin{table*}[!t]
\renewcommand{\arraystretch}{1.06}
\caption{Carrier-centric taxonomy of selected memory mechanisms (\textbf{Part~II}: Implicit State, Explicit State and Adaptive Parametric Memory). Column conventions follow Table~\ref{tab:carrier_taxonomy_a}.}
\label{tab:carrier_taxonomy_b}
\vspace{-0.6em}
\centering
\footnotesize
\begin{tabular}{
    l|
    >{\raggedright\arraybackslash}p{6.0cm}
    cccccc
}
\hlineB{2.5}
\rowcolor{CadetBlue!20}
\textbf{Method} & \textbf{Carrier} & \textbf{SG} & \textbf{US} & \textbf{Key} & \textbf{Int.} & \textbf{Edit.} & \textbf{MC} \\
\hlineB{1.5}
\multicolumn{8}{c}{\cellcolor{gray!10}\textit{II. Implicit State Memory}} \\
\hline
StateSpaceDiffuser~\citep{savov2025statespacediffuser} & Long-context Mamba state & Fixed & Dist. & None & \lowmark & \lowmark & \highmark \\
MemoryPack~\citep{wu2025memorypack} & Recurrent SemanticPack state & Fixed & Dist. & Prompt + reference & \lowmark & \lowmark & \highmark \\
ARL$^2$~\citep{li2026attend} & Clean-pass-updated linear-attention state & Fixed & Dist. & None & \lowmark & \lowmark & \highmark \\
CaR~\citep{peng2026compression} & Compressed contextual memory states & Boun. & Dist. & Pose & \lowmark & \lowmark & \highmark \\
Infinite-World~\citep{wu2026infinite} & Fixed-budget hierarchical history tokens & Fixed & Dist. & Learned query & \lowmark & \lowmark & \highmark \\
Echo-Infinity~\citep{bian2026echo} & Gated evolving memory-query state & Fixed & Dist. & None & \lowmark & \lowmark & \highmark \\
TinyHistory~\citep{zhang2025tinyhistory} & Appendable encoded-history embeddings & Linear & Item. & None & \lowmark & \lowmark & \highmark \\
\hline
\multicolumn{8}{c}{\cellcolor{gray!10}\textit{III. Explicit State Memory}} \\
\hline
MemoryWorld~\citep{zhou2026learning} & Updated world-coordinate DINO voxel map & Fixed & Struc. & Spatial & \midmark & \midmark & \midmark \\
LSM-World~\citep{wang2026latent} & World-space point cache + latent payload & Scene & Struc. & Pose + spatial & \midmark & \midmark & \midmark \\
PERSIST~\citep{garcin2026beyond} & Evolving latent 3D voxel world-frame & Fixed & Struc. & Pose & \midmark & \highmark & \midmark \\
VideoMemory~\citep{VideoMemory}  & Entity records plus reference-image banks & Linear & Hybrid & Entity & \highmark & \highmark & \highmark \\
CoTriSyGen~\citep{CoTriSyGen}  & Entity records + reference images & Fixed & Hybrid & Entity + state & \highmark & \highmark & \highmark \\
ReCA~\citep{MSVE-Bench}  & Typed narrative state + key/boundary frames & Linear & Hybrid & Narrative state & \highmark & \highmark & \highmark \\
A$^2$RD~\citep{long20262}  & Entity/event/camera records + visual media & Linear & Hybrid & Entity+event+pose & \highmark & \highmark & \highmark \\

EvoWorld~\citep{wang2025evoworld} & Colored point cloud & Scene & Struc. & Pose & \highmark & \lowmark & \lowmark \\
WorldStereo~\citep{WorldStereo} & Global point cloud + spatial-stereo views & Scene & Hybrid & Pose + 3D & \highmark & \lowmark & \midmark \\
Memory Forcing~\citep{huang2025memory} & Global point map + source frames & Scene & Hybrid & Point & \highmark & \lowmark & \midmark \\
Spatia~\citep{zhao2026spatia} & Point-cloud map + historical frames & Scene & Hybrid & Spatial & \highmark & \highmark & \midmark \\
DeepVerse~\citep{chen2025deepverse} & RGB/depth/raymap observations & Linear & Hybrid & Pose + depth & \highmark & \midmark & \highmark \\
ActionParty~\citep{pondaven2026actionparty} & Per-subject 2D coordinate states & Fixed & Entity & Entity & \highmark & \midmark & \highmark \\
AnchorWeave~\citep{wang2026anchorweave} & Pose-indexed local point-cloud memories & Linear & Struc. & Pose & \highmark & \midmark & \midmark \\
Voyager~\citep{huang2025voyager} & Culled, incrementally updated colored point-world cache & Scene & Struc. & Pose & \highmark & \midmark & \midmark \\
SPMEM~\citep{wu2026video} & TSDF/point map + sparse frames & Scene & Hybrid & Spatial & \highmark & \midmark & \midmark \\
LiveWorld~\citep{duan2026liveworld} & Static/dynamic point clouds + appearance frames & Scene & Hybrid & Spatial + entity & \highmark & \midmark & \midmark \\

\hline
\multicolumn{8}{c}{\cellcolor{gray!10}\textit{IV. Adaptive Parametric Memory}} \\
\hline

LaCT~\citep{zhang2026test}
& Large-chunk TTT-MLP fast weights
& Fixed & Dist.
& None & \lowmark & \lowmark & \highmark \\

RAD-TTT~\citep{chen2025rad}
& TTT-linear memory weights
& Fixed & Dist.
& None & \lowmark & \lowmark & \highmark \\
SlowFast-VGen~\citep{hong2025slowfast}
& Rollout-adapted temporal LoRA
& Fixed & Dist.
& None & \lowmark & \midmark & \highmark \\
ISPA~\citep{fu2026towards}
& Layer-wise $\Delta W$ $+$ residual KV cache
& Fixed$+$Linear & Hybrid
& None & \lowmark & \midmark & \highmark \\
HippoCampus~\citep{peng2026hippocampus}
& Online fast-weight global memory
& Fixed & Dist.
& None & \lowmark & \midmark & \highmark \\

\hlineB{2.5}
\end{tabular}
\end{table*}
\endgroup

\subsection{Implicit State Memory}
\label{sec:ISM}

\probox{
\textbf{Definition.}\quad Implicit state memory retains rollout history in persistent model-native neural states whose semantics are learned rather than explicitly prescribed by an observation-level decoding space or a predefined world-state schema. These states persist across autoregressive steps and condition subsequent generation.
}

Let $\mathbf{M}^{\mathrm{imp}}_n$ denote the implicit memory available before generating $\mathbf{y}_n$. Rather than requiring a static summary of a distant prefix, implicit memory may evolve with the rollout through architecture-specific state transitions:
\begin{equation}
  \mathbf{M}^{\mathrm{imp}}_{n+1}
  =
  \mathcal{S}_{\mathrm{imp}}
  \left(
    \mathbf{M}^{\mathrm{imp}}_n,
    \mathbf{y}_n,
    \mathbf{C}_n,
    \mathbf{c}
  \right),
\end{equation}
where $\mathcal{S}_{\mathrm{imp}}$ abstracts mechanisms, \textit{e.g.}, cache extension or replacement, recurrent or state-space transitions, and learned history encoding or compression. The resulting state may consist of itemized neural entries or a consolidated representation. Its defining property is that historical information is retained in a learned neural state without direct observation-level correspondence or an explicit semantic or geometric schema. Transient activations that disappear within a single generation step are therefore not considered memory.

Implicit states can retain information about appearance, motion, geometry, events, and other aspects of rollout history without committing to a hand-designed representation for each quantity. This flexibility enables tight integration with the generator's internal computation and supports learned aggregation or compression over time. The same implicit semantics, however, make stored information more difficult to interpret, localize, edit, or verify than observation-aligned or explicitly structured memory.

We organize implicit state memory according to how the persistent neural state is formed and propagated. \ding{192} \emph{Attention-cache states} retain historical key-value activations or derived cache entries associated with earlier tokens or blocks. \ding{193} \emph{Recurrent and state-space states} progressively consolidate history into an evolving neural state through recurrent or state-space transitions. \ding{194} \emph{Encoded history states} preserve dedicated neural representations produced from historical observations or states by a feature extractor, history encoder, tokenizer, or compressor; these representations may take the form of feature maps, embeddings, compact tokens, or dedicated memory slots.

\subsubsection{Attention-Cache States}

Attention-cache memory retains rollout history as persistent key-value (KV) states within Transformer attention layers. We distinguish such memory from a transient computational cache used only to accelerate inference. In our scope, a KV cache serves as memory when historical states are maintained and managed across AR steps so that they can influence later generation after the corresponding evidence is no longer guaranteed to remain in the active local context. For layer $\ell$, attention-cache memory can be represented as:
\begin{equation}
  \mathbf{M}^{\mathrm{KV}}_n
  =
  \left\{
    \left(
      K^{(\ell)}_i,
      V^{(\ell)}_i,
      p_i
    \right)
    \,\middle|\,
    i\in\mathcal{I}_n,\ \ell\in\mathcal{L}
  \right\},
\end{equation}
where $\mathcal{I}_n$ indexes the historical units retained at rollout step $n$, $\mathcal{L}$ denotes the attention layers carrying memory, and $p_i$ denotes associated positional or indexing metadata when present. The carrier remains an attention-cache state whether its entries are preserved exactly or transformed through selection, pooling, compression, quantization, or positional re-indexing.
Related cache organizations have close precedents in long-context language-model inference, where finite KV-cache budgets are managed through persistent sinks, selective retention, and non-uniform allocation \cite{zhang2023h2o,li2024snapkv,cai2024pyramidkv}. StreamingLLM \citep{xiao2024efficient}, for example, retains a small set of initial attention-sink states together with a rolling recent window.

\vspace{-1.2em}
\paragraph{Retention Structures.}
Existing systems differ first in how finite cache capacity is distributed over rollout history. Sink-plus-local designs, \emph{e.g.}, LongLive~\citep{yang2025longlive} and Rolling Forcing~\citep{liu2026rolling}, preserve a small set of early reference states together with a rolling recent cache, favoring stable anchors and local continuity over uniform historical coverage \citep{zhu2026omni}. Recent--remote designs, \emph{e.g.}, RELIC~\citep{hong2025relic} and Context Forcing~\citep{chen2026contextforcing},  retain recent entries at higher fidelity while compressing more distant history. Multi-tier caches, \emph{e.g.}, MemRoPE~\citep{kim2026memrope}, Anchor Forcing~\citep{yang2026anchorforcing}, and PackForcing~\citep{mao2026packforcing}, further separate historical states by temporal scale or interaction stage \citep{hong2025sneakpeek, wu2026echo}. Selective consolidation, \emph{e.g.}, Sparse Forcing~\citep{xu2026sparse} and FadeMem~\citep{lu2026fademem}, instead preserves salient remote blocks or progressively merges older states under a bounded budget~\citep{zhu2025memorize}. These designs trade historical coverage against memory cost: stronger compression or selection reduces cache growth but may remove low-frequency evidence whose relevance emerges only later.

\vspace{-1.2em}
\paragraph{Addressable Remote Cache.}
Recent work increasingly treats access to distant cache states as a separate challenge from retaining them \citep{xu2026teaching, ji2025memflow}. Geometry- or action-aware retrieval, \emph{e.g.}, WorldKV~\citep{yi2026worldkv} and ReWorld~\citep{chen2026reworld}, uses camera, action, or pose correspondence to reactivate historical KV states associated with relevant locations rather than relying only on temporal proximity \citep{xu2026teaching, zhang2026liveanimate, ji2025memflow, ye2026dysink}. Sparse remote retention, \emph{e.g.}, Wonder~\citep{xu2026wonder}, preserves selected full-fidelity distant context alongside recent local states, avoiding premature consolidation of potentially reusable evidence. Positional remapping, \emph{e.g.}, WorldTrace~\citep{wu2026addressable}, addresses a different access failure: distant KV states may remain stored yet become difficult to attend to when their temporal offsets fall outside the training distribution, and can therefore be reassigned to in-distribution virtual positions. Together, these approaches show that cache capacity and cache accessibility are distinct: historical evidence may remain present but still be difficult to recover reliably.

\vspace{-1.2em}
\paragraph{Specialized Cache Organization.}
Cache structure can also exploit heterogeneity within the attention mechanism \citep{tian2026head, chen2026past, zhao2026densitykv}. Head-aware policies, \emph{e.g.}, Dummy Forcing~\citep{guo2026efficient} and Forcing-KV~\citep{ji2026forcing}, allocate retention or pruning differently across attention heads according to their dependence on historical context or their structural and dynamic roles. Recall-and-refinement designs, \emph{e.g.}, TetherCache~\citep{meng2026tethercache}, additionally combine relevance-aware retrieval with lightweight modification of recalled states to reduce contamination from drifted historical activations. Such specialization can improve efficiency and robustness, but also makes memory quality depend on whether the inferred head roles, relevance signals, or correction rules remain valid as the rollout evolves.

Attention-cache memory retains relatively fine-grained neural evidence without immediately consolidating the entire history into a shared recurrent state. Its itemized or hierarchical organization can support selective access when suitable indices or retrieval mechanisms are available, while compression, consolidation, and specialization help control the cost of an expanding cache. The central trade-off is that larger caches preserve broader historical coverage but introduce competition from stale or irrelevant states, whereas bounded caches reduce cost at the risk of premature eviction; positional mismatch and erroneous historical activations can further make retained evidence difficult or unsafe to reuse.

\subsubsection{Recurrent and State-Space States}

Recurrent and state-space memory represents rollout history through an evolving neural state rather than a separate entry for each historical unit. Let $\mathbf{s}_n$ denote the recurrent state available before generating $\mathbf{y}_n$. The corresponding memory carrier can be written as:
\begin{equation}
  \mathbf{M}^{\mathrm{rec}}_n
  =
    \mathbf{s}_n, \qquad
  \mathbf{s}_n
  \in \mathcal{S},
\end{equation}
where $\mathcal{S}$ denotes the state space of the recurrent representation. 
The carrier may take the form of an LSTM or recurrent latent state, an SSM state such as Mamba~\citep{gu2023mamba}, or an accumulated linear-attention state. The broader idea of carrying compact historical state across segments has precedents in long-context sequence modeling, such as Recurrent Memory Transformer~\citep{bulatov2022recurrent} and Compressive Transformer~\citep{rae2019compressive}. Unlike itemized attention-cache memory, recurrent carriers progressively consolidate history into an evolving state rather than retaining independently addressable states for individual past units.

\vspace{-1.2em}
\paragraph{Recurrent Consolidation.}
Segment-level recurrence, \emph{e.g.}, MALT~\citep{yu2025malt}, carries a compact latent state from one generated segment to the next, allowing historical information to persist without retaining the complete preceding sequence \citep{ghafoorian2026rehyat, huang2026ultra, dalva2026adastate}. Global recurrent memory, \emph{e.g.}, MemoryPack~\citep{wu2025memorypack}, further maintains a persistent summary alongside a separate local visual context, separating long-range state from high-fidelity recent evidence \citep{liu2026worldweaver, gan2026taomate, gao2026visko, ding2026ripple}. Recurrent AR diffusion, \emph{e.g.}, RAD~\citep{chen2025rad}, extends this principle across different state realizations, including LSTM-, Mamba-, and test-time sequence states. These designs differ in the transition mechanism and state representation, but share the same principle of carrying history forward through repeated state updates rather than explicit historical retrieval.

\vspace{-1.2em}
\paragraph{State-Space and Local-Global Designs.}
State-space models, \emph{e.g.}, Long-Context SSM~\citep{po2025long} and StateSpaceDiffuser~\citep{savov2025statespacediffuser}, maintain structured recurrent states whose persistent size can remain independent of the rollout horizon. Local-global hybrids, \emph{e.g.}, VideoSSM~\citep{yu2025videossm}, Hybrid Forcing~\citep{li2026hybridforcing}, and ARL$^2$~\citep{li2026attend}, complement such consolidated remote memory with a short exact attention or visual path \citep{zhen2026soulx}. The local component preserves recent high-resolution evidence, while the recurrent component summarizes information that extends beyond the active window, providing a practical compromise between detailed local conditioning and bounded long-range state.

By removing the explicit historical-entry axis, recurrent and state-space memory can keep persistent state compact as the rollout grows, making it well suited to streaming generation. This efficiency comes from continual consolidation: once past evidence has been absorbed into the shared state, it cannot generally be revisited independently, making rare information susceptible to interference or irreversible loss and allowing state errors to propagate through subsequent updates. Local-global hybrids mitigate this trade-off by preserving exact recent evidence alongside a compact recurrent representation of more distant history.

\subsubsection{Encoded History States}

Encoded history memory retains rollout history in dedicated neural representations produced by a feature extractor, history encoder, tokenizer, or compressor. It can be expressed abstractly as:
\begin{equation}
  \mathbf{M}^{\mathrm{enc}}_n
  =
  E_{\mathrm{mem}}
  \left(
    \mathbf{y}_{\mathcal{I}_n},
    \mathbf{c}
  \right),
  \qquad
  \mathcal{I}_n \subseteq \{1,\ldots,n-1\}.
\end{equation}
where $E_{\mathrm{mem}}$ denotes a memory-encoding operator whose output is retained as a persistent neural state. Unlike attention-cache memory, the retained representation is not simply the native KV state of the generator; unlike recurrent memory, it need not arise solely through sequential state transitions. Its defining property is that historical information is explicitly re-encoded into a learned memory representation. Accordingly, encoded history states need not be visually decodable or maintain the observation-level correspondence required by VAE-space visual memory.

Encoded history states vary in how much structure from the source history they retain. Some preserve identifiable temporal or spatial correspondence to individual observations, while others maintain an ordered sequence of more abstract embeddings or tokens. Stronger compression can merge multiple historical units into a smaller latent set or fixed collection of memory slots, progressively relaxing correspondence to any particular source observation.

\vspace{-1.2em}
\paragraph{Source-Preserving Encodings.}
Source-aligned features retain neural representations extracted from identifiable historical chunks, preserving their temporal origin while moving beyond directly decodable visual latents. Appendable history embeddings, \emph{e.g.}, TinyHistory~\citep{zhang2025tinyhistory}, instead encode newly generated history into compact neural entries that remain ordered along a discrete history axis \citep{song2026interactiveavatar}. These designs preserve relatively fine-grained source structure and can maintain separable historical evidence, but their memory size may still increase as new observations are encoded.

\vspace{-1.2em}
\paragraph{Compressed and Slot-Based Encodings.}
Hierarchical or tokenized compression, \emph{e.g.}, Infinite-World~\citep{wu2026infinite}, summarizes longer histories into a smaller collection of latent memory units, reducing the cost of retaining distant context while weakening direct correspondence to individual observations. More consolidated representations, \emph{e.g.}, Echo-Infinity~\citep{bian2026echo} and GIM-World~\citep{wei2026geometry}, further encode history into compact sequences or fixed memory-query states whose individual units need not correspond to a specific past frame or segment. Increasing consolidation improves memory efficiency, but also makes recovery depend more strongly on what the encoder chose to preserve when the historical evidence was first compressed.

Encoded history memory provides a flexible middle ground between retaining native historical activations and collapsing history into a single recurrent state: its structure can range from source-aligned features to compact learned memory units. This flexibility introduces an information bottleneck, since evidence discarded during encoding may become relevant only after a later revisit, event, or interaction. Its effectiveness also depends on compatibility between the learned memory representation and the downstream generator, as retained information may remain difficult to exploit when the two representation spaces are poorly aligned.

\subsection{Explicit State Memory}
\label{sec:ESM}

\probox{
\textbf{Definition.}\quad Explicit state memory represents rollout history through persistent world-state variables with declared semantic, spatial, geometric, dynamic, or relational meanings. Rather than retaining historical evidence only in its original or learned latent form, it maintains interpretable state that can be updated as the generated world evolves.
}

Unlike visual memory, which preserves observation-aligned evidence, and implicit memory, which retains learned neural states, explicit memory assigns stored variables a predefined world-level meaning. Examples include entity identities and attributes, object poses, camera states, occupancy or depth fields, spatial relations, and event or interaction states. Importantly, explicit memory need not be an archive of past observations: it may instead maintain an estimate of the current world state inferred and updated from historical evidence. Conversely, attaching an opaque learned feature to an entity identifier or spatial coordinate does not by itself make the feature explicit; in hybrid representations, explicit variables and implicit features are treated according to their respective carriers.

We represent explicit memory as a collection of typed state variables:
\begin{equation}
\mathbf{M}^{\mathrm{exp}}_n
=
\left\{
  s_{n,i}
  =
  \left(
    \rho_i,
    \kappa_i,
    v_{n,i}
  \right)
\right\}_{i=1}^{N_n},
\qquad
\kappa_i \in \mathcal{T}_{\mathrm{exp}},
\end{equation}
where $\rho_i$ identifies the referent or scope of the state, $\kappa_i$ specifies its declared type, $v_{n,i}$ denotes its value at rollout step $n$, and $\mathcal{T}_{\mathrm{exp}}$ denotes the vocabulary of explicit state types. Depending on the representation, a variable may describe an entity, spatial structure, relation, dynamic state, or persistent consequence of an event or intervention. Historical records such as trajectories or transition logs also qualify when their fields retain a declared world-state meaning and are maintained to constrain subsequent generation.

Explicit states follow the same memory lifecycle introduced in Section~\S\ref{sec:2.4}: they may be read to condition generation and revised as new observations, generated outputs, or interventions provide evidence about the evolving world. Their maintenance may rely on geometric fusion, state estimation, an external simulator, or a learned transition model; what makes the carrier explicit is the declared meaning of the maintained state rather than the particular update mechanism. Accordingly, a target camera pose or action is a control signal rather than memory by itself, while a retained camera trajectory, action history, or action-induced state change can serve as explicit memory when it persists across AR steps.

We organize explicit state memory into \ding{192} \emph{entity-centric states} and \ding{193} \emph{spatial and geometric states}. Relational, dynamic, event, and transition variables are included according to the world-state referents they describe: for example, an object trajectory is entity-centric, whereas a camera trajectory is spatial. These categories characterize the form of the maintained world state; temporal snapshots, trajectories, versioned histories, and transition records are shared organizational patterns rather than separate carrier classes.

\subsubsection{Entity-Centric States}

Entity-centric memory maintains persistent world-state variables associated with individual tracked objects, characters, or agents throughout the rollout:
\begin{equation}
\mathcal{E}_n
=
\left\{
  \left(
    \mathrm{id}_k,
    e_{n,k}
  \right)
\right\}_{k\in\mathcal{K}_n},
\qquad
e_{n,k}
=
\left(
  \alpha_{n,k},
  p_{n,k},
  d_{n,k},
  s_{n,k}
\right),
\end{equation}
where $\mathrm{id}_k$ identifies an entity across rollout steps; $\alpha_{n,k}$ denotes typed attributes; $p_{n,k}$ denotes position or pose; $d_{n,k}$ denotes dynamic quantities such as velocity; and $s_{n,k}$ denotes interaction, lifecycle, or event status. Individual systems may instantiate only a subset of these fields, but each field classified as explicit must retain a declared world-level meaning. An opaque neural representation remains an implicit component even when associated with a persistent entity identifier.

\vspace{-1.2em}
\paragraph{Identity and Attribute States.}
Identity and attribute states bind a persistent entity identifier to interpretable properties such as category, role, and declared appearance attributes \citep{NarraStream-Bench, CoTriSyGen, VideoMemory}. By maintaining these properties independently of the current observation, they support object permanence and entity-specific state updates across occlusion, viewpoint changes, and visual variation. Low-dimensional entity states, \emph{e.g.}, ActionParty~\citep{pondaven2026actionparty}, associate persistent identities with compact pose or state variables that evolve during rollout. Richer structured records, \emph{e.g.}, A$^2$RD~\citep{long20262}, expose typed attributes, relations, or task states, although such representations appear primarily in specialized simulator-based or agentic settings among the systems considered here \citep{lai2026groundshot}. In continuous AR video generators, fine-grained appearance is therefore often carried by complementary visual or implicit memory rather than fully factorized entity records.

\vspace{-1.2em}
\paragraph{Dynamic and Interaction States.}
Dynamic and interaction states maintain time-varying, entity-bound quantities such as position, pose, velocity, visibility, contact, and lifecycle or event status. Separating these variables from rendered observations allows an entity state to continue evolving when the entity is occluded or outside the current view, and allows state-changing interactions to be recorded without requiring their visual evidence to remain locally available~\citep{yang2026spiral, kang2026egox, zhao2026population, chen2026code}. Action-conditioned state transitions, \emph{e.g.}, MultiGen~\citep{po2026multigen}, update entity pose or motion from external controls, while persistent off-screen dynamics, \emph{e.g.}, LiveWorld~\citep{duan2026liveworld} and WorldDirector~\citep{wang2026worlddirector}, maintain evolving object-motion states across intervals in which the corresponding entities are not directly observed \citep{li2026learning}. Event-aware entity memory, \emph{e.g.}, ActWorld~\citep{xiong2026actworld}, further couples persistent object-identity tokens with interaction-dependent event updates so that state-changing evidence can survive beyond the frames in which it occurs.

Across these designs, entity-centric memory separates world-state evolution from pixel synthesis by binding persistent identities to interpretable attributes and time-varying state variables. This supports localized state maintenance across occlusion, viewpoint changes, and interactions, but its reliability depends on both the adequacy of the declared state space and the correctness of its updates. Restricted kinematic or task-specific schemas may omit relevant aspects of an entity, while errors in state estimation, action binding, or temporal synchronization can propagate into trajectories that remain internally consistent yet physically or causally incorrect. Fine-grained appearance and other difficult-to-factorize information therefore often remain complementary responsibilities of visual or implicit memory.

\subsubsection{Spatial and Geometric States}

Spatial and geometric memory maintains persistent world-state variables tied to camera frames, image-plane locations, local or world coordinates, or topological structures:
\begin{equation}
\mathcal{X}_n^\text{geo}
=
\left\{
  \left(
    \xi_j,g_{n,j}
  \right)
\right\}_{j=1}^{N_n},
\qquad
\xi_j
\in
\Omega_{\mathrm{cam}}
\cup
\Omega_{\mathrm{view}}
\cup
\Omega_{\mathrm{local}}
\cup
\Omega_{\mathrm{world}}
\cup
\Omega_{\mathrm{topo}}.
\end{equation}
Here $\xi_j$ specifies the spatial referent or indexing domain of a state variable, while $g_{n,j}$ denotes a typed quantity such as pose, depth, occupancy, visibility, or connectivity. Learned features may be co-located with these structures, but only variables with declared geometric or spatial meaning are treated as explicit state; opaque feature payloads remain visual or implicit components.

\vspace{-1.2em}
\paragraph{Camera Poses and Trajectories.}
Camera-state memory retains historical viewpoints as extrinsic transforms and, when needed, intrinsics expressed in a common reference frame; a trajectory is a persistent sequence of such states rather than a camera command used only for the next generation step. Although camera state carries little scene content by itself, it provides the geometric reference needed to relate historical observations across viewpoints and distinguish observer motion from scene motion. Hybrid designs, \emph{e.g.}, WorldMem~\citep{xiao2025worldmem} and AnchorWeave~\citep{wang2026anchorweave}, associate pose records with stored frame tokens or local geometric fragments so that historical evidence can be aligned with later target views. The pose variables are explicit spatial state, whereas associated visual tokens or learned features retain their original carrier type. Their usefulness depends on accurate calibration and temporal synchronization, since pose drift or pose-evidence mismatch can make otherwise valid history geometrically inconsistent. A planned target pose remains a control signal unless it is subsequently retained as part of rollout history.

\vspace{-1.2em}
\paragraph{2D Scene Representations.}
View-aligned 2D memory preserves typed image-plane or ray-based fields such as depth, surface normals, segmentation, optical flow, visibility, or point coordinates across AR steps \citep{chen2025deepverse, WorldStereo}. Unlike RGB memory, these fields describe interpretable scene properties and can provide geometric or motion constraints for later generation. Multi-field temporal memory, \emph{e.g.}, WorldWeaver~\citep{liu2026worldweaver}, carries aligned RGB together with depth and optical-flow states, while autoregressive geometry conditioning, \emph{e.g.}, Geometry-as-Context~\citep{hu2026geometry}, propagates generated geometric fields from one view to the next. Such states remain view-dependent: they describe only the geometry in the source image plane and require correspondence or reprojection when the viewpoint changes. Retaining many views can therefore increase storage, and 2D fields alone do not provide a globally consolidated scene representation or persistent entity identity.

\vspace{-1.2em}
\paragraph{3D Scene Representations.}
World-aligned 3D memory consolidates historical evidence into points, voxels, surfels, meshes, Gaussians, or other geometric structures expressed in a shared coordinate frame \citep{chen2025teleworld, chen2026robust, zhao2026geostream, yin2026alaya}. Persistent maps, \emph{e.g.}, PERSIST~\citep{garcin2026beyond} and Voyager~\citep{huang2025voyager}, accumulate RGB-D or geometric evidence into reusable world-space state \citep{ren2025gen3c}, whereas fragment- or patch-based designs, \emph{e.g.}, AnchorWeave~\citep{wang2026anchorweave} and MosaicMem~\citep{yu2026mosaicmem}, retain localized geometric units that can be selected according to a target view \citep{qian2026matrix}. Geometry-indexed hybrid memory further uses persistent spatial structure to recover non-local visual evidence: Mem-World~\citep{zheng2026mem} associates wrist-view history with temporally evolving 4D surfels for manipulation under occlusion and camera motion, while DreamX-World~\citep{team2026dreamx} uses geometric correspondence to reactivate observations from previously explored regions. In such hybrids, the geometric scaffold is explicit state, whereas retrieved frames or learned feature payloads remain visual or implicit carriers. The main challenge is maintaining a reliable world model as the scene evolves: incorrect pose estimation or fusion can corrupt persistent geometry, while dynamic content must be distinguished from stable structure rather than indiscriminately absorbed into the map. Storage may also scale with explored scene extent, and view-dependent appearance often remains in a complementary visual carrier.

\vspace{-1.2em}
\paragraph{Topological Spatial Layouts.}
Topological memory represents scene organization through persistent nodes and typed edges that encode place or region connectivity, adjacency, containment, or other non-metric spatial relations. Editable layout states, \emph{e.g.}, MultiGen~\citep{po2026multigen}, maintain persistent vertices and edges across rollouts, while richer structured systems, \emph{e.g.}, A$^2$RD~\citep{long20262}, incorporate scene-graph or
relational records in more specialized settings. By abstracting away metric geometry and appearance, topological states provide a compact representation for coarse spatial organization and revisitation but cannot directly support viewpoint projection or rendering. Spatial edges belong to this category, whereas entity attributes remain entity-centric and non-spatial event or causal relations are better represented as state-transition records associated with the relevant entities or world states.

Spatial and geometric memory makes persistence explicit by anchoring world state to viewpoints, image planes, metric coordinates, or topological structure. Increasing spatial structure can support stronger cross-view consistency and scene revisitation, but also introduces dependence on calibration, correspondence, and state-fusion accuracy; practical systems therefore often combine explicit geometry with visual or implicit carriers for appearance and other information that is difficult to maintain as structured spatial state.

\subsection{Adaptive Parametric Memory}
\label{sec:PM}

\probox{
\textbf{Definition.}\quad Adaptive parametric memory retains sequence-, instance-, or experience-specific information in designated parameter states that are adapted from prior evidence, persist beyond the adaptation event, and influence subsequent generation through the model's forward computation.
}

Unlike visual, implicit, and explicit state memory, adaptive parametric memory places persistence in an adapted parameter state rather than in stored observations, activations, or declared world-state variables. The memory-bearing parameters may reside within the generator or in a structurally separated adaptive module. In either case, historical information influences later generation through the adapted forward computation rather than solely through a non-parametric memory input.

Let $\boldsymbol{\phi}^{\mathrm{mem}}_n$ denote the memory-bearing parameter state available before generating $\mathbf{y}_n$. We write:
\begin{equation}
\mathbf{M}^{\mathrm{par}}_n
=
\boldsymbol{\phi}^{\mathrm{mem}}_n
\in
\Phi_{\mathrm{adapt}},
\qquad
\boldsymbol{\phi}^{\mathrm{mem}}_{n+1}
=
\mathcal{A}
\left(
\boldsymbol{\phi}^{\mathrm{mem}}_n,
\mathcal{D}_n
\right),
\end{equation}
where $\Phi_{\mathrm{adapt}}$ denotes the space of parameters assigned a memory-bearing role, $\mathcal{A}$ denotes the adaptation procedure, and $\mathcal{D}_n$ denotes the sequence-, instance-, or experience-specific evidence used for adaptation. The state need not be updated at every AR step and may remain unchanged between adaptation events while continuing to affect subsequent generation.

Not all learned parameters qualify as parametric memory. Parameters obtained through ordinary pretraining or task-level fine-tuning constitute general model knowledge unless they are subsequently assigned an experience-dependent memory role. Likewise, an adapter, LoRA module \citep{hu2021lora}, or fast-weight layer \citep{ba2016using} is not parametric memory merely by architectural form; it qualifies when its adapted parameter state persists and carries information used by later generation. Optimized latent codes, input noise, or other non-parameter variables remain non-parametric carriers when the model parameters themselves are unchanged.

We distinguish two forms according to where the memory-bearing parameters reside in the architecture. \ding{192} \emph{Internal parametric memory} stores the adapted state within the generator's native computation, such as fast-weight sequence layers or selected backbone parameters. \ding{193} \emph{Modular parametric memory} stores it in a structurally separated adaptive component, such as a temporary LoRA module, dedicated fast-weight memory, or reusable parameter block.

\subsubsection{Internal Parametric Memory}

Internal parametric memory stores rollout-dependent information directly in adaptive generator parameters, such as fast-weight layers \citep{ba2016using} or selected backbone weights updated during generation. 
This avoids a separate memory bank but tightly couples information to the model, making individual memories hard to isolate or revise. This formulation builds on fast-weight methods and test-time training (TTT) \citep{sun2020test}, which encode experience via online parameter updates, and neural-memory architectures such as Titans~\citep{behrouz2024titans}.

\vspace{-1.2em}
\paragraph{Fast-Weight Recurrence.}
Fast-weight methods maintain history by repeatedly adapting internal sequence-model parameters during rollout. Large-chunk test-time training (\emph{e.g.}, LaCT~\citep{zhang2026test}) propagates TTT fast weights across AR video chunks while retaining sliding-window attention for detailed local context. The TTT instantiation of RAD~\citep{chen2025rad}, which we refer to as RAD-TTT, similarly realizes the global recurrent state through adaptive TTT weights while preserving an overlapping local attention path. In both cases, historical dependence is progressively consolidated into weights that remain active in subsequent forward passes.

\vspace{-1.2em}
\paragraph{Parametric Absorption.}
Internal parameter memory can also be formed by transferring historical context into existing generator weights rather than through recurrent TTT updates. ISPA~\citep{fu2026towards}, for example, estimates an instance-specific modulation of attention projection weights that compensates for historical KV states removed from selected layers, allowing part of the remote context to be absorbed into the model parameterization. This design converts an otherwise expanding activation history into a bounded parameter state without requiring the same update mechanism as fast-weight recurrence.

Across these designs, internal parametric memory exchanges explicit historical access for bounded persistence within the generator. Once evidence from different times has been consolidated into shared weights, its individual contributions become difficult to recover, and erroneous or stale adaptations may continue to affect later generation until the parameter state is revised or reset.

\subsubsection{Modular Parametric Memory}

Modular parametric memory retains experience-dependent information in an adaptive parameter block that is structurally separated from the base generator. Typical carriers include temporary LoRA parameters or dedicated fast-weight memory modules. Because the memory-bearing state is separated from the backbone, it can in principle be reset, stored, replaced, or reused more independently than internal parametric memory while still influencing generation through the augmented forward computation.

\vspace{-1.2em}
\paragraph{Rollout-Adaptive Modules.}
Temporary parameter modules can accumulate sequence-specific experience during causal generation. SlowFast-VGen~\citep{hong2025slowfast}, for example, updates temporal LoRA parameters from local input--output trajectories during inference and carries the adapted state into subsequent action-conditioned video generation, using the parameters as episodic memory beyond the immediate context. HippoCampus~\citep{peng2026hippocampus} adopts a complementary hybrid design in which a frozen video backbone is augmented with a lightweight fast-weight global memory that adapts online across chunks, together with sliding-window and hierarchically compressed token memories for more detailed historical evidence.

\vspace{-1.2em}
\paragraph{Reusable Parameter Memories.}
Parameter states can also be stored and reused as distinct memory units beyond a single rollout. TTOM~\citep{qu2026ttom}, for example, optimizes lightweight LoRA parameters from compositional generation experience and stores them in a keyed parametric memory that supports later insertion, retrieval, update, and deletion. Because TTOM operates over a stream of video-generation requests rather than a single causal AR video rollout, we treat it as a scope-qualified example that illustrates the broader potential of modular parameter memories rather than as a primary long-video AR mechanism.

Modularization improves the separability and lifecycle control of parametric memory but does not make its contents directly interpretable: information remains distributed across the adapted weights. Moreover, multiple parameter memories introduce additional questions of retrieval, composition, interference, and stale-state management, while rollout-adaptive modules may reinforce errors when generated experience is repeatedly absorbed into their parameters.

\section{Functions: What Does Memory Preserve?}
\label{sec:functions}

The memory-conditioned formulation in Section~\S\ref{sec:background} establishes \emph{why} temporal persistence is necessary, while the carrier taxonomy in Section~\S\ref{sec:representation} categorizes \emph{what} representational substrates carry historical information, including visual observations, model-native latent states, explicit world-state variables, and adaptive parameter states. These representational forms do not by themselves specify what historical information must remain effective in future generation. We therefore define:
\obsbox{
\textbf{Memory function} as the specific generative responsibility that a persistent state fulfills after the original evidence is no longer directly available in the bounded local context.
}
Under this view, memory is assessed not merely by how much history it retains, but by whether the retained state preserves the information required for future generation. Importantly, preservation does not imply freezing the generated world. Rather, memory should maintain properties that ought to remain invariant while allowing legitimate changes in pose, appearance conditions, motion, relations, and world state. We refer to this recurring requirement as \emph{selective invariance}.

\begin{figure*}[!t]
\centering
\vspace{-0.4em}
\includegraphics[width=0.9\linewidth]{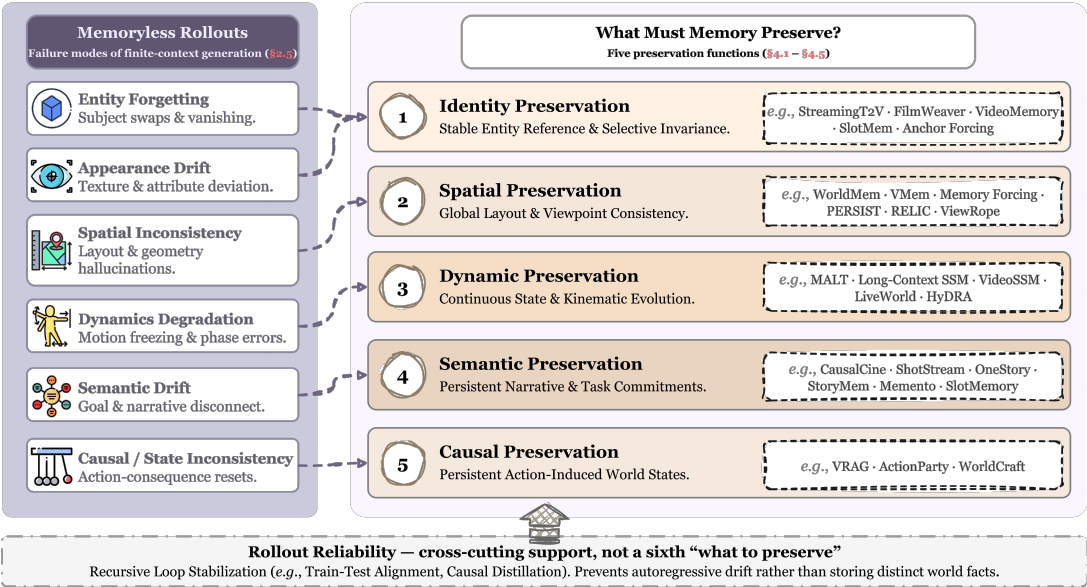}
\vspace{-0.6em}
\caption{Functional overview of memory mechanisms. (\textbf{\textit{Left}}) Long-horizon AR generation suffers from distinct failure modes when constrained by bounded local context. (\textbf{\textit{Right}}) We map these failure modes to five preservation functions that a robust memory system must fulfill for temporal persistence. (\textbf{\textit{Bottom}}) Rollout Reliability provides cross-cutting support for stabilizing the recursive self-generation loop rather than defining an additional preservation function.}
\label{fig:figure_function}
\vspace{-0.6em}
\end{figure*}

As summarized in Figure~\ref{fig:figure_function}, these functional responsibilities correspond to the failure modes of bounded-context rollouts identified in Section~\S\ref{sec:2.5}. Specifically, \ding{182} \emph{identity preservation} maintains instance-defining properties and mitigates entity forgetting and appearance drift; \ding{183} \emph{spatial preservation} maintains scene structure across navigation and revisitation; \ding{184} \emph{dynamic preservation} sustains motion and evolving state when their earlier evidence leaves the active context; \ding{185} \emph{semantic preservation} maintains rollout-specific roles, relations, events, and other high-level commitments; and \ding{186} \emph{causal preservation} maintains the delayed consequences of state-changing interventions. A single carrier may support multiple preservation functions, while a single function may be realized jointly by several carriers.

\begin{figure*}[!t]
\centering
\vspace{-0.4em}
\includegraphics[width=1\linewidth]{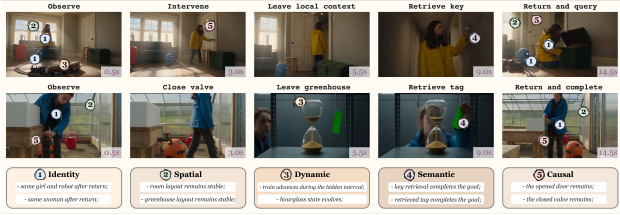}
\vspace{-1.6em}
\caption{Illustrative manifestations of memory functions in long-horizon video generation. Two example rollouts show how memory supports \ding{192} \textit{identity}, \ding{193} \textit{spatial}, \ding{194} \textit{dynamic}, \ding{195} \textit{semantic}, and \ding{196} \textit{causal} preservation across separated observations. Successful generation requires maintaining persistent entities and layouts, tracking evolving states, recalling task-relevant information, and preserving the consequences of past interventions beyond the local context.}
\label{fig:figure_function_2}
\vspace{-0.6em}
\end{figure*}

These preservation requirements often arise jointly in long-horizon generation, as illustrated in Figure~\ref{fig:figure_function_2}. Camera navigation primarily stresses spatial preservation, unobserved moving entities require dynamic preservation, and object manipulation introduces causal dependencies whose consequences may persist long after the original action. Multi-shot generation similarly places extended demands on identity and semantic preservation. We therefore treat interactive and multi-shot settings as application regimes that expose different combinations of preservation requirements rather than as additional memory functions. \Cref{tab:function_taxonomy_a,tab:function_taxonomy_b} summarize representative methods under this taxonomy, while the corresponding evaluation settings are discussed in Section~\S\ref{sec:evaluation}. The following subsections examine the five preservation requirements and the trade-offs involved in satisfying them over long autoregressive rollouts.

\subsection{Identity Preservation}
\label{sec:function_identity}

\probox{
\textbf{Definition.}\quad Identity preservation requires memory to maintain a stable reference to an individuated entity across temporal gaps, occlusion, viewpoint changes, and reappearance. It concerns whether the same entity is recovered, rather than whether all of its visible attributes, pose, or state remain unchanged.
}

Identity preservation is particularly stressed when local temporal continuity is disrupted by occlusion, shot transitions, prompt changes, or multi-subject interactions. Following the principle of \emph{selective invariance}, preserving identity does not require freezing appearance: pose, viewpoint, illumination, articulation, expression, clothing, and interaction state may legitimately evolve, while instance-defining properties such as facial structure, characteristic texture or markings, and distinctive geometry should remain stable. Excessively rigid reuse of historical evidence can suppress valid changes, whereas overly adaptive memory can gradually transform an entity into a plausible but different instance.

\vspace{-1.2em}
\paragraph{Reference Anchoring.}
The most direct strategy is to retain historical identity evidence as a long-range anchor rather than relying on local visual continuity alone. FilmWeaver~\citep{luo2026filmweaver} caches shot-level guidance, while Gloria~\citep{yang2026gloria} stores content anchors for long-horizon character consistency. StreamingT2V~\citep{henschel2025streamingt2v} preserves first-chunk appearance features, whereas LongLive-RAG~\citep{hu2026longlive} retrieves historical VAE blocks and MAG~\citep{zhu2025memorize} retains compressed KV states as longer-range references. Across multi-shot generation, OneStory~\citep{an2026onestory}, StoryMem~\citep{zhang2025storymem}, and Memento~\citep{Memento} selectively preserve or reconstruct subject evidence across scene transitions. Implicit states can serve a similar role: Anchor Forcing~\citep{yang2026anchorforcing} maintains anchor memory together with tri-region RoPE to distinguish persistent and local context. Recent streaming designs extend the same principle under bounded computation. Knot Forcing~\citep{xiao2025knot}, for example, retains reference-image KV states as a global identity anchor while combining them with a local sliding context and cross-chunk temporal knots; CineWeaver~\citep{huang2026cineweaver} maintains reference-conditioned anchors across shot transitions, while Visko Orbis~\citep{gao2026visko} uses bounded multi-scale memory during extended prompt-switchable rollouts. These mechanisms differ in carrier and retention policy, but share a common limitation: stale anchors may restore outdated appearance, suppress legitimate state changes, or over-constrain new viewpoints. Long rollouts therefore require selective updating, weighting, or invalidation of historical references.

\vspace{-1.2em}
\paragraph{Entity-Aware Addressing.}
Reference retention alone becomes insufficient when multiple entities coexist or reappear intermittently, because the generator must also determine \emph{whose} historical evidence should be recalled. VideoMemory~\citep{VideoMemory} maintains a dynamic visual--semantic memory bank, EM-Vid~\citep{vandersanden2026vid} indexes latent patches by entity, and SlotMem~\citep{SlotMem} stores role-wise character slots. IAMFlow~\citep{NarraStream-Bench} maintains an identity-aware entity registry, while SlotMemory~\citep{SlotMemory} organizes KV evidence around semantic objects. Such entity-aware organization reduces interference among visually similar or concurrently present subjects by exposing memory associated with the active instance. However, explicit addressing introduces a characteristic failure mode: \emph{binding error}. If the current observation or semantic role is matched to the wrong memory entry, the retrieved evidence may remain internally coherent yet belong to another entity, producing identity swaps or attribute leakage. Reliable identity preservation therefore requires stable associations among visual observations, semantic roles, and memory addresses throughout entry, absence, and reappearance.

\vspace{-1.2em}
\paragraph{Persistence Through Absence.}
A related requirement is \emph{object permanence}: an individuated entity should remain represented even while temporarily unobserved \citep{tian2026streamchar, zhu2026omni, vandersanden2026vid}. Without persistent state, a subject that leaves the camera frustum or disappears across a shot boundary may be forgotten, duplicated, or later reconstructed as a different instance. InfinityStory~\citep{elmoghany2026infinitystory} constrains multi-subject continuity across shot transitions, while HyDRA~\citep{chen2026out} retrieves appearance and motion cues when hidden subjects re-enter. More generally, reference anchors and entity-indexed states can maintain evidence for inactive subjects until they again become relevant. Persistence, however, should not reduce to replaying the last visible state. While absent, an entity may undergo plausible kinematic, semantic, or causal changes; its later appearance should therefore preserve identity while reflecting the world state that has evolved in the meantime.

Across these strategies, identity preservation depends on three related capabilities: retaining discriminative evidence, associating it with the correct entity, and maintaining that association through periods of absence. Increasing memory capacity alone does not guarantee identity persistence if references become stale or relevant evidence is routed to the wrong instance. Memory-revealing evaluation should therefore test reappearance after a verified information gap, preferably under increasing temporal, viewpoint, or shot separation, rather than relying on short-range visual similarity alone; EntityBench~\citep{he2026entitybench} provides one representative recurrence-based setting.

\begingroup
\setlength{\tabcolsep}{4.8pt}
\begin{table*}[!t]
\vspace{-1em}
\renewcommand{\arraystretch}{0.98}
\caption{Functional taxonomy of representative memory methods in AR video generation (\textbf{Part~I}: identity and spatial preservation). \textbf{Methods} are grouped by their \textit{primary} preservation function; \textbf{I/S/D/Sem/C} indicate multi-label coverage of identity, spatial, dynamic, semantic, and causal preservation. \textbf{Carrier} icons \visualicon~\impliciticon~\expliciticon~\paraicon~denote visual, implicit-state, explicit-state, and adaptive parametric memory, respectively. \textbf{MS/Int/LV} denote multi-shot, interactive, and long-video settings. Multiple labels may apply; associated benchmarks are discussed in Section~\S\ref{sec:evaluation}.}
\label{tab:function_taxonomy_a}
\vspace{-0.6em}
\centering
\footnotesize
\begin{tabular}{
    l|
    >{\raggedright\arraybackslash}p{7.4cm}
    ccccc
    c
    c
}
\hlineB{2.5}
\rowcolor{CadetBlue!20}
\textbf{Method} & \textbf{Mechanism} & \textbf{I} & \textbf{S} & \textbf{D} & \textbf{Sem} & \textbf{C} & \textbf{Carrier} & \textbf{Setting} \\
\hlineB{1.5}
\multicolumn{9}{c}{\cellcolor{gray!10}\textit{I. Identity Preservation}} \\
\hline
CineWeaver~\citep{huang2026cineweaver} & Reference-conditioned multi-shot anchors & \greenmark &   &   & \greenmark &   & \visualicon & MS, LV \\
Gloria~\citep{yang2026gloria} & Content anchors & \greenmark &   &   &   &   & \visualicon & MS, LV \\
StreamingT2V~\citep{henschel2025streamingt2v} & First-chunk appearance condition & \greenmark &   & \greenmark  &   &   & \visualicon & LV \\
LongLive-RAG~\citep{hu2026longlive} & Retrieved historical VAE blocks & \greenmark &   &   &   &   & \visualicon & LV \\
FilmWeaver~\citep{luo2026filmweaver} & Shot cache guidance & \greenmark &   & \greenmark &   &   & \impliciticon & MS \\
Anchor Forcing~\citep{yang2026anchorforcing} & Anchor memory + tri-region RoPE & \greenmark &   &   &   &   & \impliciticon & Int, LV \\
Knot Forcing~\citep{xiao2025knot} & Reference KV + cross-chunk temporal knots & \greenmark &   & \greenmark  &   &   & \impliciticon & Int, LV \\
MAG~\citep{zhu2025memorize} & Compressed historical KV cache & \greenmark & \greenmark  &   &   &   & \impliciticon & LV \\
UnityShots~\citep{UnityShots} & Boundary-aware AV summaries & \greenmark &  &  \greenmark &  &  & \impliciticon \visualicon & MS, LV \\
EM-Vid~\citep{vandersanden2026vid} & Entity-indexed latent patches & \greenmark &   &   &   &   & \expliciticon & MS \\
IAMFlow~\citep{NarraStream-Bench} & Identity-aware entity registry & \greenmark &   &   & \greenmark  &   & \expliciticon & MS, LV \\
VideoMemory~\citep{VideoMemory} & Dynamic visual--semantic bank & \greenmark &    &   & \greenmark  &   & \expliciticon \visualicon & MS, LV \\
SlotMem~\citep{SlotMem} & Role-wise character slots & \greenmark &   &   &   &   & \expliciticon \impliciticon & MS, LV \\
\hline
\multicolumn{9}{c}{\cellcolor{gray!10}\textit{II. Spatial Preservation}} \\
\hline
Matrix-Game 3.0~\citep{wang2026matrix} & Camera-retrieved VAE history & \greenmark  & \greenmark &   &   &   & \visualicon & Int, LV \\
Context-as-Memory~\citep{yu2025context} & Co-visible view retrieval &   & \greenmark &   &   &   & \visualicon & Int, LV \\
PlenopticDreamer~\citep{fu2026plenoptic} & Retrieved video--camera pairs &   & \greenmark & \greenmark  &   &   & \visualicon & Int, LV \\
MemCam~\citep{gao2026memcam} & Camera-conditioned frame memory &   & \greenmark &   &   &   & \visualicon & Int, LV \\
MemLearner~\citep{yu2026memlearner} & Learned context-memory query &   & \greenmark &   &   &   & \visualicon & Int, LV \\
I3DM~\citep{li2026i3dm} & 3D-aware historical-frame injection &   & \greenmark &   &   &   & \visualicon \impliciticon & Int, LV \\
MosaicMem~\citep{yu2026mosaicmem} & 3D-addressed latent-patch retrieval and composition &   & \greenmark &   &   &   & \visualicon \expliciticon & Int, LV \\
COVRAG~\citep{joo2026retrieve} & Coverage-maximizing retrieval &   & \greenmark &   &   &   & \visualicon \expliciticon & Int, LV \\
UCM~\citep{xu2026ucm} & Pose-aware retrieval with temporal PE warping &   & \greenmark &   &   &   & \visualicon \expliciticon & Int, LV \\
VMem~\citep{li2025vmem} & Surfel-indexed view retrieval &   & \greenmark &    &   &   & \visualicon \expliciticon & Int \\
Mem-World~\citep{zheng2026mem} & 4D surfel-indexed manipulation memory &   & \greenmark & \greenmark &   &   & \visualicon \expliciticon & Int, LV \\
DreamX-World~\citep{team2026dreamx} & Geometry-correspondence visual retrieval &   & \greenmark &   &   &   & \visualicon \expliciticon & Int, LV \\
INSPATIO-WORLD~\citep{inspatio2026world} & Spatiotemporal AR + spatial cache &  & \greenmark & \greenmark &  &  & \visualicon \impliciticon \expliciticon & Int, LV \\
HippoCampus~\citep{peng2026hippocampus} & Hierarchical history compression with online fast weights & \greenmark  & \greenmark & \greenmark  &   &  & \visualicon \impliciticon \paraicon & Int, LV \\

LingBot-World~\citep{team2026advancing} & Block-causal attention over cached history &   & \greenmark & \greenmark  &  \greenmark &  & \impliciticon & Int, LV \\
WorldKV~\citep{yi2026worldkv} & Camera/action-retrieved historical KV &   & \greenmark &   &   &   & \impliciticon & Int, LV \\
Wonder~\citep{xu2026wonder} & Sparse full-fidelity distant-state retention &   & \greenmark &   &   &   & \impliciticon & Int, LV \\
WorldTrace~\citep{wu2026addressable} & Addressable remapping of remote states &   & \greenmark &   &   &   & \impliciticon & Int, LV \\
ReWorld~\citep{chen2026reworld} & Pose-indexed landmark memory &   & \greenmark &   &   &   & \impliciticon & Int, LV \\
RELIC~\citep{hong2025relic} & Pose-aware compressed KV memory &   & \greenmark &   &   &   & \impliciticon & Int, LV \\
ViewRoPE~\citep{xiang2026geometry} & Geometry-aware RoPE &   & \greenmark &   &   &   & \impliciticon & Int, LV \\
Infinite-World~\citep{wu2026infinite} & Pose-free hierarchical latent mem. &   & \greenmark &   &   &   & \impliciticon & Int, LV \\
GIM-World~\citep{wei2026geometry} & Geometry-supervised fixed-size memory tokens &   & \greenmark &   &   &   & \impliciticon & Int, LV \\
WorldPack~\citep{oshima2025worldpack} & Geometry-aware frame compression &   & \greenmark &   &   &   & \impliciticon \visualicon & Int, LV \\

EvoWorld~\citep{wang2025evoworld} & Explicit panoramic 3D memory &   & \greenmark &   &   &   & \expliciticon & Int, LV \\
PERSIST~\citep{garcin2026beyond} & Editable latent 3D state &   & \greenmark &   &   & \greenmark & \expliciticon & Int, LV \\
WorldCam~\citep{nam2026worldcam} & Camera-pose geometric index &   & \greenmark &   &   &   & \expliciticon & Int \\
WorldMem~\citep{xiao2025worldmem} & Pose/time-indexed frame memory &   & \greenmark &   &   & \greenmark & \expliciticon \visualicon & Int, LV \\
Spatia~\citep{zhao2026spatia} & Updatable spatial memory &   & \greenmark &   &   &   & \expliciticon \visualicon & Int, LV \\
MoVerse~\citep{zhou2026moverse} & Panoramic Gaussian scaffold &   & \greenmark &   &   &   & \expliciticon \visualicon & Int, LV \\
WorldStereo~\citep{WorldStereo} & 3D geometric stereo memory &   & \greenmark &   &   &   & \expliciticon \visualicon & Int, LV \\
SPMEM~\citep{wu2026video} & Persistent 3D spatial cache &   & \greenmark &   &   &   & \expliciticon \visualicon & Int, LV \\
Memory Forcing~\citep{huang2025memory} & Point-to-frame geometric retrieval &   & \greenmark &   &   &   & \expliciticon \visualicon & Int, LV \\
LSM-World~\citep{wang2026latent} & Latent-space 3D cache &   & \greenmark &   &   &   & \expliciticon \visualicon & Int, LV \\
DeepVerse~\citep{chen2025deepverse} & RGB-D-pose overlap conditioning &   & \greenmark &  \greenmark &   &   & \expliciticon \visualicon & Int, LV \\
AlayaWorld~\citep{team2026alayaworld} & Reprojected 3D cache with temporal compression &   & \greenmark & \greenmark  &   &  & \expliciticon \impliciticon \visualicon & Int, LV \\
\hline
\multicolumn{9}{r}{\textit{Continued on next page.}} \\
\hlineB{2.5}
\end{tabular}%
\vspace{-3em}
\end{table*}
\endgroup

\begingroup
\setlength{\tabcolsep}{4.2pt}
\begin{table*}[!t]
\renewcommand{\arraystretch}{1.02}
\caption{Functional taxonomy of representative memory methods in AR video generation (\textbf{Part~II}: dynamic, semantic, and causal preservation, with cross-cutting rollout reliability).
Column conventions follow Table~\ref{tab:function_taxonomy_a}.
Rollout-reliability methods stabilize recursive generation rather than store a distinct world fact.}
\label{tab:function_taxonomy_b}
\vspace{-0.6em}
\centering
\footnotesize
\begin{tabular}{
    l|
    >{\raggedright\arraybackslash}p{7.4cm}
    ccccc
    c
    c
}
\hlineB{2.5}
\rowcolor{CadetBlue!20}
\textbf{Method} & \textbf{Mechanism} & \textbf{I} & \textbf{S} & \textbf{D} & \textbf{Sem} & \textbf{C} & \textbf{Carrier} & \textbf{Setting} \\
\hlineB{1.5}
\multicolumn{9}{c}{\cellcolor{gray!10}\textit{III. Dynamic Preservation}} \\
\hline
DFoT~\citep{song2025history} & History-guided flexible context &  &  & \greenmark &  &  & \visualicon & LV \\
VideoAR~\citep{ji2026videoar} & Next-frame/scale AR prediction &  &  & \greenmark &  &  & \visualicon \impliciticon & LV \\
Visko Orbis~\citep{gao2026visko} & Bounded multi-scale memory for prompt switching & \greenmark & \greenmark & \greenmark & \greenmark &  & \impliciticon & Int, LV \\
MALT~\citep{yu2025malt} & Memory-augmented latent transformers &  &  & \greenmark &  &  & \impliciticon & LV \\
Long-Context SSM~\citep{po2025long} & State-space world memory &  & \greenmark & \greenmark &  &  & \impliciticon & Int, LV \\
StateSpaceDiffuser~\citep{savov2025statespacediffuser} & Long-context SSM diffusion & \greenmark  & \greenmark & \greenmark &  &  & \impliciticon & Int, LV \\
VideoSSM~\citep{yu2025videossm} & Hybrid state-space memory & \greenmark &  & \greenmark &  &  & \impliciticon & LV \\
Grounded Forcing~\citep{chen2026groundedforcing} & Semantics vs.\ proximal dynamics & \greenmark &  & \greenmark & \greenmark &  & \impliciticon & LV \\
SANA-WM~\citep{SANA-WM} & Hybrid linear world modeling &  & \greenmark & \greenmark &  &  & \impliciticon & Int, LV \\
HyDRA~\citep{chen2026out} & Hybrid static/dynamic retrieval & \greenmark &  \greenmark & \greenmark &  & & \impliciticon \visualicon & Int, LV \\
RAD~\citep{chen2025rad} & Recurrent global memory + local attn. &  & \greenmark & \greenmark &  &  & \impliciticon \paraicon & LV \\

LiveWorld~\citep{duan2026liveworld} & Out-of-sight dynamic advancement &  &  & \greenmark &  & \greenmark & \expliciticon \impliciticon & Int, LV \\

SlowFast-VGen~\citep{hong2025slowfast} & Temporal LoRA fast weights &  & \greenmark & \greenmark &  &  & \paraicon & Int, LV \\
\hline
\multicolumn{9}{c}{\cellcolor{gray!10}\textit{IV. Semantic Preservation}} \\
\hline
OneStory~\citep{an2026onestory} & Adaptive subject/story memory & \greenmark &  &  & \greenmark &  & \visualicon & MS, LV \\
StoryMem~\citep{zhang2025storymem} & Multi-shot storytelling memory & \greenmark &  &  & \greenmark &  & \visualicon & MS, LV \\
Memento~\citep{Memento} & Reconstruct-to-remember subjects & \greenmark &  &  & \greenmark &  & \visualicon \impliciticon & MS, LV \\

CausalCine~\citep{meng2026causalcine} & Content-routed cached evidence & \greenmark &  &  & \greenmark &  & \impliciticon & MS, LV \\
ShotStream~\citep{luo2026shotstream} & Global/local streaming caches & \greenmark &  & \greenmark & \greenmark &  & \impliciticon & MS, Int, LV \\
SWIFT~\citep{tan2026swift} & Semantic-injection cache at prompt transitions &  &  & \greenmark & \greenmark &  & \impliciticon  & Int, LV \\
MemoryPack~\citep{wu2025memorypack} & Recurrent long-range semantic state &  &  & \greenmark & \greenmark &  & \impliciticon \visualicon & MS, LV \\

SlotMemory~\citep{SlotMemory} & Object-centric KV memory & \greenmark &  &  & \greenmark &  & \impliciticon \expliciticon & MS, LV \\
InfinityStory~\citep{elmoghany2026infinitystory} & Character-aware shot transitions & \greenmark & \greenmark &  & \greenmark &  & \expliciticon \visualicon & MS, LV \\
ReCA~\citep{MSVE-Bench} & Typed narrative state + key/boundary frames & \greenmark & \greenmark &  & \greenmark & \greenmark & \expliciticon \visualicon & MS, LV \\
\hline
\multicolumn{9}{c}{\cellcolor{gray!10}\textit{V. Causal Preservation}} \\
\hline
Khora~\citep{zhao2026population} & Population-scale multi-agent world state & \greenmark & \greenmark & \greenmark & \greenmark & \greenmark & \expliciticon & Int, LV \\
VRAG~\citep{chen2025vrag} & Explicit global state conditioning &  & \greenmark &  &  & \greenmark & \expliciticon \visualicon & Int, LV \\
WorldDirector~\citep{wang2026worlddirector} & Persistent object-motion world state & \greenmark & \greenmark & \greenmark & \greenmark & \greenmark & \expliciticon \visualicon & Int, LV \\
ActionParty~\citep{pondaven2026actionparty} & Subject-specific action binding & \greenmark & \greenmark &  &  & \greenmark & \expliciticon \impliciticon & Int \\
WorldCraft~\citep{gu2026worldcraft} & Object manipulation + memory refresh & \greenmark & \greenmark & \greenmark & \greenmark & \greenmark & \expliciticon \impliciticon & Int, LV \\
ActWorld~\citep{xiong2026actworld} & Action-aware event and object states & \greenmark & \greenmark & \greenmark &  & \greenmark & \expliciticon \impliciticon & Int, LV \\
\hline
\multicolumn{9}{c}{\cellcolor{gray!10}\textit{VI. Cross-cutting Rollout Reliability}} \\
\hline
FIFO-Diffusion~\citep{kim2024fifo} & FIFO latent queue &  &  &  &  &  & \visualicon & LV \\
FramePack~\citep{zhang2026frame} & Progressive frame-context packing &  &  &  &  &  & \visualicon & LV \\
Self Forcing~\citep{huang2026self} & Train-test rollout alignment &  &  &  &  &  & \impliciticon & LV \\
Causal Forcing~\citep{zhu2026causal} & Causal distillation for real-time AR &  &  &  &  &  & \impliciticon & Int, LV \\
Rolling Forcing~\citep{liu2026rolling} & Rolling long-horizon supervision & &  &  &  &  & \impliciticon & Int, LV \\
Context Forcing~\citep{chen2026contextforcing} & Long-context consistency training &  &  &  &  &  & \impliciticon & LV \\
PackForcing~\citep{mao2026packforcing} & Short-train long-sample forcing &  &  &  &  &  & \impliciticon & LV \\
Future Forcing~\citep{Future-Forcing} & Future-aware KV retention &  &  &  &  &  & \impliciticon & LV \\
Head Forcing~\citep{tian2026head} & Head-heterogeneous KV policy &  &  &  &  &  & \impliciticon & LV \\
SGF~\citep{zhuang2026self} & Future-loss supervised KV writing &  &  &  &  &  & \impliciticon & LV \\
MAGI-1~\citep{teng2025magi} & Chunk-wise AR rollout &  &  &  &  &  & \impliciticon & LV \\
NOVA~\citep{deng2025autoregressive} & Frame-wise AR without VQ &  &  &  &  &  & \impliciticon & LV \\
Echo-Infinity~\citep{bian2026echo} & Evolving memory-query state &  &  &  &  &  & \impliciticon & LV \\
LongLive~\citep{yang2025longlive} & Persistent sink KV with rolling recaching &  &  &  &  &  & \impliciticon & Int, LV \\
Infinity-RoPE~\citep{yesiltepe2025infinityrope} & Block-relativistic RoPE + KV flush/cut &  &  &  &  &  & \impliciticon & Int, LV \\
CausVid~\citep{yin2025slow} & Bidirectional-to-causal distillation + KV cache &  &  &  &  &  & \impliciticon & Int, LV \\
ISPA~\citep{fu2026towards} & Closed-form KV absorption into weights &  &  &  &  &  & \paraicon \impliciticon & LV \\
\hlineB{2.5}
\end{tabular}%
\end{table*}
\endgroup

\subsection{Spatial Preservation}
\label{sec:function_spatial}

\probox{
\textbf{Definition.}\quad Spatial preservation requires memory to maintain the layout, geometry, visibility, and viewpoint relations of the generated world across temporal gaps and camera motion. It concerns whether historical evidence remains spatially consistent and queryable from later viewpoints, rather than whether a particular entity identity or narrative fact is preserved.
}

Spatial preservation requires the generator to recover a coherent world configuration after relevant observations have left the bounded local context. The task therefore shifts from recalling \emph{what was observed} to determining \emph{what should be visible} from the current viewpoint. When a camera revisits a location or observes a known region from a new angle, geometry, object placement, and appearance should remain compatible with prior evidence while allowing legitimate occlusion, illumination, and scene changes. This requirement mitigates layout inconsistency, structural duplication, topological collapse, and failed scene revisitation. Unlike identity preservation, which associates history with entities, spatial preservation associates it with places, viewpoints, or world coordinates. Temporal recency is consequently a poor proxy for relevance: an old observation may overlap the target view, whereas the most recent frames may cover an unrelated region.

\vspace{-1.2em}
\paragraph{Spatially Addressed Retrieval.}
A common strategy is to retrieve historical evidence according to spatial correspondence rather than temporal proximity. WorldMem~\citep{xiao2025worldmem} indexes observations by pose and time, while Context-as-Memory~\citep{yu2025context} retrieves co-visible views for the target camera. WorldCam~\citep{nam2026worldcam}, COVRAG~\citep{joo2026retrieve}, MemCam~\citep{gao2026memcam}, I3DM~\citep{li2026i3dm}, PlenopticDreamer~\citep{fu2026plenoptic}, and ConsistWorld \citep{xu2026consistworldevidenceroutingconsistent} similarly select historical observations through camera correspondence, 3D-aware conditioning, or matched video--camera context. More explicit geometric keys are used by VMem~\citep{li2025vmem}, which indexes views through surfels, Memory Forcing~\citep{huang2025memory}, which maps visible 3D points back to source frames, and DeepVerse~\citep{chen2025deepverse}, which conditions retrieval on RGB-D-pose overlap. Learned or latent retrieval provides another realization: MemLearner~\citep{yu2026memlearner} learns to query contextual memory, whereas Matrix-Game~3.0~\citep{wang2026matrix} retrieves historical VAE chunks according to camera state. Recent systems extend the same principle to longer-range latent and cache histories. WorldKV~\citep{yi2026worldkv} reactivates historical KV blocks using camera and action correspondence, ReWorld~\citep{chen2026reworld} queries pose-indexed landmark memory under bounded local attention, and DreamX-World~\citep{team2026dreamx} retrieves non-local visual evidence through geometric correspondence. Mem-World~\citep{zheng2026mem} further associates manipulation history with temporally evolving 4D surfels, enabling recall under occlusion and wrist-camera motion. These mechanisms make spatial relevance explicitly queryable, but their reliability depends on correspondence accuracy: pose drift, depth error, calibration mismatch, or dynamic-object contamination can retrieve spatially incompatible evidence and produce ghosting, duplication, or geometric hallucination.

\vspace{-1.2em}
\paragraph{Persistent Scene State.}
Rather than repeatedly retrieving source observations, another family consolidates history into a persistent world-aligned state, such as point or voxel maps, occupancy fields, Gaussian representations, or structured spatial caches. SPMEM~\citep{wu2026video} maintains an incrementally updated 3D spatial cache, EvoWorld~\citep{wang2025evoworld} accumulates a panoramic 3D world representation, Spatia \citep{zhao2026spatia} updates a persistent spatial map, OctWorld \citep{lv2026octworld} incrementally fuses observations into a sparse octree map, and LSM-World \citep{wang2026latent} stores world-aligned latent features. PERSIST~\citep{garcin2026beyond} maintains an editable latent 3D state, MoVerse~\citep{zhou2026moverse} uses a panoramic Gaussian scaffold, and WorldStereo~\citep{WorldStereo} fuses geometric stereo memory with camera-guided generation. Hybrid systems can additionally use persistent geometry to recover appearance evidence: DreamX-World~\citep{team2026dreamx} reactivates observations from previously explored regions through geometric correspondence, while Mem-World~\citep{zheng2026mem} couples retrieved wrist-view history with evolving 4D surfels. Consolidating observations into a shared world frame supports navigation, loop closure, and rendering beyond the current field of view, but introduces state-correctness risks \citep{chou2026captain}. Pose or fusion errors can become persistent structural errors, dynamic content may be incorrectly absorbed into otherwise stable geometry, and scene-scaled representations may grow with explored extent. Once such a state is corrupted, later generation can repeatedly reinforce the same spatial inconsistency.

\vspace{-1.2em}
\paragraph{Implicit Spatial Alignment.}
Spatial consistency can also be maintained without reconstructing an explicit world model. WorldPack~\citep{oshima2025worldpack} geometrically packs historical frame tokens, RELIC~\citep{hong2025relic} retains pose-aware compressed KV states, and UCM~\citep{xu2026ucm} aligns historical tokens using time-aware pose information. ViewRoPE~\citep{xiang2026geometry} injects geometry-aware rotary embeddings, Infinite-World~\citep{wu2026infinite} maintains hierarchical latent history, and GIM-World~\citep{wei2026geometry} and WorldCrafter \citep{yu2026worldcrafterconsistentvideoworld} compress past observations into geometry-aware implicit tokens. MosaicMem~\citep{yu2026mosaicmem} combines localized 3D correspondence with generative completion, preserving spatial anchors while allowing regions that should change to be resynthesized. For long-range cache memory, accessibility itself also becomes important: Wonder~\citep{xu2026wonder} retains sparse full-fidelity distant states alongside recent context, whereas WorldTrace~\citep{wu2026addressable} remaps remote states whose temporal positions fall outside the training range so that retained evidence remains attendable. These approaches avoid the cost and rigidity of explicit reconstruction and integrate naturally with large generative backbones, but spatial correspondence is less directly inspectable. Historical information may remain stored yet become difficult to access or may preserve plausible appearance without enforcing metric geometry, making retrieval accessibility and geometric fidelity distinct failure sources.

Across these strategies, spatial preservation depends on both \emph{where} historical evidence is stored and whether the relevant evidence remains correctly accessible from a later viewpoint. Geometry-addressed retrieval preserves source observations, persistent scene states consolidate them into a shared world frame, and implicit alignment encodes spatial correspondence within latent or attention states. The resulting trade-off is not simply memory capacity versus reconstruction cost: failures may arise from missing coverage, incorrect spatial addressing, corrupted world-state updates, or weak geometric utilization. Scene revisitation therefore provides a useful memory-revealing condition because it requires spatial information to survive an observation gap and become actionable again from a changed viewpoint.

\subsection{Dynamic Preservation}
\label{sec:function_dynamic}

\probox{
\textbf{Definition.}\quad Dynamic preservation requires memory to maintain the evolving physical or behavioral state of entities and scenes across time, including motion, pose, velocity, visibility, and lifecycle status. It concerns continuity under temporal evolution, rather than persistent consequences specifically induced by an explicit intervention or action.
}

Dynamic preservation requires the generator to maintain evolving states after the evidence needed to infer them has left the bounded local context. While identity preservation determines \emph{what} persists and spatial preservation constrains \emph{where} it belongs, dynamic preservation concerns \emph{how} that state should continue to evolve. Failures therefore appear as motion freezing, velocity or pose drift, phase inconsistency, duplicated transitions, or implausible state changes. The relevant memory need not retain every past frame, but it must preserve sufficient information about temporal momentum, latent process state, or recent evolution to predict a state that remains compatible with the preceding trajectory.

\vspace{-1.2em}
\paragraph{State and Phase Continuity.}
The basic requirement is that future state remains consistent with prior evolution rather than reverting to the last observation or being resampled from a static prior. Different carriers realize this constraint at different levels. DFoT~\citep{song2025history} conditions generation on flexible recent history, while VideoAR~\citep{ji2026videoar} propagates dynamics through autoregressive frame- and scale-level prediction. MALT~\citep{yu2025malt} maintains recurrent latent memory, and Long-Context SSM~\citep{po2025long}, StateSpaceDiffuser~\citep{savov2025statespacediffuser}, VideoSSM~\citep{yu2025videossm}, and RAD~\citep{chen2025rad} consolidate temporal evolution into recurrent or state-space summaries. Grounded Forcing~\citep{chen2026groundedforcing} separates persistent semantic anchors from proximal dynamics, whereas SANA-WM~\citep{SANA-WM} and SlowFast-VGen~\citep{hong2025slowfast} maintain longer-horizon world or action dynamics through implicit or adaptive states. These mechanisms are especially important for phase-sensitive processes such as gait, rotation, or repeated articulation, where a small temporal offset can yield duplicated cycles, skipped transitions, or progressive phase drift. Their common trade-off is temporal compression: compact states reduce the cost of retaining motion history but may discard velocity, phase, or interaction cues that become decisive later, whereas richer histories preserve such cues at higher computational cost and expose more generated errors to subsequent prediction.

\vspace{-1.2em}
\paragraph{Out-of-Sight Evolution.}
Dynamic preservation becomes more demanding when state evolution continues while the corresponding entity or region is unobserved. An agent behind an occluder, a vehicle outside the camera frustum, or an articulated object undergoing continued motion should not remain frozen at its last visible state. LiveWorld~\citep{duan2026liveworld} advances unobserved dynamic entities against a persistent scene representation, HyDRA~\citep{chen2026out} retrieves static and dynamic evidence when hidden subjects re-enter, and INSPATIO-WORLD~\citep{inspatio2026world} combines spatiotemporal autoregression with spatial memory to maintain evolving world content. Recent explicit-state designs make this requirement more direct. WorldDirector~\citep{wang2026worlddirector} propagates semantic motion states independently of current visibility, allowing dynamic objects to evolve while another region is rendered; Mem-World~\citep{zheng2026mem} associates manipulation history with temporally evolving surface elements so that object state can be recovered after hand--object occlusion and rapid wrist-camera motion. These examples illustrate a distinction from causal preservation: dynamic memory predicts how an already evolving state progresses through time, whereas Section~\S\ref{sec:function_causal} concerns changes whose persistence is specifically induced by an intervention. In either case, simply replaying the last visible observation is insufficient; memory must maintain an estimate of the state reached during the interval of absence.

\vspace{-1.2em}
\paragraph{Stability under Self-Rollout.}
Dynamic state is particularly vulnerable to recursive generation because prediction errors alter the history from which subsequent motion is inferred. Small errors in velocity, pose, or phase can therefore accumulate into motion damping, jitter, repeated cycles, trajectory deviation, or eventual collapse toward static scenes. This is not only a retention problem: a memory state may faithfully preserve an already incorrect trajectory and thereby reinforce it. Training strategies that better expose the model to its own generated history, such as Self Forcing~\citep{huang2026self}, Causal Forcing~\citep{zhu2026causal}, Rolling Forcing~\citep{liu2026rolling}, and Context Forcing~\citep{chen2026contextforcing}, can reduce the mismatch between reference-conditioned training and recursive deployment. History-management designs such as FIFO-Diffusion~\citep{kim2024fifo}, FramePack~\citep{zhang2026frame}, and LongLive~\citep{yang2025longlive} provide complementary support by controlling how generated history remains available over extended rollouts. These mechanisms should be interpreted as reliability supports rather than additional dynamic facts stored in memory: the underlying requirement is to keep both the estimated state and the transition process sufficiently accurate as self-generated errors enter future conditioning.

Across these settings, dynamic preservation requires more than short-range temporal smoothness. Memory must retain enough information about motion and latent process state to continue evolution across temporal gaps, update that estimate while entities are unobserved, and prevent recursive prediction errors from progressively changing the implied dynamics. Different visual, recurrent, explicit, and parametric states provide different balances between temporal detail and compression, but larger or longer memory is useful only when the retained dynamics remain correct and exploitable by later generation.

\subsection{Semantic Preservation}
\label{sec:function_semantic}

\probox{
\textbf{Definition.}\quad Semantic preservation requires memory to maintain high-level commitments established earlier in the rollout, such as goals, roles, relations, events, instructions, and narrative facts. It concerns consistency of meaning and interpretation, rather than instance-level identity, metric layout, or physical state evolution alone.
}

Semantic preservation requires the generator to uphold commitments that remain relevant after their original evidence has left the bounded local context. While identity, spatial, and dynamic preservation constrain \emph{what} persists, \emph{where} it belongs, and \emph{how} its physical state evolves, semantic preservation governs what the generated world has established, implied, completed, or left unresolved. Failures therefore appear as role conflation, forgotten goals, contradictory relations, repeated or skipped events, and narrative drift. A static initial prompt is insufficient because many such commitments emerge during the rollout itself and must continue to constrain later generation unless they are legitimately revised.

\vspace{-1.2em}
\paragraph{Commitment and Referential Grounding.}
A semantic commitment is useful only if both its content and its referent remain available. Once generation establishes a role, relation, instruction, or scene-level constraint, subsequent outputs should remain compatible with it unless later evidence explicitly changes that state. MemoryPack~\citep{wu2025memorypack} maintains a recurrent long-range semantic state alongside local visual context, while CausalCine~\citep{meng2026causalcine} routes cached evidence according to content relevance and SlotMemory~\citep{SlotMemory} organizes historical KV evidence around semantic objects. These mechanisms differ in representation and retrieval policy, but address the same requirement: previously established meaning must remain bound to the intended entity, location, or role.
If this association fails, the model may remember an abstract relation yet assign it to the wrong subject, producing role swaps, misplaced objectives, or contradictory interactions; conversely, an ungrounded textual summary may preserve narrative facts without providing sufficient visual evidence to realize them consistently.

\vspace{-1.2em}
\paragraph{Event and Narrative State.}
Long-horizon generation must also track how events and higher-level plans progress over time. A sequence can remain locally plausible while violating its own history, for example, by repeating a completed action, forgetting that an objective has been resolved, or rendering a participant absent without a valid transition. ReCA~\citep{MSVE-Bench} maintains typed narrative state together with key and boundary frames, providing a more explicit representation of cross-segment story progression. InfinityStory~\citep{elmoghany2026infinitystory} maintains character-aware transitions, and ShotStream~\citep{luo2026shotstream} combines global and local streaming context. 
OneStory~\citep{an2026onestory}, StoryMem~\citep{zhang2025storymem}, and Memento~\citep{Memento} further maintain or reconstruct subject- and story-level evidence across shots \citep{duan2026long}.
These designs illustrate why semantic memory is often more selective than dense visual history: event status, role assignment, and unresolved goals may be temporally sparse but remain binding for long intervals. Compression is therefore useful, but excessive abstraction can discard visually subtle facts that determine whether a narrative transition remains valid.

\vspace{-1.2em}
\paragraph{Adaptation to Contextual Shifts.}
Semantic preservation becomes particularly difficult when external conditions change during generation. Prompt updates, shot boundaries, and interaction-dependent instructions may legitimately redirect the rollout, but new conditions should revise only the commitments they supersede rather than indiscriminately overwrite prior state. SWIFT~\citep{tan2026swift} addresses prompt transitions by augmenting visual history with a semantic-injection cache, adapting the active temporal window, and retaining segment-level semantic anchors. Visko Orbis~\citep{gao2026visko} uses bounded multi-scale memory to preserve subjects, scenes, and visual style during continuous prompt switching, while CineWeaver~\citep{huang2026cineweaver} carries reference-conditioned anchors across multi-shot transitions to maintain subject and narrative coherence. These systems expose semantic updating as a selective-revision problem: retaining every previous condition can make memory stale and over-constrain a new segment, whereas aggressive replacement can erase commitments that should remain valid. Over long rollouts, repeated local update errors can accumulate into global semantic drift even when each individual segment remains fluent and visually plausible.

Across these mechanisms, semantic preservation requires persistent commitments, correct grounding, and controlled revision as the generated trajectory evolves. Long-term retrieval, cross-shot conditioning, object-centric organization, and compact narrative states provide different ways to maintain such information, but success depends on preserving the \emph{meaning that remains operative} rather than merely retaining more history. Memory-revealing evaluation should therefore test delayed consistency of roles, relations, goals, or event state after the supporting evidence is no longer locally available, rather than relying on local prompt alignment or narrative fluency alone.

\subsection{Causal Preservation}
\label{sec:function_causal}

\probox{
\textbf{Definition.}\quad Causal preservation requires memory to maintain the persistent consequences of state-changing actions or interventions so that later generation remains consistent with the resulting world state. It concerns changes that remain operative because an intervention occurred, rather than autonomous temporal evolution or semantic coherence alone.
}

Causal preservation requires the generator to retain intervention-induced state changes after the originating action and its immediate visual evidence have left the bounded local context. The relevant question is not merely what the world previously looked like or how it would evolve on its own, but \emph{what has been changed and must remain changed}. Camera motion alters observation without modifying the underlying world state; opening a door, relocating an object, or transferring a tool instead establishes a new state that later synthesis must respect. Failures therefore appear as action amnesia, state reset, restoration of obsolete configurations, or downstream behavior inconsistent with the intervention. Causal preservation is particularly important in interactive generation, where user or agent actions can modify the world over long temporal gaps.

\vspace{-1.2em}
\paragraph{Intervention-State Binding.}
The first requirement is to bind each state-changing intervention to the entity, location, relation, or state variable that it modifies \citep{hao2026egosim, zhao2026population}. VRAG~\citep{chen2025vrag} conditions generation on a global state, ActionParty~\citep{pondaven2026actionparty} maintains subject-specific action bindings, and WorldMem~\citep{xiao2025worldmem} associates interactive scene state with pose- and time-indexed observations. More explicit event-centered memory is used by ActWorld~\citep{xiong2026actworld}, which retains object-identity and event-update tokens while routing history according to interaction importance, preventing brief but consequential transitions from being discarded by recency-biased compression. WorldDirector~\citep{wang2026worlddirector} similarly maintains explicit object-motion states that continue to constrain later chunks after the initiating command or affected object is no longer visible. Across these designs, recording that an action occurred is insufficient: memory must identify \emph{what changed} and \emph{which new state became operative}. Incorrect binding can produce action-object swaps, misplaced effects, or interventions that are acknowledged semantically but never grounded in the rendered world.

\vspace{-1.2em}
\paragraph{Persistence and Consequence Propagation.}
Once an intervention changes the world, the updated state must remain effective after the affected region leaves the active context and must constrain subsequent possibilities \citep{chen2025teleworld, chen2026code}. PERSIST~\citep{garcin2026beyond} maintains an editable latent 3D world state, while WorldCraft~\citep{gu2026worldcraft} refreshes memory after object manipulation so that later views reflect updated object positions rather than the pre-intervention configuration. This requirement distinguishes causal preservation from ordinary spatial revisitation: reconstructing a previously seen region is insufficient if the reconstructed state ignores modifications that occurred in the meantime. Moreover, interventions can alter future constraints as well as appearance, \textit{e.g.}, opening a door changes navigability, relocating an obstacle changes reachable trajectories, and transferring an object changes subsequent ownership or interaction possibilities. The memory target is therefore not the isolated action token but the post-intervention world state and the constraints implied by it. A common failure is \emph{prior reversion}: because the modified state may be sparse or locally observed, the generator can fall back to a visually likely canonical configuration after the evidence disappears, even when identity, geometry, and short-term motion remain individually plausible.

\vspace{-1.2em}
\paragraph{Ordered Revision and Conflict Resolution.}
Causal state is mutable: later interventions may extend, reverse, or supersede earlier ones. Memory must therefore preserve not all historical versions equally, but the currently operative state induced by their ordered sequence. A door opened in one segment and closed in a later segment should subsequently remain closed; retaining both observations without resolving their temporal relation can instead produce oscillation, state averaging, or retrieval of an obsolete configuration. Causal preservation consequently requires revision, overwrite, and invalidation in addition to retention. This creates a reliability trade-off: aggressive overwriting may erase still-valid consequences, whereas indiscriminate accumulation leaves stale evidence available to compete with newer state. The central challenge is thus to maintain a revisable world state whose updates respect intervention order and whose current version remains sufficiently influential to override obsolete observations and generic model priors.

Across these mechanisms, causal preservation requires three linked capabilities: grounding an intervention to the state it changes, maintaining and propagating the resulting consequences after local evidence disappears, and revising that state when later interventions supersede it. This makes causal preservation stricter than spatial revisitation or autonomous out-of-sight evolution: the state recovered after a gap must be the \emph{post-intervention} state, not simply a plausible continuation of the last observation. Memory-revealing evaluation should therefore separate observation-changing navigation from state-changing intervention protocols and test whether delayed outputs still depend on the earlier action once its local evidence has been removed.

\subsection{Coupled Functions}
\label{sec:function_coupling}

The five preservation functions provide distinct analytical views of what historical information must remain effective, but practical long-horizon generation often requires several of them simultaneously. An entity may retain its visual identity yet reappear at an incompatible location; a scene may close a spatial loop while reverting an earlier intervention; or a character may remain visually consistent while violating an established role. Such failures arise because identity, spatial, dynamic, semantic, and causal constraints must remain mutually compatible as the generated world evolves. Importantly, this coordination does not require a single unified representation: different carriers may maintain complementary aspects of history, provided that their contributions remain consistent when later generation depends on them.

\vspace{-1.2em}
\paragraph{Referential Coupling.}
Identity and semantic preservation interact through \emph{referential grounding}. Commitments such as ownership, task assignment, social roles, or object relations have generative force only when they remain associated with the intended entity; conversely, preserving visual appearance is insufficient if the model forgets what that entity represents or how it relates to others. A failure of this association can therefore produce a visually stable character assigned the wrong role, or a correctly remembered relation grounded to the wrong instance. Reliable long-horizon generation must consequently maintain compatibility between instance-level evidence and higher-level identifiers, roles, and relations, while still allowing legitimate changes in appearance or semantic state.

\vspace{-1.2em}
\paragraph{Spatio-Temporal Coupling.}
Spatial and dynamic preservation interact whenever entities evolve within a persistent environment. A moving object must remain consistent not only in velocity, phase, or articulation, but also in its trajectory relative to scene geometry; similarly, a persistent spatial representation must distinguish stable structure from entities whose positions or visibility change over time. If this coordination fails, a generator may synthesize locally plausible motion that crosses incompatible geometry, reintroduce an object at a position inconsistent with its trajectory, or preserve a stable map while populating it with temporally disconnected states. Dynamic information therefore needs to remain spatially grounded, whether through explicit coordinates and geometric states or through implicit pose- and view-aligned representations.

\vspace{-1.2em}
\paragraph{Intervention Coupling.}
Causal preservation often places joint demands on the other functions because an intervention must remain attached to the correct entity, grounded in the correct region, propagated through subsequent state evolution, and reflected in later semantic commitments. Transferring a tool, for example, can simultaneously modify its location, ownership or assignment, subsequent motion, and future interaction possibilities. Preserving only one component is insufficient: a model may remember that an action occurred while rendering the pre-intervention state, apply the change to the wrong object, or reproduce the immediate visual edit without its downstream consequences. Causal consistency therefore depends on coordinated preservation of the state variables that jointly define the post-intervention world, while later interventions may legitimately revise a subset of those constraints.

\vspace{-1.2em}
\paragraph{System-Level Coordination.}
The functional view specifies \emph{what} must persist; realizing these requirements depends on how memory is written, read, updated, managed, and integrated throughout the rollout. A system may preserve identity through visual anchors, spatial structure through a world-aligned state, and semantic or causal commitments through entity- or event-level variables, yet still fail if these memories become stale, are retrieved for the wrong query, or exert insufficient influence on later synthesis. Functional claims are therefore most informative when interpreted together with the temporal or intervention gap, memory carrier and access mechanism, and computational regime under which persistence is demonstrated. Rollout-reliability approaches provide an additional cross-cutting support: Self Forcing~\citep{huang2026self} exposes generation to self-produced histories, Rolling Forcing~\citep{liu2026rolling} extends supervision across causal rollouts, Context Forcing~\citep{chen2026contextforcing} targets long-context consistency, and PackForcing~\citep{mao2026packforcing} controls how historical context is represented over extended generation. These mechanisms can improve whether preservation responsibilities survive recursive self-generation, but do not constitute an additional content function. With the functional requirements established, Section~\S\ref{sec:operations} next examines the memory lifecycle that realizes them, followed by their learning in Section~\S\ref{sec:learning} and evaluation in Section~\S\ref{sec:evaluation}.

\section{Operations: How Does Memory Work?}
\label{sec:operations}

\begin{figure*}[!t]
\centering
\vspace{-0.4em}
\includegraphics[width=1\linewidth]{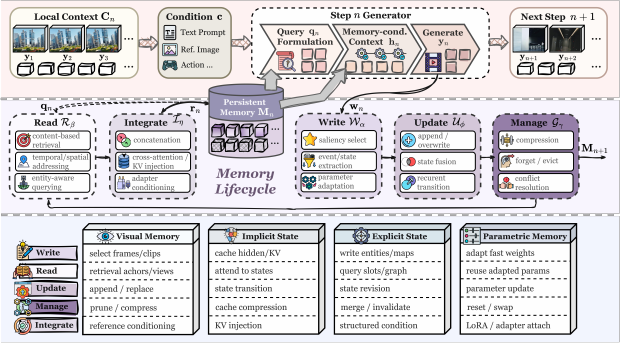}
\vspace{-1.6em}
\caption{Operational lifecycle of memory in autoregressive video generation. (\textbf{\textit{Top}}) At each generation step, relevant information is read from persistent memory and integrated with the local context and external conditions; the resulting output is then written back, updated, and managed to form the memory state for subsequent steps. (\textbf{\textit{Bottom}}) Representative realizations of these operations across visual, implicit-state, explicit-state, and parametric memory.}
\label{fig:figure_operation}
\vspace{-0.6em}
\end{figure*}

The preceding sections characterize memory by its representational carriers and preservation responsibilities. A persistent state $\mathbf{M}_n$, however, becomes effective only through operations that determine how historical information enters memory, remains accessible, evolves over time, and influences subsequent generation. We therefore view memory as an active interface that mediates between past evidence and future generation, governed by five core operations:
\obsbox{
\begin{itemize}[leftmargin=1.6em]
    \item[\ding{182}] \textbf{Writing:} \textit{how is new information written into memory?}
    \vspace{0.6em}
    \item[\ding{183}] \textbf{Reading:} \textit{how is relevant historical evidence read for the current step?}
    \vspace{0.6em}
    \item[\ding{184}] \textbf{Updating:} \textit{how is persistent state updated as the generated world evolves?}
    \vspace{0.6em}
    \item[\ding{185}] \textbf{Management:} \textit{how is bounded memory capacity managed?}
    \vspace{0.6em}
    \item[\ding{186}] \textbf{Integration:} \textit{how is retrieved memory integrated into the generator?}
\end{itemize}
}
Together, these operations form the memory lifecycle introduced in Section~\S\ref{sec:2.4} and illustrated in Figure~\ref{fig:figure_operation}. Before generating the next visual unit, relevant history is \textit{read} and \textit{integrated} with the current context; after generation, new evidence is \textit{written}, used to \textit{update} the persistent state, and \textit{managed} under resource and validity constraints. This operational view complements the carrier taxonomy in Section~\S\ref{sec:representation} and the preservation functions in Section~\S\ref{sec:functions}. The same visual buffer, KV cache, recurrent state, or explicit world state may implement different operational policies, while similar policies can be instantiated across different carriers and serve different preservation responsibilities. Operations therefore provide a common basis for comparing how AR video generators admit, access, revise, maintain, and ultimately use historical information during sequential rollout. Tables~\ref{tab:operations} and \ref{tab:operations_b} summarize representative methods from this perspective.

\subsection{Memory Writing}
\label{sec:5.1_writing}

Memory writing governs how newly synthesized or observed evidence is admitted into the persistent state $\mathbf{M}_n$. Evidence that is not committed may become inaccessible once it leaves the bounded local context, making writing the primary admission gate of the memory lifecycle. Formally, given the current visual unit $\mathbf{y}_n$, bounded context $\mathbf{C}_n$, and generalized condition $\mathbf{c}$, a writing operator extracts a memory candidate $\mathbf{w}_n$:
\begin{equation}
\mathbf{w}_n = \mathcal{W}_\alpha(\mathbf{y}_n, \mathbf{C}_n, \mathbf{c}),
\end{equation}
where $\mathcal{W}_\alpha$ may be parameterized or heuristic. The resulting $\mathbf{w}_n$ may comprise raw frames, latent tokens, KV activations, entity states, event records, or geometric fragments. Writing therefore involves two related decisions: \emph{admission}, which determines whether and which evidence should persist, and \emph{binding}, which determines where admitted evidence is stored when the memory is addressable.

\vspace{-1.2em}
\paragraph{Exhaustive Admission.}
The simplest policy admits every newly generated unit or corresponding internal activation before subsequent memory management \citep{yu2025context, zhang2026frame, zhang2025tinyhistory}. For example, FIFO-Diffusion~\citep{kim2024fifo} appends generated latent frames to a FIFO queue, while dense frame or KV buffers similarly defer selection until later truncation or compression. Exhaustive admission minimizes write-time omission and preserves local evidence at high fidelity, but also stores redundant observations and self-generated artifacts indiscriminately. Its effective cost therefore grows with the amount of retained history unless later management imposes a bounded window or compression policy. In this sense, exhaustive writing does not eliminate selection; it shifts selection from admission to downstream memory management.

\vspace{-1.2em}
\paragraph{Anchor-Based Admission.}
Rather than admitting the full rollout history, anchor-based policies preserve a sparse set of reference states expected to remain useful over longer horizons \citep{lu2026reward, zhao2026relax, chen2026pyramid}. LongLive~\citep{yang2025longlive} and Anchor Forcing~\citep{yang2026anchorforcing}, for example, maintain persistent sink or anchor KV states alongside rolling local history. More recent designs make anchor admission itself dynamic: FreqForcing~\citep{li2026freqforcing} periodically commits generated anchors to a persistent bank that remains available after the corresponding local context has advanced. Such anchors provide stable long-range references at low storage cost, but can become stale when appearance, viewpoint, or world state legitimately changes. Their effectiveness therefore depends on complementary updating or management mechanisms that refresh, reweight, or invalidate obsolete anchors.

\vspace{-1.2em}
\paragraph{Utility-Based Selection.}
A more adaptive policy admits evidence according to its estimated future relevance \citep{EgoLCD, xu2026sparse, wu2026video}. The writing criterion $\mathcal{W}_\alpha$ may depend on saliency, motion, attention, novelty, uncertainty, reconstruction error, or coverage of underrepresented entities and regions. Grounded Forcing~\citep{chen2026groundedforcing}, for instance, maintains a diversity-updated global anchor bank, while RECAP-Forcing~\citep{xu2026recap} performs finer-grained admission by scoring appearance novelty and retaining informative historical patches in a bounded long-term bank. Such policies allocate capacity more selectively than exhaustive writing, but future utility is only partially observable at admission time. Evidence that appears redundant locally, \textit{e.g.}, a peripheral object, background structure, or briefly observed landmark, may become essential only after later revisitation or interaction, making premature omission difficult to recover from.

\vspace{-1.2em}
\paragraph{Transition-Triggered Writing.}
In interactive or condition-varying generation, persistence may be triggered by changes in the generated world or its governing conditions rather than by continuous importance scoring \citep{yang2026anchorforcing, CoTriSyGen}. Relevant triggers include action onsets, object contacts, entity-state changes, scene boundaries, and semantic or prompt transitions. ActWorld~\citep{xiong2026actworld}, for example, couples persistent object-identity states with interaction-dependent event updates so that consequential changes survive beyond the frames in which they occur. Semantic transitions provide a complementary case: SWIFT~\citep{tan2026swift} augments history around prompt changes with semantic memory and retains segment-level anchors for subsequent generation. Transition-triggered writing is particularly useful when sparse changes have long-lived consequences, but it depends critically on reliable trigger detection and association. Missed or delayed triggers can leave memory incomplete, whereas incorrect action, \textit{e.g.}, entity or event, state binding can persistently attach a valid change to the wrong referent.

\vspace{-1.2em}
\paragraph{Structured Binding.}
For addressable memories, admission alone is insufficient: newly written evidence must also be assigned to the appropriate entity, location, map element, or event state \citep{vandersanden2026vid, VideoMemory, wu2026video}. We denote such addressed writing as:
\begin{equation}
\mathbf{w}_{n}[k] \leftarrow
\mathcal{W}_\alpha(\mathbf{y}_n, \mathbf{C}_n, \mathbf{c}; k),
\end{equation}
where $k$ may denote an entity identifier, spatial coordinate, topological node, map element, or transition index. Different carriers instantiate this principle through different keys. ActionParty \citep{pondaven2026actionparty} binds evolving states to persistent subject identities, Memory Forcing~\citep{huang2025memory} associates observations with an updated global point map, and ReWorld~\citep{chen2026reworld} organizes long-range memory through pose-indexed landmarks. Structured binding makes retained evidence explicitly queryable, but transfers substantial responsibility to the indexing mechanism: identity switches, geometric projection errors, or event-association failures can make correctly observed evidence persist under an incorrect address and contaminate subsequent reads and updates.

Across these policies, memory writing must balance three risks: \emph{omission}, when future-relevant evidence is never retained; \emph{contamination}, when redundant or unreliable generated evidence is admitted; and \emph{misbinding}, when valid evidence is stored under an incorrect address. Exhaustive admission favors coverage but increases resource and contamination costs, whereas selective policies improve efficiency at the risk of irreversible omission; structured writing improves accessibility but depends on reliable correspondence. Because subsequent reading, updating, management, and integration operate on the evidence established at write time, admission and binding errors can propagate throughout the remaining memory lifecycle.

\begin{table*}[!t]
\vspace{-1em}
\renewcommand{\arraystretch}{0.9}
\caption{Operation-centric taxonomy of selected memory mechanisms (\textbf{Part~I}: Visual Memory \& Implicit State Memory). Operation codes follow the mechanism headings in Sections~\S\ref{sec:5.1_writing}--\S\ref{sec:5.5_integration}; for example, EA, SI, and CRR denote Exhaustive Admission, Spatial Indexing, and Cache Refresh and Remapping, respectively. \textbf{Schedule:} each non-empty cell begins with a cadence icon, indicating that the memory operation is performed at every
\protect\cadenceframe~Frame,
\protect\cadenceblock~Block,
\protect\cadencechunk~Chunk, or
\protect\cadenceshot~Shot.
When present, a second icon specifies a trigger:
\protect\conditionquery~Query,
\protect\conditionevent~Event,
\protect\conditionprompt~Prompt, or
\protect\conditionadaptive~Adaptive.
\textbf{Policy:} \ruleicon~= rule-based, \learnedicon~= learned, and \hybridicon = hybrid.}
\label{tab:operations}
\vspace{-0.6em}
\centering
\scriptsize
\setlength{\tabcolsep}{3.2pt}
\begin{tabular}{
  >{\raggedright\arraybackslash}p{3.2cm}|
  >{\raggedright\arraybackslash}p{3.6cm}
  ccccccc
}
\hlineB{2.5}
\rowcolor{CadetBlue!20}
\textbf{Method} & \textbf{Unit} & \textbf{Write} & \textbf{Read} & \textbf{Update} & \textbf{Manage} & \textbf{Integ.} & \textbf{Schedule} & \textbf{Policy} \\
\hlineB{1.5}
\multicolumn{9}{c}{\cellcolor{gray!10}\textit{I. Visual Memory}} \\
\hline
WorldPack~\citep{oshima2025worldpack} & VAE-frame tokens & \opwrite{EA} & \opread{SI} & \opupdate{CRR} & \opmanage{Comp} & \opinject{AC} & \cadenceframe & \ruleicon \\
StreamingT2V~\citep{henschel2025streamingt2v} & Chunk features & \opwrite{EA}\,\opwrite{AA} & \opread{TR} & \opupdate{CRR} & \opmanage{AMR}\,\opmanage{RT} & \opinject{AC}\,\opinject{AMG} & \cadencechunk & \ruleicon \\
Context-as-Memory~\citep{yu2025context} & RGB frames & \opwrite{EA} & \opread{SI} & \opupdate{AE} & --- & \opinject{CC} & \scheduleicons{\cadenceframe}{\conditionquery} & \ruleicon \\
VMem~\citep{li2025vmem} & Surfel-indexed views & \opwrite{SB} & \opread{SI} & \opupdate{AE}\,\opupdate{SSU} & \opmanage{UAR} & \opinject{CC} & \scheduleicons{\cadencechunk}{\conditionquery} & \ruleicon \\
WorldMem~\citep{xiao2025worldmem} & Visual tokens & \opwrite{EA} & \opread{SI}\,\opread{CR} & \opupdate{AE} & --- & \opinject{AC} & \cadenceframe & \ruleicon \\
FramePack~\citep{zhang2026frame} & VAE-frame history & \opwrite{EA} & \opread{TR}\,\opread{CR} & \opupdate{CRR} & \opmanage{Comp} & \opinject{CC} & \cadenceframe & \ruleicon \\
MagicWorld~\citep{li2025magicworld} & Latent frames & \opwrite{EA} & \opread{CR} & \opupdate{CRR} & \opmanage{AMR}\,\opmanage{RT} & \opinject{CC} & \cadencechunk & \ruleicon \\
WorldPlay~\citep{sun2025worldplay} & VAE video blocks & \opwrite{EA} & \opread{TR}\,\opread{SI} & \opupdate{AE} & --- & \opinject{KVI} & \cadencechunk & \ruleicon \\
Matrix-Game 3.0~\citep{wang2026matrix} & VAE-frame latents & \opwrite{EA} & \opread{SI} & \opupdate{AE} & --- & \opinject{KVI} & \cadencechunk & \ruleicon \\
Pathwise-TTC~\citep{Pathwise-TTC} & First-frame anchor & \opwrite{AA} & \opread{TR} & --- & \opmanage{AMR} & \opinject{CC} & \cadencechunk & \ruleicon \\
MosaicMem~\citep{yu2026mosaicmem} & Latent patches & \opwrite{SB} & \opread{SI} & \opupdate{AE}\,\opupdate{SSU} & --- & \opinject{GAC}\,\opinject{AC} & \cadencechunk & \ruleicon \\

VRAG~\citep{chen2025vrag} & Latent-frame buffer & \opwrite{EA} & \opread{StR} & \opupdate{CRR} & \opmanage{RT} & \opinject{CC}\,\opinject{SSC} & \cadenceframe & \ruleicon \\
WorldCam~\citep{nam2026worldcam} & Pose-indexed latents & \opwrite{EA}\,\opwrite{SB} & \opread{SI} & \opupdate{AE}\,\opupdate{CRR} & \opmanage{AMR}\,\opmanage{RT} & \opinject{CC} & \cadenceblock & \ruleicon \\

FIFO-Diffusion~\citep{kim2024fifo} & Latent-frame queue & \opwrite{EA} & \opread{TR} & \opupdate{CRR} & \opmanage{RT} & \opinject{CC} & \cadenceframe & \ruleicon \\

MemLearner~\citep{yu2026memlearner} & Latent frames & \opwrite{EA} & \opread{AR} & \opupdate{AE} & --- & \opinject{AC} & \cadencechunk & \learnedicon \\

HyDRA~\citep{chen2026out} & VAE memory latents & \opwrite{EA} & \opread{CR}\,\opread{AR} & \opupdate{AE} & --- & \opinject{AC} & \cadenceblock & \hybridicon \\
LongLive-RAG~\citep{hu2026longlive} & Latent blocks & \opwrite{EA} & \opread{CR} & \opupdate{AE}\,\opupdate{CRR} & \opmanage{AMR}\,\opmanage{RT} & \opinject{KVI} & \cadenceblock & \hybridicon \\
MemCam~\citep{gao2026memcam} & Frame features & \opwrite{EA} & \opread{SI} & \opupdate{AE} & \opmanage{Comp} & \opinject{CC} & \cadencechunk & \hybridicon \\
I3DM~\citep{li2026i3dm} & RGB frames & \opwrite{EA} & \opread{SI} & \opupdate{AE} & --- & \opinject{GAC} & \cadencechunk & \hybridicon \\

\hlineB{1.5}
\multicolumn{9}{c}{\cellcolor{gray!10}\textit{II. Implicit State Memory}} \\
\hline

EgoLCD~\citep{EgoLCD} & Global/local KV & \opwrite{US} & \opread{CR}\,\opread{TR} & \opupdate{CRR} & \opmanage{RT}\,\opmanage{UAR} & \opinject{KVI} & \scheduleicons{\cadencechunk}{\conditionquery} & \ruleicon \\

Closing the Loop~\citep{Closing-the-Loop} & Clean KV chunks & \opwrite{EA} & \opread{SI} & \opupdate{CRR} & \opmanage{RT} & \opinject{KVI} & \scheduleicons{\cadencechunk}{\conditionevent} & \ruleicon \\
Future Forcing~\citep{Future-Forcing} & Historical KV & \opwrite{EA} & \opread{AR} & \opupdate{CRR} & \opmanage{UAR} & \opinject{KVI} & \scheduleicons{\cadenceframe}{\conditionadaptive} & \ruleicon \\

Relax Forcing~\citep{zhao2026relax} & Sink/history/tail KV & \opwrite{EA}\,\opwrite{AA} & \opread{CR}\,\opread{TR} & \opupdate{CRR} & \opmanage{AMR}\,\opmanage{RT}\,\opmanage{UAR} & \opinject{KVI} & \cadencechunk & \ruleicon \\
Anchor Forcing~\citep{yang2026anchorforcing} & Tri-region KV & \opwrite{EA}\,\opwrite{AA}\,\opwrite{TW} & \opread{TR} & \opupdate{CRR}\,\opupdate{Rev} & \opmanage{AMR}\,\opmanage{RT} & \opinject{KVI} & \scheduleicons{\cadenceframe}{\conditionprompt} & \ruleicon \\
LongLive~\citep{yang2025longlive} & Sink/local KV & \opwrite{EA}\,\opwrite{AA} & \opread{TR} & \opupdate{CRR}\,\opupdate{Rev} & \opmanage{AMR}\,\opmanage{RT} & \opinject{KVI} & \scheduleicons{\cadenceframe}{\conditionprompt} & \ruleicon \\
Rolling Forcing~\citep{liu2026rolling} & Sink/rolling KV & \opwrite{EA}\,\opwrite{AA} & \opread{TR} & \opupdate{CRR} & \opmanage{AMR}\,\opmanage{RT} & \opinject{KVI} & \cadenceframe & \ruleicon \\
RELIC~\citep{hong2025relic} & Recent/remote KV & \opwrite{EA} & \opread{TR}\,\opread{AR} & \opupdate{CRR} & \opmanage{Comp} & \opinject{KVI} & \cadenceblock & \ruleicon \\
Context Forcing~\citep{chen2026contextforcing} & Sink/slow/fast KV & \opwrite{EA}\,\opwrite{AA}\,\opwrite{US} & \opread{TR} & \opupdate{CRR} & \opmanage{AMR}\,\opmanage{RT}\,\opmanage{Comp} & \opinject{KVI} & \scheduleicons{\cadencechunk}{\conditionadaptive} & \ruleicon \\
FadeMem~\citep{lu2026fademem} & Consolidated KV & \opwrite{EA}\,\opwrite{AA} & \opread{TR}\,\opread{AR} & \opupdate{CRR} & \opmanage{AMR}\,\opmanage{Comp} & \opinject{KVI} & \scheduleicons{\cadenceblock}{\conditionevent} & \ruleicon \\

Deep Forcing~\citep{yi2025deep} & Sink/recent KV & \opwrite{EA}\,\opwrite{AA} & \opread{TR}\,\opread{AR} & \opupdate{CRR} & \opmanage{AMR}\,\opmanage{RT}\,\opmanage{UAR} & \opinject{KVI} & \scheduleicons{\cadenceframe}{\conditionadaptive} & \ruleicon \\
Head Forcing~\citep{tian2026head} & Head-wise KV & \opwrite{EA}\,\opwrite{AA} & \opread{TR}\,\opread{AR} & \opupdate{CRR} & \opmanage{AMR}\,\opmanage{RT}\,\opmanage{UAR}\,\opmanage{Comp} & \opinject{KVI} & \scheduleicons{\cadenceblock}{\conditionadaptive} & \ruleicon \\

LingBot-World~\citep{team2026advancing} & Block-causal history KV & \opwrite{EA} & \opread{TR}\,\opread{AR} & \opupdate{AE} & --- & \opinject{KVI} & \cadenceblock & \ruleicon \\
SGF~\citep{zhuang2026self} & Historical context KV & \opwrite{EA}\,\opwrite{AA} & \opread{TR}\,\opread{AR} & \opupdate{AE} & \opmanage{AMR}\,\opmanage{RT} & \opinject{KVI} & \cadenceframe & \ruleicon \\
Causal Forcing~\citep{zhu2026causal} & Causal history KV & \opwrite{EA} & \opread{TR}\,\opread{AR} & \opupdate{AE} & --- & \opinject{KVI} & \cadencechunk & \ruleicon \\
MAGI-1~\citep{teng2025magi} & Causal video chunks & \opwrite{EA} & \opread{TR}\,\opread{AR} & \opupdate{AE} & --- & \opinject{KVI} & \cadencechunk & \ruleicon \\

SANA-WM~\citep{SANA-WM} & Hybrid linear-attention state & \opwrite{EA} & \opread{TR}\,\opread{AR} & \opupdate{RC} & \opmanage{RT}\,\opmanage{Comp} & \opinject{AMG} & \cadenceframe & \learnedicon \\
MALT~\citep{yu2025malt} & Hidden states & \opwrite{EA} & \opread{AR} & \opupdate{RC} & \opmanage{Comp} & \opinject{AC} & \cadencechunk & \learnedicon \\
MAG~\citep{zhu2025memorize} & Block-summary KV & \opwrite{EA} & \opread{AR} & \opupdate{AE} & \opmanage{Comp} & \opinject{KVI} & \cadenceblock & \learnedicon \\
Sparse Forcing~\citep{xu2026sparse} & Persistent/local KV & \opwrite{US} & \opread{TR}\,\opread{AR} & \opupdate{CRR} & \opmanage{AMR}\,\opmanage{RT}\,\opmanage{UAR} & \opinject{KVI} & \scheduleicons{\cadenceblock}{\conditionadaptive} & \learnedicon \\
PaFu-KV~\citep{chen2026past} & Salience-scored KV & \opwrite{EA} & \opread{AR} & \opupdate{CRR} & \opmanage{UAR} & \opinject{KVI} & \scheduleicons{\cadencechunk}{\conditionadaptive} & \learnedicon \\

VideoSSM~\citep{yu2025videossm} & Local KV + SSM state & \opwrite{EA}\,\opwrite{AA} & \opread{TR}\,\opread{AR} & \opupdate{RC}\,\opupdate{CRR} & \opmanage{AMR}\,\opmanage{RT}\,\opmanage{Comp} & \opinject{KVI}\,\opinject{AMG} & \scheduleicons{\cadencechunk}{\conditionprompt} & \hybridicon \\
VideoMLA~\citep{yesiltepe2026videomla} & Low-rank latent KV & \opwrite{EA} & \opread{TR}\,\opread{AR} & \opupdate{AE} & \opmanage{RT}\,\opmanage{Comp} & \opinject{KVI} & \cadencechunk & \hybridicon \\
OmniMem~\citep{OmniMem} & Multi-branch KV & \opwrite{EA} & \opread{CR}\,\opread{TR} & \opupdate{CRR} & \opmanage{RT}\,\opmanage{UAR}\,\opmanage{Comp} & \opinject{AMG} & \cadencechunk & \hybridicon \\
PackForcing~\citep{mao2026packforcing} & Three-tier KV & \opwrite{EA} & \opread{CR}\,\opread{TR} & \opupdate{CRR} & \opmanage{AMR}\,\opmanage{RT}\,\opmanage{Comp} & \opinject{KVI} & \cadenceblock & \hybridicon \\
GIM-World~\citep{wei2026geometry} & Pose-indexed VAE history & \opwrite{EA} & \opread{AR} & \opupdate{AE}\,\opupdate{CRR} & \opmanage{UAR}\,\opmanage{Comp} & \opinject{CC} & \cadenceframe & \hybridicon \\

\hline
\multicolumn{9}{r}{\textit{Continued on next page.}} \\
\hlineB{2.5}
\end{tabular}
\vspace{-2.4em}
\end{table*}

\begin{table*}[!t]
\vspace{-1em}
\renewcommand{\arraystretch}{0.9}
\caption{Operation-centric taxonomy of selected memory mechanisms (\textbf{Part~II}: Implicit State, Explicit State and Adaptive Parametric Memory). Column conventions and abbreviations follow Table~\ref{tab:operations}.}
\label{tab:operations_b}
\vspace{-0.6em}
\centering
\scriptsize
\setlength{\tabcolsep}{2.4pt}
\begin{tabular}{
  >{\raggedright\arraybackslash}p{3.2cm}|
  >{\raggedright\arraybackslash}p{3.6cm}
  ccccccc
}
\hlineB{2.5}
\rowcolor{CadetBlue!20}
\textbf{Method} & \textbf{Unit} & \textbf{Write} & \textbf{Read} & \textbf{Update} & \textbf{Manage} & \textbf{Integ.} & \textbf{Schedule} & \textbf{Policy} \\
\hlineB{1.5}
\multicolumn{9}{c}{\cellcolor{gray!10}\textit{II. Implicit State Memory}} \\
\hline

MemRoPE~\citep{kim2026memrope} & EMA tokens + KV & \opwrite{EA}\,\opwrite{AA} & \opread{TR}\,\opread{AR} & \opupdate{RC}\,\opupdate{CRR} & \opmanage{AMR}\,\opmanage{RT}\,\opmanage{Comp} & \opinject{KVI} & \cadencechunk & \ruleicon \\

Pyramid Forcing~\citep{chen2026pyramid} & Head-wise KV & \opwrite{EA}\,\opwrite{AA} & \opread{TR}\,\opread{AR} & \opupdate{CRR} & \opmanage{AMR}\,\opmanage{RT}\,\opmanage{Comp} & \opinject{KVI} & \scheduleicons{\cadenceframe}{\conditionadaptive} & \ruleicon \\
Infinity-RoPE~\citep{yesiltepe2025infinityrope} & Sink/rolling KV & \opwrite{EA}\,\opwrite{AA} & \opread{TR}\,\opread{AR} & \opupdate{CRR} & \opmanage{AMR}\,\opmanage{RT} & \opinject{KVI} & \scheduleicons{\cadenceblock}{\conditionprompt} & \ruleicon \\
Reward Forcing~\citep{lu2026reward} & EMA sink + local KV & \opwrite{EA}\,\opwrite{AA} & \opread{TR}\,\opread{AR} & \opupdate{RC} & \opmanage{AMR}\,\opmanage{RT}\,\opmanage{Comp} & \opinject{KVI} & \cadencechunk & \ruleicon \\
Rolling Sink~\citep{li2026rolling} & Sink/rolling KV & \opwrite{EA}\,\opwrite{AA} & \opread{TR}\,\opread{AR} & \opupdate{CRR} & \opmanage{AMR}\,\opmanage{RT} & \opinject{KVI} & \cadenceblock & \ruleicon \\

Self Forcing~\citep{huang2026self} & Rolling KV & \opwrite{EA} & \opread{TR}\,\opread{AR} & \opupdate{AE} & \opmanage{RT} & \opinject{KVI} & \cadenceframe & \ruleicon \\
Self-Forcing++~\citep{cui2026self} & Rolling KV & \opwrite{EA} & \opread{TR}\,\opread{AR} & \opupdate{AE} & \opmanage{RT} & \opinject{KVI} & \cadencechunk & \ruleicon \\
INSPATIO-WORLD~\citep{inspatio2026world} & Anchor/recent KV & \opwrite{EA}\,\opwrite{AA} & \opread{TR}\,\opread{AR} & \opupdate{AE} & \opmanage{AMR}\,\opmanage{RT} & \opinject{KVI} & \cadencechunk & \ruleicon \\
WorldCraft~\citep{gu2026worldcraft} & KV + trajectory state & \opwrite{EA}\,\opwrite{AA} & \opread{SI}\,\opread{AR} & \opupdate{CRR}\,\opupdate{Rev} & \opmanage{VAF} & \opinject{KVI}\,\opinject{GAC} & \scheduleicons{\cadencechunk}{\conditionevent} & \ruleicon \\

Echo-Infinity~\citep{bian2026echo} & Memory queries + KV & \opwrite{EA}\,\opwrite{AA} & \opread{TR}\,\opread{AR} & \opupdate{RC}\,\opupdate{CRR} & \opmanage{AMR}\,\opmanage{RT}\,\opmanage{Comp} & \opinject{KVI} & \scheduleicons{\cadencechunk}{\conditionadaptive} & \learnedicon \\
TinyHistory~\citep{zhang2025tinyhistory} & History embeddings & \opwrite{EA} & \opread{TR} & \opupdate{AE} & \opmanage{Comp} & \opinject{CC} & \cadencechunk & \learnedicon \\
Long-Context SSM~\citep{po2025long} & SSM state & \opwrite{EA} & \opread{TR}\,\opread{AR} & \opupdate{RC} & --- & \opinject{AMG} & \cadenceframe & \learnedicon \\
StateSpaceDiffuser~\citep{savov2025statespacediffuser} & Mamba state & \opwrite{EA} & \opread{StR} & \opupdate{RC} & --- & \opinject{CC} & \cadenceframe & \learnedicon \\
RAD~\citep{chen2025rad} & Recurrent states & \opwrite{EA} & \opread{TR}\,\opread{StR} & \opupdate{RC} & --- & \opinject{AMG} & \cadenceframe & \learnedicon \\
MemoryPack~\citep{wu2025memorypack} & SemanticPack state & \opwrite{EA} & \opread{TR}\,\opread{AR} & \opupdate{RC} & \opmanage{Comp} & \opinject{AC} & \cadencechunk & \learnedicon \\
ARL$^2$~\citep{li2026attend} & Linear-attn. state & \opwrite{EA} & \opread{TR}\,\opread{AR} & \opupdate{RC} & --- & \opinject{AMG} & \cadenceframe & \learnedicon \\

Hybrid Forcing~\citep{li2026hybridforcing} & Linear-attn. state & \opwrite{EA} & \opread{TR}\,\opread{AR} & \opupdate{RC} & \opmanage{RT}\,\opmanage{Comp} & \opinject{AMG} & \cadencechunk & \hybridicon \\
Infinite-World~\citep{wu2026infinite} & History tokens & \opwrite{EA} & \opread{TR} & \opupdate{RC} & \opmanage{Comp} & \opinject{CC} & \scheduleicons{\cadencechunk}{\conditionadaptive} & \hybridicon \\
ViewRoPE~\citep{xiang2026geometry} & Geometry-rotated KV & \opwrite{EA} & \opread{SI}\,\opread{AR} & \opupdate{AE} & --- & \opinject{KVI} & \cadenceframe & \hybridicon \\
\hline
\multicolumn{9}{c}{\cellcolor{gray!10}\textit{III. Explicit State Memory}} \\
\hline
Voyager~\citep{huang2025voyager} & RGB-D point cache & \opwrite{EA}\,\opwrite{SB} & \opread{SI} & \opupdate{AE} & \opmanage{UAR} & \opinject{CC}\,\opinject{GAC} & \scheduleicons{\cadenceframe}{\conditionadaptive} & \ruleicon \\
DeepVerse~\citep{chen2025deepverse} & RGB-D-ray states & \opwrite{EA}\,\opwrite{SB} & \opread{TR}\,\opread{SI} & \opupdate{AE}\,\opupdate{CRR} & \opmanage{RT} & \opinject{CC} & \cadencechunk & \ruleicon \\
Lyra 2.0~\citep{shen2026lyra} & Historical frames + per-frame 3D cache & \opwrite{EA}\,\opwrite{SB} & \opread{TR}\,\opread{SI} & \opupdate{AE} & \opmanage{Comp} & \opinject{CC}\,\opinject{GAC} & \scheduleicons{\cadencechunk}{\conditionquery} & \ruleicon \\

PERSIST~\citep{garcin2026beyond} & Latent 3D world-frame & \opwrite{SB} & \opread{SI} & \opupdate{WST} & --- & \opinject{CC}\,\opinject{GAC} & \cadenceframe & \learnedicon \\
SlotMem~\citep{SlotMem} & Role-wise character slots & \opwrite{US}\,\opwrite{SB} & \opread{SA} & \opupdate{SSU} & \opmanage{Comp} & \opinject{AC} & \scheduleicons{\cadencechunk}{\conditionprompt} & \learnedicon \\

Memento~\citep{Memento} & Reconstructed subject state & \opwrite{US} & \opread{CR}\,\opread{AR} & \opupdate{CRR} & \opmanage{UAR} & \opinject{CC} & \cadenceshot & \learnedicon \\

SPMEM~\citep{wu2026video} & Map + frame tiers & \opwrite{US}\,\opwrite{SB} & \opread{TR}\,\opread{SI} & \opupdate{AE}\,\opupdate{SSU} & \opmanage{RT} & \opinject{CC}\,\opinject{AC}\,\opinject{GAC} & \scheduleicons{\cadencechunk}{\conditionadaptive} & \hybridicon \\
ActionParty~\citep{pondaven2026actionparty} & Subject states & \opwrite{SB} & \opread{SA} & \opupdate{WST} & \opmanage{RT} & \opinject{AC}\,\opinject{SSC} & \cadenceframe & \hybridicon \\
A$^2$RD~\citep{long20262} & Typed records + media & \opwrite{EA}\,\opwrite{SB} & \opread{CR}\,\opread{SA} & \opupdate{AE} & --- & \opinject{CC}\,\opinject{SSC} & \scheduleicons{\cadencechunk}{\conditionadaptive} & \hybridicon \\
LiveWorld~\citep{duan2026liveworld} & Point-cloud states & \opwrite{US}\,\opwrite{SB} & \opread{TR}\,\opread{SI} & \opupdate{SSU}\,\opupdate{WST} & \opmanage{UAR} & \opinject{AC}\,\opinject{GAC} & \scheduleicons{\cadencechunk}{\conditionevent} & \hybridicon \\
Memory Forcing~\citep{huang2025memory} & Point map + frames & \opwrite{US}\,\opwrite{SB} & \opread{TR}\,\opread{SI} & \opupdate{SSU} & \opmanage{UAR} & \opinject{AC} & \scheduleicons{\cadenceblock}{\conditionadaptive} & \hybridicon \\
EvoWorld~\citep{wang2025evoworld} & Colored point cloud & \opwrite{SB} & \opread{SI} & \opupdate{SSU} & \opmanage{UAR} & \opinject{AC}\,\opinject{GAC} & \cadencechunk & \hybridicon \\
WorldStereo~\citep{WorldStereo} & 2D bank + 3D cache & \opwrite{SB} & \opread{SI} & \opupdate{AE}\,\opupdate{SSU} & \opmanage{Comp} & \opinject{AC}\,\opinject{GAC} & \cadencechunk & \hybridicon \\
Spatia~\citep{zhao2026spatia} & Point map + frames & \opwrite{SB} & \opread{SI} & \opupdate{SSU} & --- & \opinject{CC}\,\opinject{GAC} & \cadencechunk & \hybridicon \\
LSM-World~\citep{wang2026latent} & World-space latents & \opwrite{EA}\,\opwrite{SB} & \opread{SI} & \opupdate{AE} & --- & \opinject{GAC} & \cadencechunk & \hybridicon \\
AnchorWeave~\citep{wang2026anchorweave} & Local point clouds & \opwrite{EA}\,\opwrite{SB} & \opread{SI} & \opupdate{AE} & --- & \opinject{AMG}\,\opinject{GAC} & \cadencechunk & \hybridicon \\
MemoryWorld~\citep{zhou2026learning} & DINO voxel map & \opwrite{SB} & \opread{SI}\,\opread{AR} & \opupdate{SSU} & \opmanage{Comp} & \opinject{AC}\,\opinject{GAC} & \cadencechunk & \hybridicon \\
EM-Vid~\citep{vandersanden2026vid} & Entity-indexed latent patches & \opwrite{US}\,\opwrite{SB} & \opread{SA} & \opupdate{AE} & \opmanage{UAR} & \opinject{CC}\,\opinject{SSC} & \scheduleicons{\cadenceshot}{\conditionprompt} & \hybridicon \\
IAMFlow~\citep{NarraStream-Bench} & Entity ID/attribute registry & \opwrite{US}\,\opwrite{SB} & \opread{SA} & \opupdate{AE}\,\opupdate{Rev} & \opmanage{UAR} & \opinject{KVI}\,\opinject{SSC} & \scheduleicons{\cadencechunk}{\conditionprompt} & \hybridicon \\
InfinityStory~\citep{elmoghany2026infinitystory} & Character/transition state & \opwrite{SB} & --- & \opupdate{SSU} & --- & \opinject{SSC} & \scheduleicons{\cadenceshot}{\conditionevent} & \hybridicon \\

ActWorld~\citep{xiong2026actworld} & Event/object-token bank & \opwrite{US}\,\opwrite{SB}\,\opwrite{TW} & \opread{SA} & \opupdate{WST} & \opmanage{RT}\,\opmanage{UAR}\,\opmanage{Comp} & \opinject{CC}\,\opinject{AMG}\,\opinject{SSC} & \scheduleicons{\cadencechunk}{\conditionevent} & \hybridicon \\
MoVerse~\citep{zhou2026moverse} & Panoramic Gaussian scaffold & \opwrite{SB} & \opread{SI} & \opupdate{SSU} & --- & \opinject{GAC} & \cadenceblock & \hybridicon \\
AlayaWorld~\citep{team2026alayaworld} & Sink/history/spatial cache & \opwrite{EA}\,\opwrite{AA}\,\opwrite{SB} & \opread{TR}\,\opread{SI} & \opupdate{AE}\,\opupdate{CRR}\,\opupdate{SSU} & \opmanage{AMR}\,\opmanage{RT}\,\opmanage{Comp} & \opinject{CC}\,\opinject{GAC} & \cadencechunk & \hybridicon \\
VideoMemory~\citep{VideoMemory} & Entity/image banks & \opwrite{SB} & \opread{CR}\,\opread{SA} & \opupdate{AE} & --- & \opinject{CC}\,\opinject{SSC} & \scheduleicons{\cadenceshot}{\conditionprompt} & \hybridicon \\
CoTriSyGen~\citep{CoTriSyGen} & Entity/image records & \opwrite{TW}\,\opwrite{SB} & \opread{CR}\,\opread{SA} & \opupdate{AE}\,\opupdate{SSU} & --- & \opinject{CC}\,\opinject{SSC} & \scheduleicons{\cadenceshot}{\conditionevent} & \hybridicon \\

\hline
\multicolumn{9}{c}{\cellcolor{gray!10}\textit{IV. Adaptive Parametric Memory}} \\
\hline
ISPA~\citep{fu2026towards} & $\Delta W$ + residual KV & \opwrite{EA} & \opread{AR} & \opupdate{RC} & \opmanage{UAR} & \opinject{KVI}\,\opinject{AMG} & \scheduleicons{\cadenceframe}{\conditionevent} & \ruleicon \\
SlowFast-VGen~\citep{hong2025slowfast} & Temporal LoRA & \opwrite{EA} & --- & \opupdate{RC} & \opmanage{Comp} & \opinject{AMG} & \cadencechunk & \learnedicon \\
LaCT~\citep{zhang2026test} & Fast weights & \opwrite{EA} & \opread{AR} & \opupdate{RC} & \opmanage{Comp} & \opinject{AMG} & \cadencechunk & \learnedicon \\
HippoCampus~\citep{peng2026hippocampus} & Fast weights + hierarchical tokens & \opwrite{EA} & \opread{TR}\,\opread{AR} & \opupdate{RC}\,\opupdate{CRR} & \opmanage{RT}\,\opmanage{Comp} & \opinject{KVI}\,\opinject{AMG} & \cadencechunk & \hybridicon \\
\hlineB{2.5}
\end{tabular}
\vspace{-3em}
\end{table*}

\subsection{Memory Reading}
\label{sec:5.2_reading}

Memory reading governs how retained historical evidence is made accessible to the generator at a given rollout step. While writing determines what enters persistent memory, reading determines which retained information is selected and exposed for the current prediction. A memory entry that remains stored but cannot be located, is retrieved for the wrong query, or becomes difficult to access over long temporal gaps is functionally unavailable even without being physically removed. Formally, given a query state $\mathbf{q}_n$ derived from the current context, conditional inputs, action signals, or latent activations, the reading operation extracts a memory context $\mathbf{r}_n$:
\begin{equation}
    \mathbf{r}_n = \mathcal{R}_\beta(\mathbf{q}_n, \mathbf{M}_n),
\end{equation}
where $\mathcal{R}_\beta$ denotes a parameterized or heuristic retrieval, routing, attention, or projection operator. The resulting $\mathbf{r}_n$ may consist of RGB observations, latent, KV states, explicit records, or projected spatial states. The central challenge is therefore to recover historically relevant evidence with sufficient precision and computational efficiency, particularly when relevant information is temporally remote from the current step.

\vspace{-1.2em}
\paragraph{Temporal Reading.}
The simplest policy exposes memory according to temporal proximity. Sliding windows, recent frame concatenation, and rolling KV or latent contexts instantiate this strategy \citep{liu2026rolling, zhao2026relax, chen2026contextforcing}. FIFO-Diffusion~\citep{kim2024fifo}, for example, reads from a bounded queue of recent latent frames, while LongLive~\citep{yang2025longlive} combines a rolling recent KV context with persistent reference states. Temporal reading is efficient and well-suited to short-range motion and appearance continuity, but recency is an unreliable proxy for long-term relevance. A distant observation may contain the entity, layout, or event state needed for the current prediction, whereas nearby history may contain only occlusion or transient views. Purely temporal access therefore remains vulnerable when dependencies extend beyond the retained local horizon.

\vspace{-1.2em}
\paragraph{Content-Based Retrieval.}
To recover nonlocal evidence, many systems select memory according to visual, semantic, or latent similarity \citep{hu2026longlive, OmniMem, Memento}. The query $\mathbf{q}_n$ may be formed from the current latent, prompt representation, hidden feature, subject embedding, or scene descriptor and matched against stored entries through similarity scoring or learned retrieval. Recent methods make this selection increasingly query dependent. Resampling Forcing \citep{guo2025end} selects the most relevant historical frames for each current query, Focused Forcing~\citep{cai2026focused} performs content aware selection at the frame and attention head levels, and T-RFlow \citep{chang2026towards} retrieves a fixed budget of past clips according to cross-clip relevance. Such retrieval extends effective access beyond temporal neighborhoods, but similarity alone can be ambiguous. Representation drift or visually similar distractors may yield plausible evidence referring to the wrong entity, scene, or event.

\vspace{-1.2em}
\paragraph{Semantic Addressing.}
When memory is organized into semantically meaningful units, reading can use entity identifiers, roles, textual summaries, event keys, or relation records as access signals \citep{SlotMem, vandersanden2026vid, NarraStream-Bench}. VideoMemory \citep{VideoMemory}, for example, maintains a visual semantic memory bank associated with persistent entities, while SlotMemory~\citep{SlotMemory} organizes historical KV evidence around semantic objects. Such addressing allows the generator to retrieve information by \emph{what} or \emph{who} is relevant rather than temporal proximity alone, which is useful when entities or events reappear after prolonged absence. Its main failure mode is binding error. An incorrect entity, role, or event query can retrieve internally coherent evidence associated with the wrong referent, leading to identity swaps, relation errors, or inconsistent event continuation.

\vspace{-1.2em}
\paragraph{Spatial Indexing.}
For spatial memories, relevance can instead be determined through explicit geometric correspondence. The query may use camera pose, estimated depth, target view parameters, or topological coordinates to access co-visible observations, spatial map elements, point-based states, or pose-indexed memories from historical views \citep{kong2025causnvs, xu2026teaching}:
\begin{equation}
    \mathbf{r}_n = \mathcal{R}_\beta(\pi_n, \mathbf{M}_n),
\end{equation}
where $\pi_n$ denotes the current or target projection model and corresponds to a specialized query with $\mathbf{q}_n = \pi_n$. Light Interaction~\citep{lu2026light} uses camera-aware relevance to prune and reconstruct historical context before it is accessed by the current chunk. Spatial access can also operate directly over latent or neural memories. WorldKV~\citep{yi2026worldkv} uses camera and action correspondence to reactivate relevant historical KV blocks, while ReWorld~\citep{chen2026reworld} queries pose-indexed landmark memory. DreamX-World~\citep{team2026dreamx} retrieves nonlocal visual evidence through geometric correspondence, and Mem-World~\citep{zheng2026mem} associates manipulation history with evolving 4D surfels for retrieval under occlusion and camera motion. These mechanisms support revisitation without temporal proximity, but pose drift, depth error, calibration mismatch, or incorrect correspondence can retrieve spatially incompatible evidence and introduce persistent geometric inconsistency.

\vspace{-1.2em}
\paragraph{State-Conditioned Retrieval.}
In interactive or stateful world models, the relevant query may depend on the current estimated world state rather than on appearance or location alone. Memory can therefore be accessed according to entity status, interaction history, task state, or previously established consequences. ActWorld~\citep{xiong2026actworld}, for example, routes historical information according to interaction importance while maintaining persistent object and event states. Such state-conditioned access is useful when the required evidence concerns an updated object configuration or an event whose visual cause is no longer present in the local context. Its reliability depends on state synchronization. If the query is formed from a stale, incorrectly transitioned, or internally inconsistent world state, the retrieved evidence may be individually plausible while contradicting the accumulated rollout history.

\vspace{-1.2em}
\paragraph{Attentive Routing.}
Implicit memory architectures often expose retained neural states through attention rather than an explicitly specified retrieval key \citep{yu2026memlearner, Future-Forcing, xu2026sparse}. Given query features from the current generation step, the model reads stored KV entries, recurrent summaries, or encoded history as:
\begin{equation}
    \mathbf{r}_n =
    \mathrm{Attn}
    (\mathbf{q}_n,
    \mathbf{K}_{\mathbf{M}_n},
    \mathbf{V}_{\mathbf{M}_n}).
\end{equation}
This mechanism is differentiable and can support fine-grained, query-specific access without requiring predefined semantic or geometric addresses. SparSTAR~\citep{lee2026sparstar}, for example, recomputes relevance scores across scales, layers, and attention heads and reads only the highest scoring historical KV blocks. Long-lived attention states also expose a distinct accessibility problem: retained evidence may remain physically present but become difficult to attend to. Wonder~\citep{xu2026wonder} preserves sparse full-fidelity distant states alongside recent context, while WorldTrace~\citep{wu2026addressable} remaps remote states to temporal positions within the model's effective attention range. Attentive reading therefore depends not only on retaining useful states, but also on keeping them distinguishable and reachable under the current query distribution.

Memory reading ultimately balances relevance, precision, accessibility, and cost. Temporal access is efficient but limited by recency; content retrieval extends the accessible horizon but depends on discriminative queries; semantic and spatial addressing provide more structured access while inheriting binding or correspondence errors; state conditioned retrieval requires a synchronized estimate of the evolving world; and attentive routing offers flexible neural access but can suffer from dilution or positional mismatch. These distinctions also clarify an important operational boundary: reading determines which memory is exposed as $\mathbf{r}_n$, whereas how strongly and through which pathway that retrieved evidence affects generation is the responsibility of memory integration in Section~\S\ref{sec:5.5_integration}.

\subsection{Memory Updating}
\label{sec:5.3_updating}

Memory updating governs how persistent memory changes after new evidence has been admitted. Writing determines what enters memory, whereas updating determines how that evidence modifies, consolidates, or corrects the existing state. This distinction is particularly important in long rollouts: insufficient updating leaves memory stale, while excessive updating can overwrite valid long range commitments. Formally, given a write candidate $\mathbf{w}_n$, the update operator produces an intermediate state:
\begin{equation}
    \widetilde{\mathbf{M}}_{n+1}
    =
    \mathcal{U}_\phi
    (\mathbf{M}_n, \mathbf{w}_n, \mathbf{C}_n, \mathbf{c}),
\end{equation}
where $\mathcal{U}_\phi$ may implement append operations, recurrent transitions, cache transformations, structured state updates, world state transitions, or corrective revision. The intermediate state $\widetilde{\mathbf{M}}_{n+1}$ is subsequently processed by memory management to obtain $\mathbf{M}_{n+1}$. Updating therefore concerns how memory content evolves, while Section~\S\ref{sec:5.4_manage} concerns how that content is retained under resource and validity constraints. The central challenge is to balance stability and plasticity, preserving established information while remaining responsive to legitimate changes and detected errors.

\vspace{-1.2em}
\paragraph{Append-Only Extension.}
The simplest update extends memory with each newly admitted entry without modifying previously stored evidence \citep{yu2025context, wang2026matrix, xiao2025worldmem}. Frame banks, latent queues, and retrieval histories instantiate this policy. Append-only extension avoids destructive overwriting and preserves episodic evidence for later access, but does not reconcile conflicts across time. Obsolete observations, duplicated records, and self-generated artifacts can therefore coexist with valid evidence as the rollout progresses. Reading must then determine which version is relevant, while management must control the growth. Append-only updating thus favors historical coverage at the cost of increasing ambiguity and downstream maintenance.

\vspace{-1.2em}
\paragraph{Recurrent Consolidation.}
A different strategy repeatedly folds new evidence into a compact, bounded latent state that persists across rollout steps \citep{yu2025malt, bian2026echo, yu2025videossm}:
\begin{equation}
    \mathbf{s}_{n+1}
    =
    \mathcal{U}_\phi(\mathbf{s}_n,\mathbf{w}_n).
\end{equation}
Recurrent latents, state-space memories, and linear-attention summaries follow this principle, replacing an expanding history with an evolving state of fixed or slowly growing size. Recent methods instantiate the transition in different ways. AdaState~\citep{dalva2026adastate} replaces a static first-frame anchor with a hidden state that is recurrently denoised and updated with each generated chunk. Ripple~\citep{ding2026ripple} maintains modality-specific recurrent states that continuously summarize audio and visual context, with cross-modal interaction between them. Related fixed-capacity designs such as S2DiT~\citep{zhao2026s2dit} and Steady-Forcing~\citep{minar2026steady} consolidate long context into recurrent linear-attention or global states \citep{lee2025edeline}. Such updates scale favorably with rollout length, but consolidation sacrifices episodic separability. Rare evidence may be attenuated, and an erroneous transition can persist because later states inherit the compressed result rather than the original observations.

\vspace{-1.2em}
\paragraph{Cache Refresh and Remapping.}
For attention-based memory, updating may transform persistent KV states as the rollout advances \citep{yang2025longlive, zhao2026relax, yesiltepe2025infinityrope}. This includes refreshing local context, transferring information across chunk boundaries, rebuilding condition-dependent caches, and remapping retained states to remain compatible with the current positional structure. Knot Forcing~\citep{xiao2025knot}, for example, refreshes local history while propagating cross-chunk temporal knots, whereas UniSwap~\citep{zhang2026uniswap} reanchors and remaps retained KV states as the target context advances. Visko Orbis~\citep{gao2026visko} incrementally updates a bounded multi-scale state as generated chunks are committed, retaining recent content at finer granularity while progressively consolidating older history. These mechanisms preserve model-native evidence without explicit decoding, but cache states remain coupled to positional encodings and activation statistics. Incorrect refresh or remapping can therefore make retained evidence stale or incompatible even when its underlying content remains useful.

\vspace{-1.2em}
\paragraph{Structured State Update.}
Explicit memories update interpretable spatial, geometric, or entity states rather than only extending neural history \cite{wu2026video, wang2025evoworld, zhou2026learning, chen2026physstream}. New evidence may revise image-aligned perceptual fields, register observations into a common coordinate system, or modify identity-bound attributes and interaction states. WorldWeaver~\citep{liu2026worldweaver} maintains long-horizon memory using jointly modeled RGB, depth, and optical-flow information, exploiting perceptual signals that remain more stable than RGB alone. PERSIST~\citep{garcin2026beyond} maintains an editable latent 3D world state, while Mem-World~\citep{zheng2026mem} associates manipulation history with temporally evolving 4D surfels. Structured updates improve queryability and allow multiple observations to contribute to a persistent state, but their reliability depends on correspondence and estimation accuracy. Pose drift, depth error, identity mismatch, or incorrect registration can consolidate incompatible evidence and turn a local estimation error into a persistent memory error.

\vspace{-1.2em}
\paragraph{World-State Transition.}
In stateful and interactive generators, memory may also evolve because the represented world itself changes \citep{pondaven2026actionparty, duan2026liveworld, garcin2026beyond}. The update can depend on an action or control signal in addition to newly rendered evidence:
\begin{equation}
    \widetilde{\mathbf{M}}_{n+1}
    =
    \mathcal{U}_\phi
    \left(
        \mathbf{M}_n,
        \mathbf{w}_n,
        \mathbf{a}_n
    \right),
\end{equation}
where $\mathbf{a}_n$ denotes the action or control applied during the current transition. Such updates cover both autonomous state evolution and intervention-induced changes. WorldDirector~\citep{wang2026worlddirector} propagates persistent object-motion states even when the corresponding objects are not currently visible, while ActWorld~\citep{xiong2026actworld} couples persistent object states with interaction-dependent event updates. WorldCraft~\citep{gu2026worldcraft} similarly refreshes world memory after object manipulation so that later views reflect the modified configuration. Programmable World Model \citep{huang2026programmable} executes explicit transition rules over a persistent global state, including off-screen entities. When no new observation is available, $\mathbf{w}_n$ may be absent and the transition must rely on the previous state and learned dynamics or control. The principal risk is incorrect transition or association, which can preserve a coherent but wrong state throughout subsequent rollout.

\vspace{-1.2em}
\paragraph{Corrective Revision.}
Updating may also revise memory when newly generated evidence indicates that previously retained state has become unreliable \citep{yang2026anchorforcing, gu2026worldcraft, NarraStream-Bench}. Unlike ordinary accumulation, corrective revision explicitly evaluates inconsistency and changes the state to prevent detected errors from continuing into future conditioning. TokenTrim~\citep{shaulov2026tokentrim}, for example, estimates drift in generated latent tokens, removes unstable positions from the historical conditioning cache, and regenerates the affected batch before the accepted state is propagated to later steps. More generally, revision may replace stale attributes, repair inconsistent state estimates, or correct erroneous associations when sufficient evidence becomes available \citep{su2026cycle}. The difficulty lies in distinguishing genuine corruption from legitimate world change or stochastic variation. Over-correction can erase valid history, whereas weak correction allows generated errors to be repeatedly consolidated.

Across these update mechanisms, three failure regimes recur. \emph{Under-updating} preserves obsolete states after the world has changed, \emph{over-updating} erases historical commitments that should remain valid, and \emph{incorrect updating} consolidates noisy or misassociated evidence into persistent state. Append-only extension largely avoids overwriting but leaves conflicts unresolved; recurrent consolidation and cache refresh provide efficient evolving neural states but can propagate compressed errors; structured and world-state updates provide stronger organization but depend on reliable correspondence and transitions; corrective revision can limit error accumulation but requires trustworthy inconsistency detection. Memory updating is therefore the operation that turns retained history into an evolving state estimate, before memory management determines what portion of that state remains persistent.

\subsection{Memory Management}
\label{sec:5.4_manage}

Memory management governs which parts of the updated state remain persistent, and at what fidelity, under finite storage, computation, and attention budgets. After memory updating produces the intermediate state $\widetilde{\mathbf{M}}_{n+1}$, management determines how this state is retained, compressed, evicted, or invalidated before it conditions subsequent rollout:
\begin{equation}
    \mathbf{M}_{n+1}
    =
    \mathcal{G}_\gamma
    (\widetilde{\mathbf{M}}_{n+1}; B),
\end{equation}
where $\mathcal{G}_\gamma$ denotes a parameterized or heuristic management policy and $B$ represents the available resource budget. This separates management from updating: updating determines how new evidence changes the memory content, whereas management determines which resulting information survives and in what representation. The central challenge is that both future relevance and reliability are only partially observable when these decisions are made.

\vspace{-1.2em}
\paragraph{Rolling Truncation.}
The simplest policy bounds memory through a moving temporal window, retaining only recent frames, latents, or KV states \citep{li2026rolling, liu2026rolling, huang2026self}. FIFO-Diffusion~\citep{kim2024fifo}, for example, maintains a bounded queue of latent history, while many streaming systems similarly discard states once they leave the active retention window. Rolling truncation provides predictable cost and preserves recent dynamics, but treats age as a proxy for relevance. Remote evidence about identity, geometry, or prior state changes can therefore disappear even when it remains necessary for later generation. This policy controls memory growth effectively, but inherits the long-range dependency failures that motivate persistent memory.

\vspace{-1.2em}
\paragraph{Anchor and Multi-Tier Retention.}
A common alternative reserves persistent capacity for selected long range references while maintaining a separate recent context \citep{yang2026anchorforcing, chen2026contextforcing, mao2026packforcing}:
\begin{equation}
    \mathbf{M}_n
    =
    \mathbf{M}^{\mathrm{anchor}}_n
    \cup
    \mathbf{M}^{\mathrm{local}}_n.
\end{equation}
LongLive~\citep{yang2025longlive} and related sink-based designs preserve persistent reference KV states alongside rolling recent history. Recent streaming systems further specialize this allocation across temporal or modality scales. JoyStreamer-Flash~\citep{li2025joystreamer} and LiveTalk~\citep{chern2025livetalk} retain persistent identity states together with rolling KV context, while Stream-T1~\citep{tu2026stream} applies different policies to states leaving the active window. Unlike anchor admission in Section~\S\ref{sec:5.1_writing}, the management problem here concerns how long admitted anchors remain available and how much capacity they receive. Persistent anchors improve long range stability, but stale references can suppress legitimate appearance, viewpoint, or world state changes.

\vspace{-1.2em}
\paragraph{Utility-Aware Retention.}
Rather than allocating capacity solely by age or fixed memory tier, adaptive policies estimate which entries deserve continued retention \citep{Future-Forcing, xu2026sparse, chen2026past, zhao2026densitykv}. Given a utility score $s_\gamma(m)$, a bounded memory can be formed as:
\begin{equation}
    \mathbf{M}_{n+1}
    =
    \operatorname{TopK}_{m \in \widetilde{\mathbf{M}}_{n+1}}
    s_\gamma(m).
\end{equation}
The score may reflect attention participation, visual novelty, retrieval statistics, uncertainty, diversity, or predicted future relevance. Forcing-KV~\citep{ji2026forcing}, for example, assigns different pruning policies to attention heads according to their structural and dynamic roles, while HeadCast~\citep{shen2026headcast} maintains head-specific history budgets \citep{zhao2026densitykv}. DySink~\citep{ye2026dysink} instead selects visually relevant historical states as dynamically changing sinks. Such policies retain informative remote evidence more efficiently than uniform truncation, but depend on an inherently uncertain estimate of future utility. Evidence that appears dispensable locally may become essential only after scene revisitation, delayed interaction, or entity reappearance.

\vspace{-1.2em}
\paragraph{Compression and Consolidation.}
When remote history remains potentially useful, management can reduce its representation cost rather than remove it \citep{samuelfast, ma2026flow}. Recent evidence may remain at high fidelity, while older states are pooled, subsampled, quantized, or projected into compact summaries:
\begin{equation}
    \mathbf{M}^{\mathrm{comp}}_n
    =
    \mathcal{C}_\gamma(\mathbf{M}_n).
\end{equation}
The operator $\mathcal{C}_\gamma$ may implement fixed hierarchical reduction or a learned compression module. Light Forcing~\citep{lv2026light} compresses history across token and block scales, while Ring Forcing~\citep{xue2026ring} adapts between global and detail-preserving compression under a fixed token budget. LongLive-2.0~\citep{chen2026longlive} further combines persistent sink structures and sliding history with low-bit KV storage, illustrating that representation precision can itself become a management variable \citep{Quant-VideoGen}. Compression extends the effective memory horizon, but introduces an information bottleneck. Fine appearance cues, small entities, subtle spatial relations, or rare events may disappear even when the compressed state remains globally coherent. This differs from recurrent updating in Section~\S\ref{sec:5.3_updating}: recurrent updating assimilates new evidence into an evolving state, whereas compression management primarily changes how retained state is represented under a resource budget.

\vspace{-1.2em}
\paragraph{Validity-Aware Forgetting.}
Memory management must also control whether retained evidence remains valid, not merely whether it fits within the available budget. Once an observation becomes obsolete or is judged unreliable, preserving it can be more harmful than removing it. A forgetting operator may therefore suppress selected entries:
\begin{equation}
    \mathbf{M}_{n+1}
    =
    \widetilde{\mathbf{M}}_{n+1}
    \setminus
    \mathcal{D}_\gamma
    (\widetilde{\mathbf{M}}_{n+1}, \mathbf{w}_n, \mathbf{c}),
\end{equation}
where $\mathcal{D}_\gamma$ identifies evidence that should no longer remain active. StableWorld~\citep{yang2026stableworld}, for example, identifies geometrically degraded generated frames and removes unreliable history rather than retaining it according to recency alone. Validity-aware forgetting complements corrective updating but serves a different role: revision changes what the memory state represents, whereas forgetting controls whether an unreliable or obsolete entry should continue to survive. Its difficulty lies in calibration, since premature invalidation can erase genuine long-range evidence, while insufficient forgetting allows stale or erroneous states to repeatedly influence future predictions.

Across these policies, memory management must balance \emph{coverage}, \emph{fidelity}, \emph{cost}, and \emph{validity}. Temporal truncation provides strict resource bounds but loses remote evidence; anchor and multi-tier retention protect selected references but can preserve stale constraints; utility-aware policies allocate capacity more selectively but depend on uncertain estimates of future relevance; compression extends the accessible horizon at the cost of information loss; and validity-aware forgetting prevents obsolete evidence from persisting but requires reliable confidence signals. Management therefore determines not simply how much history is retained, but which evidence survives, at what fidelity, and with what authority over the remaining rollout.

\subsection{Memory Integration}
\label{sec:5.5_integration}

Memory integration governs how retrieved historical evidence influences the current generative computation. Reading determines which memory becomes accessible as $\mathbf{r}_n$, whereas integration determines through which pathway and with what strength that evidence affects the next visual unit. Relevant memory may still be ineffective if it is weakly injected, incompatible with the current representation, or overwhelmed by local context. Formally, given the bounded context $\mathbf{C}_n$, retrieved memory $\mathbf{r}_n$, and generalized condition $\mathbf{c}$, an integration operator constructs the memory-conditioned context:
\begin{equation}
    \mathbf{h}_n
    =
    \mathcal{I}_\eta
    (\mathbf{C}_n, \mathbf{r}_n, \mathbf{c}),
\end{equation}
from which the next visual unit is generated:
\begin{equation}
    \mathbf{y}_n
    \sim
    p_\theta(\mathbf{y}_n \mid \mathbf{h}_n).
\end{equation}
The central challenge is therefore not merely to provide historical evidence, but to make it sufficiently influential, structurally compatible, and reliably accessible while preserving flexibility for legitimate changes in the generated world.

\vspace{-1.2em}
\paragraph{Context Concatenation.}
The simplest pathway appends retrieved memory directly to the current input context \citep{yu2025context, zhang2026frame, zhang2025tinyhistory}. Historical frames, latent prefixes, visual anchors, or retrieved tokens are processed jointly with recent observations by the generator. This strategy requires little architectural specialization and preserves the native interface of sequence-conditioned models. It is effective when retrieved evidence is compact and well aligned with the current target, but provides limited control over utilization. Distant entries may receive little effective attention, irrelevant details may interfere with current synthesis, and increasing the concatenated history consumes the same context budget that persistent memory is intended to alleviate. Direct concatenation therefore converts retrieval into available context without guaranteeing that the generator will use it appropriately.

\vspace{-1.2em}
\paragraph{Attention-Based Conditioning.}
Memory can instead enter generation through a dedicated attention pathway \citep{yu2026memlearner, yu2025malt, SlotMem}. Current generation features query the retrieved memory:
\begin{equation}
    \mathbf{h}_n
    =
    \operatorname{CrossAttn}
    \left(
        \mathbf{Q}_n,
        \mathbf{K}_{\mathbf{r}_n},
        \mathbf{V}_{\mathbf{r}_n}
    \right),
\end{equation}
where $\mathbf{Q}_n$ denotes queries from the current generative features and is distinct from the retrieval query $\mathbf{q}_n$ used in Section~\S\ref{sec:5.2_reading}. Cross-attention supports heterogeneous memories such as entity slots, retrieved visual features, and compressed history tokens while keeping them separate from the local sequence. Its effectiveness, however, depends on compatibility and calibration between current queries and historical keys. Poorly aligned representations can suppress important memory or amplify plausible but irrelevant entries, while large retrieved sets can dilute attention across competing evidence.

\vspace{-1.2em}
\paragraph{KV-State Injection.}
For transformer-based generators, retrieved neural states can be inserted directly into self-attention \citep{yang2025longlive, hong2025relic, xu2026sparse}:
\begin{equation}
\mathbf{h}_n =
\operatorname{Attn}
\left(
\mathbf{Q}_n,
[\mathbf{K}_{\mathrm{local}};\mathbf{K}_{\mathbf{r}_n}],
[\mathbf{V}_{\mathrm{local}};\mathbf{V}_{\mathbf{r}_n}]
\right).
\end{equation}
This pathway reuses persistent activations at the same interface as local context and can preserve fine visual and motion information without decoding memory into an external representation. Its main challenge is compatibility. Historical KV states remain coupled to the positional encodings, activation statistics, and numerical representation under which they were produced. LoL~\citep{cui2026lol}, for example, adjusts positional frequencies across attention heads to reduce sink-dominated utilization, while Quantized Keys Steal Attention~\citep{tuncer2026quantized} corrects attention-logit bias introduced by quantized historical keys. Wan-Streamer~\citep{huang2026wan} directly transfers committed KV states into subsequent generation blocks. These methods illustrate that preserving a cache is insufficient when its representation no longer interacts correctly with the active generator.

\vspace{-1.2em}
\paragraph{Adaptive Modulation and Gating.}
Rather than inserting memory as additional sequence elements, integration can instead regulate intermediate features or the relative influence of memory and local context during generation. A generic formulation is:
\begin{equation}
    \mathbf{h}^{(\ell)}_n
    \leftarrow
    \gamma_\eta(\mathbf{r}_n)
    \odot
    \mathbf{h}^{(\ell)}_n
    +
    \delta_\eta(\mathbf{r}_n),
\end{equation}
or, more explicitly,
\begin{equation}
\mathbf{h}_n
=
\lambda_n
\odot
\mathcal{I}_{\mathrm{mem}}(\mathbf{r}_n)
+
(1-\lambda_n)
\odot
\mathcal{I}_{\mathrm{local}}(\mathbf{C}_n),
\end{equation}
where $\lambda_n$ may depend on relevance, confidence, spatial overlap, temporal distance, or other state-dependent signals. This family allows memory influence to vary across layers, tokens, regions, or rollout steps \citep{ding2026layerrecall, cao2026recency}. Memorize When Needed~\citep{guo2026memorize} exemplifies explicit control over when persistent memory should affect long-horizon generation, while TaoMate~\citep{gan2026taomate} bridges evolving states with persistent reference anchors during streaming generation. TetherMem~\citep{li2026tether} provides a finer-grained realization by constructing attention-logit priors from query role, spatial region, and memory age, thereby regulating how strongly different historical entries affect the current block without rewriting their stored values. The principal challenge is calibration: excessive memory influence can impose stale constraints, whereas insufficient influence makes correctly retrieved evidence operationally irrelevant.

\vspace{-1.2em}
\paragraph{Geometry-Aligned Conditioning.}
Spatial memories require their retained state to be aligned with the target view before it can constrain generation \citep{garcin2026beyond, yu2026mosaicmem, WorldStereo}. Persistent point clouds, surfels, voxel states, maps, or related geometric representations can be projected into view-aligned conditions:
\begin{equation}
    \mathbf{r}^{\mathrm{view}}_n
    =
    \Pi
    (\mathbf{M}^{\mathrm{spatial}}_n,\pi_n),
\end{equation}
where $\Pi$ denotes a projection or rendering operator and $\pi_n$ specifies the target camera. The resulting depth, correspondence, occupancy, feature, or semantic maps can then condition synthesis through the generator's native control pathway. This integration provides strong spatial grounding for scene revisitation and viewpoint consistency, but transfers geometric errors directly into the generative condition. Pose drift, depth uncertainty, or dynamic-object contamination can therefore produce misaligned constraints even when the underlying memory entry was retrieved correctly. This distinguishes integration failure from the retrieval failure discussed in Section~\S\ref{sec:5.2_reading}: the correct spatial state may be selected but incorrectly aligned with the current rendering coordinates.

\vspace{-1.2em}
\paragraph{Structured Semantic Conditioning.}
When memory stores explicit entities, relations, events, or world states, integration can expose these variables as structured tokens, semantic conditions, scene representations, or other explicit controls \citep{pondaven2026actionparty, long20262, VideoMemory}. This pathway is useful when the relevant historical commitment concerns what state currently holds rather than how a previous frame appeared. It can therefore preserve entity roles, interaction consequences, narrative commitments, or task states without retaining all associated visual observations. Its limitation is perceptual under-specification. Knowing that an object moved, a relation changed, or an event occurred does not by itself determine texture, illumination, precise geometry, or other view-dependent appearance. Structured conditioning therefore often works best in conjunction with visual or spatial memory when both state consistency and rendering fidelity are required.

Across these pathways, integration determines whether accessible memory becomes effective memory. Failures can arise from insufficient influence, excessive reliance on stale evidence, or incompatibility between retained states and the current generative representation. Context concatenation provides a simple interface but weak utilization control; attention and KV injection offer direct neural access but require calibrated and compatible representations; modulation and gating regulate influence more explicitly; and geometric or structured conditioning provide stronger inductive structure at the cost of dependence on accurate intermediate states. Integration thus completes the operational lifecycle by converting retrieved history into an actionable constraint on subsequent generation.

\section{Learning: How Is Memory Learned?}
\label{sec:learning}

The preceding sections characterize memory from three complementary perspectives: what carries historical information, what must remain preserved, and how persistent state is operated during sequential rollout. We now turn to the learning question:
\obsbox{
\textit{How do generative models acquire reliable memory behavior from data, supervision, and closed-loop interaction?}
}

\begin{figure*}[!t]
\centering
\vspace{-0.4em}
\includegraphics[width=1\linewidth]{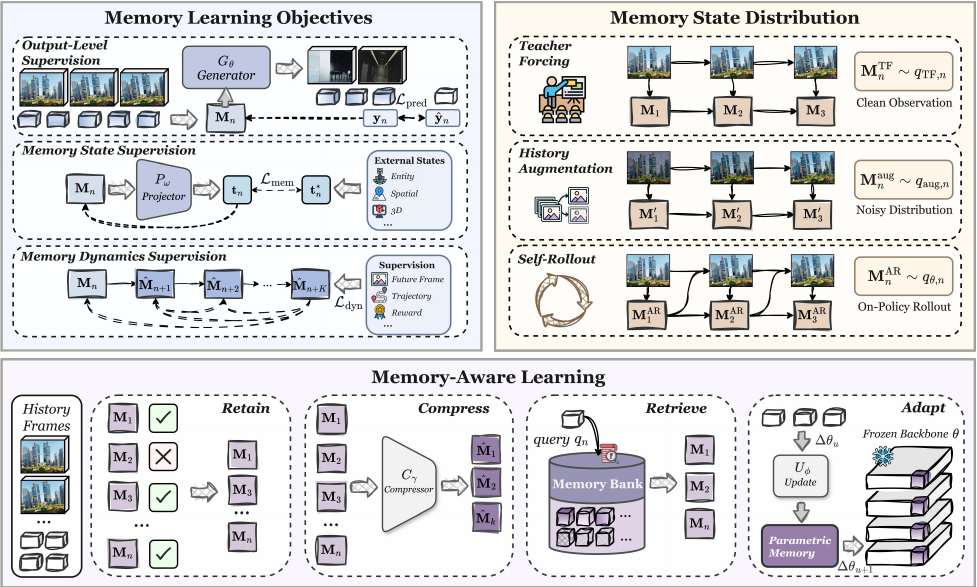}
\vspace{-1.6em}
\caption{Memory learning in autoregressive video generation. Generated outputs are written back into memory and reused as conditions for future generation, forming a closed loop in which prediction errors can alter subsequent memory states and accumulate over time.}
\label{fig:figure_neaser}
\vspace{-0.6em}
\end{figure*}

In autoregressive video generation, memory is an evolving conditional state rather than a static record of past observations. Model-generated frames or latents may be written back into memory, reshaping subsequent conditioning and allowing errors to propagate as generation progresses. As illustrated in Figure~\ref{fig:figure_neaser}, this recursive feedback process can cause prediction and memory-state errors to compound over extended horizons. Memory learning is therefore inherently closed-loop: reliable long-horizon generation requires not only accurate next-step synthesis, but also informative memory representations, coherent state evolution, and robustness to model-generated histories.

Let $\mathbf{M}_n$ denote the memory state available before generating the $n$-th autoregressive visual unit, and let $\hat{\mathbf{y}}_n$ denote the corresponding model-generated unit. With bounded local context $\mathbf{C}_n$, action or control signal $\mathbf{a}_n$, and generalized condition $\mathbf{c}$, a memory-conditioned rollout step can be written as:
\begin{equation}
\mathbf{r}_n
=
\mathcal{R}_\beta(\mathbf{q}_n,\mathbf{M}_n),
\qquad
\hat{\mathbf{y}}_n
=
G_\theta\bigl(
    \boldsymbol{\epsilon}_n;
    \mathcal{I}_\eta(\mathbf{C}_n,\mathbf{r}_n,\mathbf{a}_n,\mathbf{c})
\bigr),
\qquad
\mathbf{M}_{n+1}
=
\mathcal{F}\bigl(
    \mathbf{M}_n,
    \hat{\mathbf{y}}_n,
    \mathbf{a}_n
\bigr).
\end{equation}
Here, $\mathcal{R}_\beta$ retrieves relevant evidence from memory according to query $\mathbf{q}_n$, $\mathcal{I}$ integrates the retrieved state with the current generation context, and $G_\theta$ denotes the step-wise generator with stochastic input $\boldsymbol{\epsilon}_n$. The effective transition $\mathcal{F}$ subsumes post-generation writing, updating, and, when applicable, memory management. For compactness, its dependence on $\mathbf{C}_n$ and $\mathbf{c}$ is suppressed. For non-interactive generation, $\mathbf{a}_n=\varnothing$. Depending on the system, $\mathbf{M}_n$ may consist of visual observations, latent or KV states, recurrent states, retrieval stores, structured entity/scene states, or adaptive parameter states.

\begin{figure*}[!t]
    \centering
    \vspace{-0.4em}
    \begin{tikzpicture}[
        yscale=.88,
        root/.style={draw=black!80, fill=gray!7, rounded corners=2pt, line width=.65pt, minimum width=3.4cm, minimum height=.55cm, align=center, font=\bfseries\small},
        objective/.style={draw=SteelBlue!80!black, fill=SteelBlue!15, rounded corners=2pt, line width=.65pt, text width=2.25cm, minimum height=.82cm, align=center, font=\footnotesize},
        distribution/.style={draw=DarkOliveGreen!80!black, fill=DarkOliveGreen!15, rounded corners=2pt, line width=.65pt, text width=2.25cm, minimum height=.82cm, align=center, font=\footnotesize},
        aware/.style={draw=DarkOrange!85!black, fill=DarkOrange!15, rounded corners=2pt, line width=.65pt, text width=2.25cm, minimum height=.82cm, align=center, font=\footnotesize},
        objectiveleaf/.style={draw=SteelBlue!80!black, fill=SteelBlue!5, rounded corners=2pt, line width=.55pt, text width=3.0cm, minimum height=.70cm, align=center, font=\footnotesize},
        distributionleaf/.style={draw=DarkOliveGreen!80!black, fill=DarkOliveGreen!5, rounded corners=2pt, line width=.55pt, text width=3.0cm, minimum height=.70cm, align=center, font=\footnotesize},
        awareleaf/.style={draw=DarkOrange!85!black, fill=DarkOrange!5, rounded corners=2pt, line width=.55pt, text width=3.0cm, minimum height=.70cm, align=center, font=\footnotesize},
        objectiveexample/.style={draw=SteelBlue!25, fill=SteelBlue!5, rounded corners=2pt, line width=.35pt, text width=6.05cm, minimum height=.66cm, align=left, font=\tiny},
        distributionexample/.style={draw=DarkOliveGreen!25, fill=DarkOliveGreen!5, rounded corners=2pt, line width=.35pt, text width=6.05cm, minimum height=.66cm, align=left, font=\tiny},
        awareexample/.style={draw=DarkOrange!25, fill=DarkOrange!5, rounded corners=2pt, line width=.35pt, text width=6.05cm, minimum height=.66cm, align=left, font=\tiny},
        edge/.style={draw=black!70, line width=.45pt}
    ]
    \node[root, rotate=90] (root) at (-0.60,-3.95) {Memory Learning};
    
    \node[objective] (objective) at (2.50,1.05) {\textbf{Memory Learning\\Objectives}};
    \node[distribution] (distribution) at (2.50,-2.95) {\textbf{Memory State}\\\textbf{Distribution}};
    \node[aware] (aware) at (2.50,-7.60) {\textbf{Memory-Aware}\\\textbf{Learning}};
    
    \node[objectiveleaf] (output) at (6.25,2.35) {\textbf{Output-Level}\\\textbf{Prediction}};
    \node[objectiveleaf] (state) at (6.25,1.05) {\textbf{Memory State}\\\textbf{Supervision}};
    \node[objectiveleaf] (dynamics) at (6.25,-0.25) {\textbf{Memory Dynamics}\\\textbf{Supervision}};
    \node[objectiveexample] (outputex) at (11.60,2.35) {\textit{e.g.}, MAGI-1~\citep{teng2025magi}, VideoAR~\citep{ji2026videoar}, NOVA~\citep{deng2025autoregressive}, VideoMAR~\citep{yu2026videomar}, InfinityStar~\citep{liu2026infinitystar}};
    \node[objectiveexample] (stateex) at (11.60,1.05) {\textit{e.g.}, PERSIST~\citep{garcin2026beyond}, GIM-World~\citep{wei2026geometry}, Memory Forcing~\citep{huang2025memory}};
    \node[objectiveexample] (dynamicsex) at (11.60,-0.25) {\textit{e.g.}, Next Forcing~\citep{xu2026next}, PanoWorld~\citep{jiang2026panoworld}, LIVE~\citep{huang2026live}, Salt~\citep{ge2026salt}, VIGOR~\citep{yin2026vigor}, VideoRLVR~\citep{zhu2026video}};
    
    \node[distributionleaf] (teacher) at (6.25,-1.65) {\textbf{Teacher Forcing}};
    \node[distributionleaf] (augment) at (6.25,-2.95) {\textbf{History}\\\textbf{Augmentation}};
    \node[distributionleaf] (rollout) at (6.25,-4.25) {\textbf{Self-Rollout}};
    \node[distributionexample] (teacherex) at (11.60,-1.65) {\textit{e.g.}, ACDiT~\citep{hu2024acdit}, Ca2-VDM~\citep{gao2024ca2}, GPDiT~\citep{zhang2026generative}, Causal Forcing~\citep{zhu2026causal}};
    \node[distributionexample] (augmentex) at (11.60,-2.95) {\textit{e.g.}, GameNGen~\citep{valevski2025diffusion}, MAGI~\citep{zhou2025taming}, DFoT ~\citep{song2025history}, Diffusion Forcing~\citep{chen2024diffusion}, AR-Diffusion~\citep{sun2025ar}, Resampling Forcing~\citep{guo2025end}};
    \node[distributionexample] (rolloutex) at (11.60,-4.25) {\textit{e.g.}, Self Forcing~\citep{huang2026self}, Rolling Forcing~\citep{liu2026rolling}, Self-Forcing++~\citep{cui2026self}, Causal-rCM~\citep{zheng2026causal}, RAVEN~\citep{lu2026raven}, BAgger~\citep{po2026bagger}, OPSD-V~\citep{liu2026opsd}};
    
    \node[awareleaf] (retain) at (6.25,-5.65) {\textbf{Learning to}\\\textbf{Retain}};
    \node[awareleaf] (compress) at (6.25,-6.95) {\textbf{Learning to}\\\textbf{Compress}};
    \node[awareleaf] (retrieve) at (6.25,-8.25) {\textbf{Learning to}\\\textbf{Retrieve}};
    \node[awareleaf] (adapt) at (6.25,-9.55) {\textbf{Learning to}\\\textbf{Adapt}};
    \node[awareexample] (retainex) at (11.60,-5.65) {\textit{e.g.}, LongLive~\citep{yang2025longlive}, Reward Forcing~\citep{lu2026reward}, Sparse Forcing~\citep{xu2026sparse}, PaFu-KV~\citep{chen2026past}};
    \node[awareexample] (compressex) at (11.60,-6.95) {\textit{e.g.}, FramePack~\citep{zhang2026frame}, PackForcing~\citep{mao2026packforcing}, TinyHistory~\citep{zhang2025tinyhistory}, ARL$^2$~\citep{li2026attend}, Echo-Infinity~\citep{bian2026echo}, VideoMLA~\citep{yesiltepe2026videomla}};
    \node[awareexample] (retrieveex) at (11.60,-8.25) {\textit{e.g.}, Context-as-Memory~\citep{yu2025context}, MosaicMem~\citep{yu2026mosaicmem}, COVRAG~\citep{joo2026retrieve}, LongLive-RAG~\citep{hu2026longlive}, MemLearner~\citep{yu2026memlearner}, CaR~\citep{peng2026compression}};
    \node[awareexample] (adaptex) at (11.60,-9.55) {\textit{e.g.}, LaCT~\citep{zhang2026test}, HippoCampus~\citep{peng2026hippocampus}, RAD-TTT~\citep{chen2025rad}, ISPA~\citep{fu2026towards}};
    
    \draw[edge] (-0.30,-3.95) -- (0.65,-3.95);
    \draw[edge] (0.65,1.05) -- (0.65,-7.60);
    \draw[edge] (0.65,1.05) -- (objective.west);
    \draw[edge] (0.65,-2.95) -- (distribution.west);
    \draw[edge] (0.65,-7.60) -- (aware.west);
    
    \draw[edge] (objective.east) -- (4.50,1.05);
    \draw[edge] (4.50,2.35) -- (4.50,-0.25);
    \draw[edge] (4.50,2.35) -- (output.west);
    \draw[edge] (4.50,1.05) -- (state.west);
    \draw[edge] (4.50,-0.25) -- (dynamics.west);
    \draw[edge] (output.east) -- (outputex.west);
    \draw[edge] (state.east) -- (stateex.west);
    \draw[edge] (dynamics.east) -- (dynamicsex.west);
    
    \draw[edge] (distribution.east) -- (4.50,-2.95);
    \draw[edge] (4.50,-1.65) -- (4.50,-4.25);
    \draw[edge] (4.50,-1.65) -- (teacher.west);
    \draw[edge] (4.50,-2.95) -- (augment.west);
    \draw[edge] (4.50,-4.25) -- (rollout.west);
    \draw[edge] (teacher.east) -- (teacherex.west);
    \draw[edge] (augment.east) -- (augmentex.west);
    \draw[edge] (rollout.east) -- (rolloutex.west);
    
    \draw[edge] (aware.east) -- (4.50,-7.60);
    \draw[edge] (4.50,-5.65) -- (4.50,-9.55);
    \draw[edge] (4.50,-5.65) -- (retain.west);
    \draw[edge] (4.50,-6.95) -- (compress.west);
    \draw[edge] (4.50,-8.25) -- (retrieve.west);
    \draw[edge] (4.50,-9.55) -- (adapt.west);
    \draw[edge] (retain.east) -- (retainex.west);
    \draw[edge] (compress.east) -- (compressex.west);
    \draw[edge] (retrieve.east) -- (retrieveex.west);
    \draw[edge] (adapt.east) -- (adaptex.west);
    \end{tikzpicture}
    \vspace{-0.4em}
\caption{Taxonomy of memory learning methods for autoregressive video generation. Methods are organized by memory learning objectives, memory state distributions, and memory-aware learning; representative works are listed for each leaf category.}
\label{fig:learning_taxonomy}
\vspace{-0.6em}
\end{figure*}
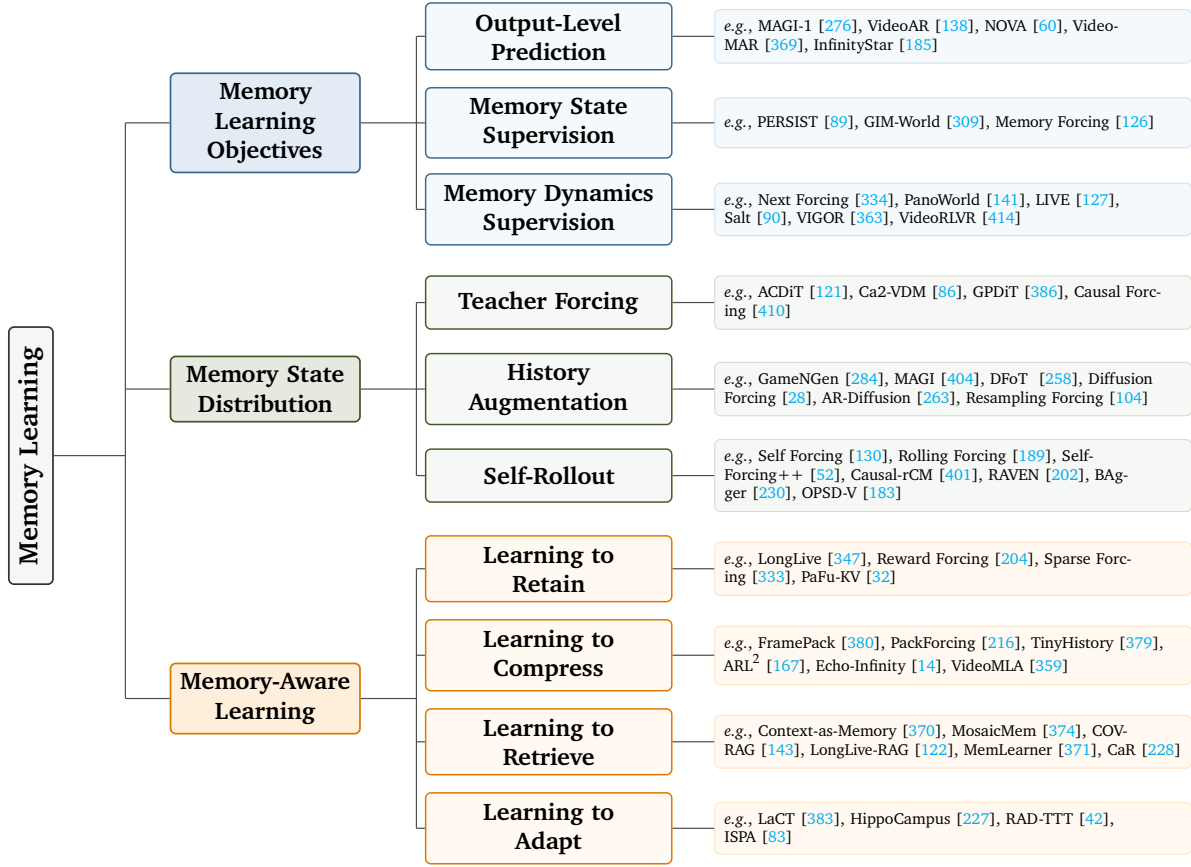

Under this abstraction, reliable memory learning for AR video generation must support both accurate synthesis and persistent memory states that remain informative and coherent throughout autoregressive rollout. These requirements are difficult to satisfy jointly: prediction errors can be written into the evolving state and shift the distribution of memory encountered later in the rollout, while finite context and memory budgets restrict the historical evidence available for subsequent generation.

As summarized in Figure~\ref{fig:learning_taxonomy}, we organize memory learning along three complementary dimensions. \ding{182} \textbf{Memory Learning Objectives} define which components of the learning process receive supervision, including generated outputs, persistent memory states, and memory dynamics. \ding{183} \textbf{Memory State Distributions} characterize the states induced by the histories encountered during training, ranging from clean teacher-forced histories to the model's own autoregressive rollouts. \ding{184} \textbf{Memory-Aware Learning} accounts for the memory interface used at deployment by training the generator under the corresponding retention, compression, retrieval, or adaptation mechanisms, whether fixed or learnable. Together, these dimensions determine which generation and memory properties are optimized, which memory-state distributions are covered during training, and whether the resulting memory mechanism remains effective under deployment-time capacity and access constraints.

\subsection{Memory Learning Objectives}
\label{sec:6.1_learning_objectives}

Memory learning objectives determine which aspects of a memory-conditioned generator receive explicit training signals. We distinguish three supervision targets. \ding{192} \emph{Output-level prediction supervision} optimizes future frames, tokens, latents, or chunks and provides the basic generative learning signal. \ding{193} \emph{Memory state supervision} directly constrains what persistent memory encodes, encouraging the retained state to preserve information required by later generation. \ding{194} \emph{Memory dynamics supervision} constrains how persistent state changes across rollout steps, including the consistency of memory updates and action-conditioned state transitions. These objectives address different parts of the closed-loop process: output losses supervise immediate synthesis, state losses shape the content and correctness of persistent memory, and dynamics losses regularize how that memory evolves over time.

\subsubsection{Output-Level Prediction Supervision}
\label{sec:output_level_prediction_supervision}

Output-level prediction supervision directly optimizes future visual units, such as discrete tokens, frames, hierarchical representations, or spatio-temporal chunks. Let $\mathbf{y}_{1:N}=\{\mathbf{y}_1,\ldots,\mathbf{y}_N\}$ denote a sequence of autoregressive visual units and $\mathbf{c}$ the external condition. Their conditional distribution factorizes as:
\begin{equation}
p_\theta(\mathbf{y}_{1:N}\mid \mathbf{c})
=
\prod_{n=1}^{N}
p_\theta(\mathbf{y}_n\mid \mathbf{y}_{<n},\mathbf{c}).
\end{equation}
Let $\mathbf{H}_n=\mathbf{y}_{<n}$ denote the complete history before predicting $\mathbf{y}_n$. Under unrestricted access to $\mathbf{H}_n$, standard next-step supervision is sufficient in principle to learn the AR conditionals. Practical generators, however, expose only a bounded local context $\mathbf{C}_n$ directly, while information from more distant history must remain available through persistent memory $\mathbf{M}_n$. The memory-conditioned predictor therefore seeks to approximate:
\begin{equation}
p_{\theta}
(\mathbf{y}_n \mid \mathbf{C}_n,\mathbf{M}_n,\mathbf{c})
\approx
p_{\mathrm{data}}
(\mathbf{y}_n \mid \mathbf{H}_n,\mathbf{c}).
\end{equation}
The learning problem is thus not induced by AR factorization itself, but by whether the bounded pair $(\mathbf{C}_n,\mathbf{M}_n)$ retains sufficient information from $\mathbf{H}_n$ for future prediction under finite context, memory, and computation.

The concrete prediction loss depends on the representation and synthesis process. Discrete models typically use next-token cross-entropy \citep{yan2021videogpt, kondratyuk2023videopoet}, whereas continuous generators commonly adopt conditional denoising or flow-matching objectives \citep{voleti2022mcvd, yang2023diffusion,davtyan2023efficient, gu2025long,teng2025magi}. Let $\mathbf{z}_n$ denote the continuous latent representation of target unit $\mathbf{y}_n$. A generic continuous prediction objective can be written as:
\begin{equation}
\mathcal{L}_{\mathrm{pred}}
=
\mathbb{E}_{n,\sigma,\boldsymbol{\epsilon}}
\left[
\left\|
f_\theta
\left(
    \mathbf{z}_n^\sigma,
    \sigma;
    \mathbf{C}_n,
    \mathbf{M}_n,
    \mathbf{c}
\right)
-
\mathbf{f}^{*}
\left(
    \mathbf{z}_n,
    \boldsymbol{\epsilon},
    \sigma
\right)
\right\|_2^2
\right],
\end{equation}
where $\boldsymbol{\epsilon}\sim\mathcal{N}(\mathbf{0},\mathbf{I})$, $\mathbf{z}_n^\sigma$ is the perturbed target latent, and $\mathbf{f}^{*}$ denotes the corresponding noise, velocity, or flow target. Such losses directly supervise the generated output. When memory is learnable end-to-end, its representation is shaped indirectly through its contribution to prediction, but the loss does not independently specify what the persistent state should encode or how it should evolve. Sections~\S\ref{sec:memory_state_supervision} and ~\S\ref{sec:memory_dynamics_supervision} consider objectives that impose more explicit constraints on these properties.

A useful way to organize output-level supervision is by the \emph{prediction structure and effective autoregressive unit}. This choice determines the number of outer rollout decisions, the dependencies resolved within each decision, and how frequently newly generated errors can enter subsequent conditioning and memory.

\vspace{-1.2em}
\paragraph{Token-Level Prediction.}
At the finest autoregressive resolution, VideoGPT \citep{yan2021videogpt} compresses videos into discrete spatio-temporal tokens with a VQ-VAE \citep{van2017neural} and predicts them using a GPT-style causal model. Emu3 \citep{wang2024emu3} and Emu3.5 \citep{cui2025emu3} place text, images, and videos in a unified discrete sequence and learn generation through next-token prediction. Loong~\citep{wang2024loong} further extends this formulation to longer temporal sequences through joint text-video ordering and progressive short-to-long training. Token-wise prediction provides a unified likelihood formulation, but high-resolution videos induce long causal chains in which local errors can propagate through many subsequent decisions. PAR~\citep{wang2025parallelized} reduces this serial burden by predicting weakly dependent distant tokens in parallel while preserving local causal order.

\vspace{-1.2em}
\paragraph{Frame-Level Prediction.}
Frame-wise models retain the frame as the outer temporal unit while resolving its internal spatial structure through a separate synthesis process. RVD \citep{yang2023diffusion} combines deterministic next-frame prediction with diffusion-based residual correction, providing an early frame-wise autoregressive diffusion formulation. DIAMOND \citep{alonso2024diffusion} and GameNGen \citep{valevski2025diffusion} predict future observations from recent frames and actions in interactive environments. NOVA \citep{deng2025autoregressive} combines temporal frame-wise causality with intra-frame spatial set prediction, whereas VideoMAR~\citep{yu2026videomar} couples inter-frame causal modeling with bidirectional masked generation within each frame. These designs shorten the outer causal chain relative to token-wise generation, while still committing a newly synthesized visual state at every frame transition.

\vspace{-1.2em}
\paragraph{Multi-Scale Prediction.}
Rather than synthesizing a target at a single resolution, hierarchical objectives decompose prediction into coarse-to-fine representations. Following the next-scale formulation introduced by VAR \citep{tian2024visual}, SAMPO \citep{wang2026sampo} and VideoAR \citep{ji2026videoar} preserve inter-frame causal ordering while progressively refining multi-scale visual states. SAMPO uses an asymmetric multi-scale tokenizer to compress future dynamics while retaining finer detail for observed frames, whereas VideoAR uses a three-dimensional multi-scale tokenizer and predicts spatio-temporal token maps from coarse to fine. Such objectives expose broad structure before fine detail and can distribute prediction difficulty across scales, but errors introduced at coarse levels may constrain all subsequent refinement.

\vspace{-1.2em}
\paragraph{Chunk-Level Prediction.}
Coarser autoregressive units jointly synthesize multiple frames or spatio-temporal regions before advancing the outer rollout. MAGI-1 \citep{teng2025magi} treats consecutive frames as fixed-length autoregressive chunks. VideoAR \citep{ren2025autoregressive} studies several prediction units, including full frames, key-detail frames, multi-scale representations, and spatio-temporal cubes, and adopts next-cube prediction as its default formulation. InfinityStar \citep{liu2026infinitystar} first establishes appearance through an image pyramid of the initial frame and then generates subsequent video chunks through a three-dimensional volume pyramid. Larger units reduce the number of outer autoregressive transitions and model broader spatio-temporal dependencies jointly, but increase per-step synthesis complexity and cause prediction errors to enter future conditioning at a coarser temporal granularity.

Output-level supervision remains the primary learning signal for generative quality, but it leaves several memory-specific ambiguities. \textbf{(\textit{i}) Supervision dilution} arises because most prediction loss is dominated by locally observable appearance and motion, whereas distant memory may be required only at sparse events such as entity reappearance, scene revisitation, or delayed intervention effects. \textbf{(\textit{ii}) Supervision delay} arises when the consequence of a memory write or update becomes observable only many rollout steps later, weakening credit assignment under finite training horizons. \textbf{(\textit{iii}) Memory-state underconstraint} arises because output losses specify observable predictions rather than the internal state that supports them: different memory representations may yield the same immediate output while differing substantially in what they preserve for later generation. These limitations motivate objectives that supervise persistent memory content and its evolution more directly.

\subsubsection{Memory State Supervision}
\label{sec:memory_state_supervision}

Memory state supervision directly constrains what the memory state $\mathbf{M}_n$ encodes at each autoregressive step, encouraging it to represent historical information required for subsequent generation. A direct formulation aligns $\mathbf{M}_n$, or a projection of it, with a step-specific supervision target:
\begin{equation}
\mathcal L_{\mathrm{mem}} =
\mathbb E_n
\left[
d\!\left(P_\omega(\mathbf{M}_n),\mathbf{m}_n^{\mathrm{tgt}}\right)
\right],
\end{equation}
where $P_\omega$ maps memory into the supervision space and $\mathbf{m}_n^{\mathrm{tgt}}$ denotes the supervision target for the memory state at step $n$. Such targets may be obtained from annotations, simulators, multi-view reconstruction, or pretrained perception models. A complementary strategy constructs explicit memory states from history and uses them as conditioning inputs during generation, thereby training the generator to exploit retained information. We therefore consider both direct supervision of memory content and state-conditioned generation. Their concrete carriers and operations are discussed in Sections~\S\ref{sec:representation} and~\S\ref{sec:operations}.

\vspace{-1.2em}
\paragraph{Explicit World-State Variable Targets.}
One class of supervision targets consists of explicitly specified world-state variables, such as camera poses, depth, occupancy, semantic maps, object attributes, trajectories, and 3D scene states. In generative video world models, PERSIST~\citep{garcin2026beyond} derives proxy world states from simulator-provided 3D frames and camera parameters, supervising both 3D state prediction and camera regression. PanoWorld~\citep{jiang2026panoworld} uses pseudo-ground-truth panoramic depth to constrain geometric memory states. Such objectives ground memory in spatial or semantic state variables shared across views, providing more direct supervision than future output prediction losses alone. However, their scalability is limited by the cost and availability of structured supervision, and predefined state variables may capture only a subset of the information required for future generation.

\vspace{-1.2em}
\paragraph{Learned Representation Targets.}
When explicit world-state targets are unavailable, memory can instead be supervised using high-dimensional features produced by pretrained perceptual or geometric models. GIM-World~\citep{wei2026geometry} uses geometric features extracted by a frozen 3D foundation model as $\mathbf{m}_n^{\mathrm{tgt}}$ and aligns the memory representation with these targets across views, encouraging $\mathbf{M}_n$ to preserve view-consistent geometry. Such learned targets reduce dependence on manually specified state variables, but their effectiveness depends on whether the pretrained representation captures information that remains relevant to future generation, rather than merely providing an auxiliary feature-matching signal.

\vspace{-1.2em}
\paragraph{State-Conditioned Generation.}
Complementary to direct state supervision, some methods train the generator to use explicitly constructed memory states as conditions. \citet{wu2026video} maintains a static point cloud and episodic key frames as spatial and visual conditions for diffusion generation. \citet{zhou2026learning} aggregates RGB-D history into a 3D feature map and trains a memory-conditioned denoising model to use this map. Memory Forcing~\citep{huang2025memory} combines geometrically indexed spatial memory with mixed training, encouraging reliance on recent temporal context during exploration and on spatial memory during revisitation. In these methods, memory is not merely an auxiliary target; it becomes part of the conditioning pathway through which the generation loss evaluates whether retained information is useful.

Overall, memory state supervision complements output-level prediction in two ways. Direct content objectives constrain what future-relevant information $\mathbf{M}_n$ encodes, whereas state-conditioned generation trains the generator to exploit constructed memory during prediction. Neither route alone ensures reliable long-horizon memory: supervision targets may cover only a subset of the information needed later, while the generator may still bypass memory when local context is sufficient. More fundamentally, these objectives constrain memory at individual steps but do not explicitly regulate how it evolves across successive autoregressive steps. We therefore next consider objectives that supervise memory dynamics.

\subsubsection{Memory Dynamics Supervision}
\label{sec:memory_dynamics_supervision}

Memory dynamics supervision constrains how persistent memory evolves across successive autoregressive steps. As new observations, generated outputs, or actions become available, memory updates should incorporate genuine state changes while preserving previously established information that remains valid. Let $\mathcal{F}^{(k)}$ denote the composition of $k$ consecutive effective memory transitions. A generic multi-step state objective can be written as:
\begin{equation}
\mathcal{L}_{\mathrm{dyn}}
=
\sum_{k=1}^{K}
w_k\,
d\!\left(
\mathcal{F}^{(k)}
\left(
    \mathbf{M}_n,
    \hat{\mathbf{y}}_{n:n+k-1},
    \mathbf{a}_{n:n+k-1}
\right),
\mathbf{M}_{n+k}^{*}
\right),
\end{equation}
where $K$ denotes the supervision horizon, $w_k$ weights the $k$-step objective, $\mathbf{M}_{n+k}^{*}$ denotes the target state after $k$ transitions, and $d(\cdot,\cdot)$ measures discrepancy between predicted and target states, possibly after projection or decoding into image, latent, semantic, geometric, or feature space. Unlike per-step state matching, dynamics supervision constrains how memory behaves over multiple transitions. Existing approaches can be broadly organized into long-horizon outcome constraints and transition-path constraints.

\vspace{-1.2em}
\paragraph{Long-Horizon Outcome Constraints.}
These objectives supervise memory dynamics through the delayed consequences of state evolution, using distant predictions, cross-time geometric relations, or task- and perception-based rewards. Next Forcing~\citep{xu2026next} trains a shared state to predict multiple future chunks rather than only the next chunk, using multi-scale temporal units to provide denser long-horizon supervision. PanoWorld~\citep{jiang2026panoworld} supervises trajectories of tracked points in world coordinates, linking the same physical points across frames to constrain spatial coherence. When explicit target states are unavailable, supervision can instead arise from geometric, task-based, or perceptual signals. VIGOR~\citep{yin2026vigor} uses cross-frame 3D point reprojection errors to penalize deformation, spatial drift, and depth conflicts; VideoRLVR~\citep{zhu2026video} uses rule-based rewards in verifiable environments to assess physical and logical states; and Persistent Robot World Models~\citep{bardhan2026persistent} generate futures of different lengths from the same self-rollout state and compare them using multi-view perceptual rewards. Rather than requiring element-wise agreement between memory states, these objectives constrain whether state evolution produces coherent long-horizon outcomes.

\vspace{-1.2em}
\paragraph{Transition-Path Constraints.}
A complementary strategy constrains the state-evolution path more directly. LIVE~\citep{huang2026live} rolls forward from a real initial state and then reconstructs that state through reverse generation, applying a diffusion loss at the end of the cycle. This recoverability objective penalizes forward trajectories that discard information required to reconstruct the initial state, while a progressive curriculum extends training from short to long cycles. Salt~\citep{ge2026salt} enforces endpoint consistency under composed denoising updates and uses cache-conditioned feature alignment under histories of varying quality. Delta Forcing~\citep{wu2026delta} estimates transition compatibility from latent discrepancies between teacher and generator trajectories, defining an adaptive trust region between teacher supervision and the student trajectory. Cyclic objectives therefore constrain information preservation over a complete path, whereas compatibility objectives regulate agreement across local transitions, cache conditions, or trajectory segments.

These two forms of dynamics supervision address complementary aspects of memory evolution. Long-horizon outcome constraints assess whether memory supports correct downstream behavior over extended rollouts, but their delayed signals can make it difficult to attribute errors to specific writes or updates. Transition-path constraints provide more localized guidance on how memory should evolve, but often require additional rollouts, reverse processes, teacher trajectories, or structural assumptions. Together, they move memory learning beyond per-step state matching toward coherent state evolution throughout long-horizon autoregressive generation.

\subsection{Memory State Distribution}
\label{sec:memory_state_distribution}

Memory state distribution concerns which persistent states the model encounters during training and how they differ from those visited at deployment. In AR video generation, this mismatch primarily arises from how history is constructed. Training commonly builds memory from ground-truth visual units, whereas deployment must progressively update memory from the model's own generated outputs. Thus, even under the same memory transition, training and inference can induce different distributions over persistent states.

Let $\mathcal{F}$ denote the effective memory transition that incorporates writing, updating, and, when applicable, memory management. Under teacher forcing and AR rollout, the memory evolution can be written as:
\begin{equation}
\begin{aligned}
\mathbf{M}_{n+1}^{\mathrm{TF}}
&=
\mathcal{F}
\left(
    \mathbf{M}_{n}^{\mathrm{TF}},
    \mathbf{y}_{n},
    \mathbf{a}_{n}
\right),
&
\mathbf{y}_{n}
&\sim
p_{\mathrm{data}}
\left(
    \,\cdot
    \mid
    \mathbf{H}_{n},
    \mathbf{a}_{\le n},
    \mathbf{c}
\right),
\\
\mathbf{M}_{n+1}^{\mathrm{AR}}
&=
\mathcal{F}
\left(
    \mathbf{M}_{n}^{\mathrm{AR}},
    \hat{\mathbf{y}}_{n},
    \mathbf{a}_{n}
\right),
&
\hat{\mathbf{y}}_{n}
&\sim
p_{\theta}
\left(
    \,\cdot
    \mid
    \mathbf{C}_{n},
    \mathbf{M}_{n}^{\mathrm{AR}},
    \mathbf{a}_{n},
    \mathbf{c}
\right).
\end{aligned}
\end{equation}
Because $\mathbf{y}_n$ and $\hat{\mathbf{y}}_n$ arise from different distributions, repeated transitions induce different distributions over persistent memory states. Let $\mathbf{M}_{n}^{\mathrm{TF}}\sim q_{\mathrm{TF},n}$ and $\mathbf{M}_{n}^{\mathrm{AR}}\sim q_{\theta,n}$ denote the corresponding state distributions. Unlike $p_\theta$, which directly parameterizes generated visual outputs, $q_{\theta,n}$ is an induced distribution produced by recursively writing model-generated outputs into memory and applying $\mathcal{F}$. In general,
\begin{equation}
q_{\mathrm{TF},n}
\neq
q_{\theta,n}.
\end{equation}
This mismatch extends the classical exposure-bias problem in sequence prediction, where models trained on reference histories must condition on their own predictions at inference, from generated outputs to the persistent states that condition future generation~\citep{bengio2015scheduled}. From a state-distribution perspective, it is also closely related to the covariate shift addressed by dataset aggregation in imitation learning, which trains policies on states induced by their own behavior~\citep{ross2011reduction}. In AR video generation, a small error in $\hat{\mathbf{y}}_n$ may be written into memory, alter subsequent states and conditioning signals, and progressively move the rollout into regions poorly covered by teacher-forced training. We therefore organize existing approaches according to how training-time memory states are constructed: \ding{192} \emph{teacher forcing} constructs states from ground-truth histories; \ding{193} \emph{history augmentation} perturbs, masks, or resamples reference histories to induce a broader distribution $q_{\mathrm{aug}}$; and \ding{194} \emph{self-rollout} constructs states through the model's own autoregressive process, directly exposing the model to states induced by its generated history.

\subsubsection{Teacher Forcing}
\label{sec:teacher_forcing}

Teacher forcing constructs the memory state $\mathbf{M}_n^{\mathrm{TF}}$ from the ground-truth history $\mathbf{H}_n=\mathbf{y}_{<n}$ and trains the model to predict $\mathbf{y}_n$. Accordingly, $q_{\mathrm{TF},n}$ is induced by ground-truth histories, whereas the deployment-time distribution $q_{\theta,n}$ is induced by histories generated recursively by the model. This regime is computationally efficient for causal Transformers because complete training sequences can be processed in parallel under causal masks without explicitly unrolling autoregressive generation.

Several AR video diffusion models instantiate this regime. ACDiT~\citep{hu2024acdit} conditions each noisy video block on preceding clean blocks. Ca2-VDM~\citep{gao2024ca2} retains a clean prefix while causally denoising subsequent frames, whereas GPDiT~\citep{zhang2026generative} predicts noisy frames from preceding clean frames using frame-level causal attention. Despite differences in prediction units and masking schemes, these methods share the defining property of teacher forcing: predictions are conditioned on ground-truth context rather than the model's own recursively generated outputs. Teacher-forced training can also provide an initialization for subsequent causal distillation. Causal Forcing~\citep{zhu2026causal} first trains a multistep autoregressive diffusion teacher under teacher forcing and then uses teacher trajectories for ODE-based initialization of a causally conditioned student.

This dependence on clean ground-truth context creates the familiar train-test distribution gap known as exposure bias. In memory-conditioned autoregressive generation, this mismatch also extends to persistent memory states. Teacher-forced training does not expose the model to errors that arise when its predictions are repeatedly fed back as future context, including appearance drift, geometric distortion, motion degradation, and conflicts with stale memory. Once such errors enter KV caches, recurrent states, retrieval stores, or explicit world states, recursive updates may drive memory into regions poorly represented under $q_{\mathrm{TF},n}$. Teacher forcing therefore offers stable and parallelizable training and can serve as an effective initialization, but it provides limited exposure to the model-induced state distribution $q_{\theta,n}$ and cannot by itself eliminate the exposure-bias-induced mismatch in memory states.

\subsubsection{History Augmentation}
\label{sec:history_augmentation}

History augmentation broadens the memory-state distribution encountered during training by transforming states constructed from ground-truth histories. These transformations alter the content, availability, or reliability of the conditioning history, exposing the model to incomplete, corrupted, selectively retained, or model-assisted memory states. Let $K_\psi(\mathbf{M}\mid\mathbf{M}')$ denote a transformation kernel that maps a teacher-forced state $\mathbf{M}'$ to an augmented state $\mathbf{M}$. The induced distribution at step $n$ is:
\begin{equation}
q_{\mathrm{aug},n}(\mathbf{M}) =
\int
K_\psi(\mathbf{M}\mid\mathbf{M}')
q_{\mathrm{TF},n}(\mathbf{M}')
\,\mathrm d\mathbf{M}' .
\end{equation}
Here, $K_\psi$ may implement noise injection, masking, history selection, or model-assisted reconstruction. When the current generator is used to construct augmented states, the kernel may additionally depend on $\theta$, with this dependence suppressed for notational simplicity. A transformation is considered history augmentation only when it affects the state used to condition subsequent predictions.

\vspace{-1.2em}
\paragraph{Temporally Unstructured Perturbations.}
The simplest strategies apply predefined, model-independent transformations to ground-truth histories without imposing an explicit temporal ordering on corruption severity. GameNGen~\citep{valevski2025diffusion} adds random noise to conditioning frames and conditions the generator on the corresponding noise level, encouraging robustness to corrupted context. MAGI~\citep{zhou2025taming} injects dynamic noise into observation latents under teacher forcing to improve robustness to errors in the conditioning history. DFoT~\citep{song2025history} varies the amount of available history, whereas Diffusion Forcing~\citep{chen2024diffusion} independently samples noise levels for temporal tokens, exposing the model to conditioning histories with nonuniform reliability. Collectively, these transformations broaden the training state distribution through corruption, truncation, and heterogeneous noise while preserving the efficiency of parallelizable training. However, because they impose no explicit temporal ordering, they do not directly capture how the reliability of conditioning history may deteriorate over time.

\vspace{-1.2em}
\paragraph{Temporally Structured Perturbations.} 
A second group of predefined methods imposes ordered corruption schedules across temporal positions, or jointly across temporal and denoising dimensions, to approximate the growth of uncertainty over the prediction horizon. Rolling Diffusion Models~\citep{ruhe2024rolling} assign progressively higher noise levels to frames farther into the prediction horizon within a rolling window. AR-Diffusion~\citep{sun2025ar} constrains corruption timesteps to be non-decreasing across frames, such that later frames are at least as corrupted as earlier ones, while SkyReels-V2~\citep{chen2025skyreels} adopts a similar schedule for long-video diffusion forcing. Diagonal Distillation~\citep{liu2026streaming} introduces Diagonal Forcing, which propagates partially denoised history through the KV cache under a prescribed cross-chunk noise schedule, aligning training with its asymmetric denoising trajectory at inference. In discrete autoregressive generation, VideoAR~\citep{ji2026videoar} combines progressively increasing bit-flip ratios across frames with cross-scale perturbation propagation and random history masking. These schemes explicitly model the deterioration of conditioning reliability over time, but their temporal profiles remain prescribed rather than induced by the model's complete AR rollouts.

\vspace{-1.2em}
\paragraph{Model-Derived Perturbations.} Beyond prescribed transformations, some methods derive augmented histories from errors produced by the current generator. Resampling Forcing~\citep{guo2025end} perturbs ground-truth history frames to intermediate diffusion timesteps and uses the online model to complete the remaining denoising trajectory autoregressively, producing histories that capture model-specific errors and their cross-frame propagation. Gradients are stopped through resampling, while a sparse causal mask preserves parallel loss evaluation. Stable Video Infinity~\citep{li2026stable} instead estimates latent and noise residuals through one-step bidirectional integration, stores them in timestep-indexed replay banks, and reinjects them into subsequent flow-matching inputs. Both methods incorporate model-derived failure patterns without unrolling complete deployment trajectories; because their histories remain initialized from or recombined with ground-truth states, the resulting distributions approximate rather than fully match $q_{\theta,n}$.

Overall, history augmentation broadens the training state distribution beyond $q_{\mathrm{TF},n}$ by exposing the generator to corrupted, incomplete, temporally ordered, or model-derived histories without requiring complete deployment-time rollouts. This broader coverage can improve robustness to unreliable conditioning, missing context, and errors accumulated in the history. However, the resulting distribution $q_{\mathrm{aug},n}$ remains only an approximation to the deployment-time distribution $q_{\theta,n}$. Predefined perturbations impose error patterns that may not reflect the coupled failures arising from autoregressive feedback, such as identity drift, geometric distortion, motion degradation, and conflicts with stale memory. Model-derived resampling and error recycling incorporate errors produced by the current generator, but they initialize from or recombine these errors with ground-truth states and therefore reproduce only part of the temporal dependencies created by repeated memory updates. More direct exposure to $q_{\theta,n}$ requires constructing memory from the model's own autoregressive rollouts, which we discuss next.

\subsubsection{Self-Rollout}
\label{sec:self_rollout}

Self-rollout constructs training memory states by unrolling the model's own autoregressive generation process. At step $n$, the generated output $\hat{\mathbf{y}}_n$ is incorporated into memory to produce the next persistent state $\mathbf{M}_{n+1}$, which then conditions subsequent predictions. Unlike teacher forcing and history augmentation, whose conditioning histories are constructed from or initialized around ground-truth trajectories, self-rollout allows the model's prediction errors to enter memory and influence subsequent state transitions. If training follows the same generation, memory-update, and memory-access procedures as deployment, the resulting states are samples from the deployment-time distribution $q_{\theta,n}$; truncated rollouts or mismatched procedures instead provide only an approximation.

Self Forcing~\citep{huang2026self} exemplifies this training paradigm in autoregressive video diffusion. During training, it generates video chunks autoregressively with a KV cache, conditions each chunk on previously generated outputs, and applies holistic video-level supervision to the resulting sequence. Feeding generated chunks back into the cache exposes training to both prediction errors and the memory states created as those errors propagate through subsequent cache updates. This differs from model-derived history perturbations, which either initialize resampling from corrupted ground-truth states or recombine stored errors with clean samples. The primary cost of self-rollout is sequential generation. Measured in denoising-network evaluations, a rollout of $N$ chunks with $S$ denoising steps per chunk requires $O(NS)$ sequential evaluations. Backpropagating through the complete rollout additionally requires retaining intermediate activations and memory states across the unrolled computation. Practical methods therefore commonly employ few-step generators, truncated backpropagation, or stochastic gradient truncation, reducing computational and memory costs at the expense of credit assignment across distant memory updates.

\vspace{-1.2em}
\paragraph{Rollout Organization.}
Self-rollout methods further differ in how generated trajectories are organized into training states. RAVEN~\citep{lu2026raven} repackages each self-rollout as an interleaved sequence of fully denoised historical endpoints and noisy intermediate states, aligning training-time attention with inference-time extrapolation and allowing later chunk losses to supervise the history representations on which they depend. Rolling Forcing~\citep{liu2026rolling} trains over non-overlapping rolling windows conditioned on self-generated histories and jointly denoises multiple frames with progressively increasing noise levels. BAgger~\citep{po2026bagger} reverses drifted model rollouts to construct corrective trajectories that move from degraded states toward earlier, cleaner states. These methods demonstrate that rollout coverage depends not only on trajectory length, but also on how visited states are repackaged, segmented into windows, or reorganized into corrective examples.

\vspace{-1.2em}
\paragraph{Rollout-State Supervision.} A second design axis concerns how supervision is propagated through states visited during self-rollout. Self Gradient Forcing~\citep{zhuang2026self} uses a no-gradient autoregressive rollout followed by parallel context-gradient reconstruction, enabling future-latent losses to supervise how generated context is encoded into causal KV memory without differentiating through the serial rollout. Other methods construct corrective signals through consistency learning, teacher guidance, distribution matching, or dense denoising targets. Causal-rCM~\citep{zheng2026causal} initializes the student with teacher-forced consistency learning before applying distribution matching on self-rollouts. Self-Forcing++~\citep{cui2026self} re-noises local windows sampled from the student's long rollouts and uses a short-horizon bidirectional teacher for distribution matching. OPSD-V~\citep{liu2026opsd} retains the student's deployment-time sampler and KV-cache updates while deriving dense velocity targets from a cleaner, autoregressively consistent teacher cache. Alaya-EVOKE~\citep{yin2026alaya} applies long-horizon distribution matching with a redesigned teacher that supervises extended rollouts. Despite using different gradient paths and corrective targets, these methods all supervise states induced by the student's recursive generation rather than ground-truth histories.

Overall, self-rollout offers the most direct way to narrow the training-deployment memory-state gap. Lengthening teacher-forced sequences exposes the model to later autoregressive steps, but the corresponding states remain distributed according to $q_{\mathrm{TF},n}$ rather than being induced by the model's own errors. Increasing rollout depth broadens coverage of $q_{\theta,n}$, but also raises sequential sampling cost, activation memory, and optimization difficulty. Finite rollout horizons or mismatches between training and deployment samplers, memory updates, or access rules leave a residual state-distribution gap, whereas gradient truncation limits how supervision propagates across distant updates even when the forward rollout remains on-policy. Self-rollout training must therefore balance rollout depth, state coverage, supervision horizon, and optimization cost under feasible computational resources.

\subsection{Memory-Aware Learning}

Even when training matches deployment in the histories used to construct memory states, a second mismatch may arise from the interface through which those states are maintained and accessed. During training, the generator may be exposed to rich or complete histories, whereas deployment is constrained by context length, memory capacity, and computation. The deployed system may therefore retain only selected historical units, compress distant evidence into compact representations, or retrieve nonlocal information on demand. These constraints change not only the amount of available history, but also the structure and distribution of the conditioning signals presented to the generator.

We define \emph{memory-aware learning} as training that explicitly incorporates deployment-time memory constraints, including capacity limits, representation transformations, and access rules. The goal is to train the generator and, when applicable, the memory policy under an interface matched to that used at inference.

Let $\mathbf{H}n$ denote the complete history before generation step $n$, $\mathbf{M}n^{\mathrm{act}}$ the active portion of persistent memory, and $\mathbf{E}n$ an external or inactive memory store. We use $B=(B{\mathrm{act}},B{\mathrm{cmp}},B{\mathrm{ret}})$ to collect the active-memory, compressed-memory, and retrieval budgets. Unlike the bounded local context $\mathbf{C}_n$, $\mathbf{M}_n^{\mathrm{act}}$ represents actively accessible persistent memory. Retention, compression, retrieval, and subsequent generation are then abstracted as:
\begin{equation}
\label{eq:memory_aware_interface}
\begin{aligned}
\widetilde{\mathbf{M}}_n^{\mathrm{act}}
&=
\mathcal{U}_\phi
\left(
    \mathbf{M}_{n-1}^{\mathrm{act}},
    \mathbf{w}_{n-1}
\right),
\\
\mathbf{M}_n^{\mathrm{act}}
&=
\mathcal{G}_\gamma
\left(
    \widetilde{\mathbf{M}}_n^{\mathrm{act}};
    B_{\mathrm{act}}
\right),
\qquad
\operatorname{size}
\left(
    \mathbf{M}_n^{\mathrm{act}}
\right)
\leq B_{\mathrm{act}},
\\
\bar{\mathbf{H}}_n
&=
\mathcal{A}
\left(
    \mathbf{H}_n,
    \mathbf{c};
    B_{\mathrm{cmp}}
\right),
\qquad
\operatorname{size}
\left(
    \bar{\mathbf{H}}_n
\right)
\leq B_{\mathrm{cmp}},
\\
\mathbf{M}_n
&:=
\left(
    \mathbf{M}_n^{\mathrm{act}},
    \bar{\mathbf{H}}_n,
    \mathbf{E}_n
\right),
\\
\mathbf{q}_n
&=
\mathcal{Q}_\theta
\left(
    \mathbf{C}_n,
    \mathbf{a}_n,
    \mathbf{c}
\right),
\qquad
\mathbf{r}_n
=
\mathcal{R}_\beta
\left(
    \mathbf{q}_n,
    \mathbf{M}_n;
    B_{\mathrm{ret}}
\right),
\\
\mathbf{h}_n
&=
\mathcal{I}_\eta
\left(
    \mathbf{C}_n,
    \mathbf{r}_n,
    \mathbf{a}_n,
    \mathbf{c}
\right),
\qquad
\hat{\mathbf{y}}_n
\sim
p_\theta
\left(
    \,\cdot
    \mid
    \mathbf{h}_n
\right).
\end{aligned}
\end{equation}
Here, $\mathbf{w}_{n-1}$ is the write candidate extracted after the preceding visual unit is generated or observed. The update operator $\mathcal{U}_\phi$ incorporates this candidate into the active state, while the management operator $\mathcal{G}_\gamma$ enforces the active-memory budget through retention, consolidation, or revision. The function $\operatorname{size}(\cdot)$ denotes the relevant resource measure, such as the number of frames, tokens, latent blocks, KV entries, or memory slots. The aggregation operator $\mathcal{A}$ produces a budgeted compressed representation $\bar{\mathbf{H}}_n$ of historical evidence. Although written as a map from the complete history for notational clarity, it must be implemented causally at deployment and may be updated incrementally. Updates to the external store $\mathbf{E}_n$ are left implicit because methods differ in whether this store is fixed, append-only, or actively managed.

The persistent memory state $\mathbf{M}_n$ collects the active, compressed, and external components used by the memory interface. The query-formation operator $\mathcal{Q}_\theta$ constructs the retrieval query $\mathbf{q}_n$, the reading operator $\mathcal{R}_\beta$ exposes a memory context $\mathbf{r}_n$ under budget $B_{\mathrm{ret}}$, and $\mathcal{I}_\eta$ integrates the retrieved information with the bounded local context and current conditions. Equation~\eqref{eq:memory_aware_interface} explicitly describes contextual and external memory; parametric adaptation can be incorporated by allowing the effective generator parameters to depend on the parameter-memory state $\mathbf{M}_n^{\mathrm{par}}$ or the accumulated update $\Delta\theta_u$. Not every method instantiates all memory components, and unused components and operators may be omitted.

A generic memory-aware objective can be written as:
\begin{equation}
\label{eq:memory_aware_objective}
\mathcal{L}_{\mathrm{aware}}
=
\mathbb{E}_{
\substack{
    (\mathbf{H}_n,\mathbf{y}_n,
     \mathbf{a}_n,\mathbf{c})
    \sim p_{\mathrm{train}},
    \\
    B\sim p_B
}}
\left[
\mathcal{L}_{\mathrm{pred}}
\left(
    \theta;
    \mathbf{y}_n,
    \mathbf{h}_n
\right)
+
\lambda
\mathcal{L}_{\mathrm{mem}}
\left(
    \mathbf{M}_n,
    \mathbf{H}_n
\right)
\right].
\end{equation}
Here, $p_{\mathrm{train}}$ denotes the selected training-state construction regime rather than only the empirical data distribution; depending on the preceding design choice, it may induce teacher-forced, augmented, or self-rollout memory states. The prediction loss $\mathcal{L}_{\mathrm{pred}}$ evaluates generation under the constrained interface, while $\mathcal{L}_{\mathrm{mem}}$ optionally regularizes information preservation, compression fidelity, retrieval quality, retention behavior, or adaptation stability. Dependence on the memory-operator parameters is suppressed for compactness. Learnable operators can be optimized jointly with the generator, whereas fixed operators still expose the generator to deployment-relevant memory conditions during training.

We distinguish four recurring pathways for memory-aware learning. \ding{192} \emph{Learning to retain} determines which historical units remain active and for how long. \ding{193} \emph{Learning to compress} maps long histories into compact representations that preserve information required for future generation. \ding{194} \emph{Learning to retrieve} selects and reactivates relevant inactive memory under a limited access budget. \ding{195} \emph{Learning to adapt} incorporates historical evidence into an updatable parameter-memory state that persists across subsequent generation steps. The first three pathways regulate contextual or external memory through explicit capacity, representation, and access constraints and are directly represented in Equation~\eqref{eq:memory_aware_interface}. Parameter adaptation instead operates through $\Delta\theta_u$ and is governed by update cost, parameter capacity, interference, and update stability.

\subsubsection{Learning to Retain}
\label{sec:learning_no_retain}

Learning to retain concerns how historical units are selected and maintained under bounded active-memory capacity. These units may be frames, tokens, latent blocks, or KV states. Memory-aware retention involves two coupled questions: which information should remain accessible, and whether the generator is trained under the retention policy used at deployment. Existing methods follow two main routes: training the generator under a predefined retention policy, or learning the retention policy jointly with the generator.

\vspace{-1.2em}
\paragraph{Predefined Retention Policies.}
One line of work trains the generator under a predefined retention policy, allowing it to operate with the same bounded history available at inference. LongLive~\citep{yang2025longlive} combines a short attention window with frame-level attention sinks, retaining a small set of early frames as long-range anchors while using recent history to support local temporal continuity. Its streaming long tuning exposes the generator to this cache layout during training. Reward Forcing~\citep{lu2026reward} introduces EMA-Sink, which initializes a fixed number of sink tokens from early frames and continuously updates them by incorporating evicted tokens through an exponential moving average. The resulting slots combine long-range context with recent dynamics under fixed memory capacity. Together, these methods show that memory-aware retention need not explicitly learn retention decisions; training the generator under the deployment-time policy can allow it to adapt to the resulting constrained memory state.

\vspace{-1.2em}
\paragraph{Learned Retention Policies.} 
A second line learns retention decisions from generation-relevant supervision, either independently or jointly with the generator \citep{Future-Forcing, tian2026head, zhuang2026self}. Sparse Forcing~\citep{xu2026sparse} learns to preserve and update salient KV blocks while dynamically selecting a local key-block neighborhood for each query. Using the same sparse mechanism during training and inference allows the generator and retention policy to co-adapt. PaFu-KV~\citep{chen2026past} instead distills a lightweight salience-estimation head from a bidirectional teacher to predict token utility and retain informative KV states under a bounded cache. Both methods make retention an optimized component of the memory interface rather than relying on a predefined layout.

A central challenge for memory-aware retention is that the short-term salience of a historical unit does not necessarily predict its long-term utility. A unit may have little influence on the current prediction but become essential when an object reappears or a scene is revisited. Conversely, a highly attended self-generated state may be unreliable because it already contains accumulated generation errors. An effective retention policy must therefore balance expected future utility, state reliability, and temporal coverage under the active-memory budget $B_{\mathrm{act}}$.

\subsubsection{Learning to Compress}
\label{sec:learning_to_compress}

Learning to compress aims to represent long histories under bounded memory resources while preserving information needed for future generation. Compression may reduce the representation cost of retained historical units while preserving their temporal structure, or consolidate a variable-length history into a bounded state. Because compression changes both the information available to the generator and the representation through which it is accessed, training should expose the model to the same compression interface used at deployment.

\vspace{-1.2em}
\paragraph{Explicit-History Compression.}
One strategy compresses retained history while preserving explicit temporal or positional structure. FramePack~\citep{zhang2026frame} encodes input frames into varying numbers of context tokens, keeping the total context length approximately constant as video duration increases, and trains the next-frame model through this packing interface. PackForcing~\citep{mao2026packforcing} partitions history into sink, mid, and recent segments, retaining higher resolution for sink and recent content while applying stronger spatiotemporal compression and dynamic selection to intermediate history. The two methods allocate representation capacity differently across time: FramePack varies token counts across frames, whereas PackForcing varies spatiotemporal resolution across history segments. VideoMLA~\citep{yesiltepe2026videomla} compresses KV representations through Multi-Head Latent Attention, replacing per-head content states with a shared low-rank latent while retaining a decoupled 3D-RoPE positional key, and adapts the generator to this representational bottleneck.

\vspace{-1.2em}
\paragraph{State-Summary Compression.}
A second strategy consolidates variable-length history into a bounded persistent state rather than retaining each historical unit explicitly. TinyHistory~\citep{zhang2025tinyhistory} pretrains a lightweight history encoder with a random-frame query objective and then integrates it into an AR video diffusion model with a content-consistency objective. ARL$^2$~\citep{li2026attend} replaces cross-frame softmax attention with a fixed-size recurrent linear-attention state and updates this state only after frame denoising is complete, so persistent memory is updated from completed frame representations rather than intermediate denoising states. Echo-Infinity~\citep{bian2026echo} uses a fixed number of learnable Memory Queries to absorb history evicted from the local window and updates them through attention and gating. Optimized jointly with the video DiT, these queries provide an evolving fixed-size summary of long histories.

The central trade-off is between compression efficiency and preservation of future-relevant information. Stronger compression reduces storage, attention cost, or memory bandwidth, but increases the risk of removing evidence whose utility appears only after long temporal gaps. Moreover, compressed states may differ substantially from the original history representations seen by the generator. Compression is therefore most effective when the generator is trained through the same representation pathway used at deployment, rather than applying compression only after training.

\subsubsection{Learning to Retrieve}
\label{sec:learning_to_retrieve}

Learning to retrieve concerns how information outside the active context is selected from an external memory store $\mathbf{E}_n$ and made accessible for future generation. Effective retrieval should recover evidence that is relevant to the current step yet unavailable from active memory, rather than merely selecting historically similar content. Existing methods follow two main routes: training with structured retrieval rules, or learning query and selection mechanisms from generation objectives or retrieval-specific supervision.

\vspace{-1.2em}
\paragraph{Structured Retrieval Rules.}
One line of work uses predefined spatial or geometric cues to select historical evidence and trains the generator with the resulting memory interface. Context-as-Memory~\citep{yu2025context} filters historical frames according to camera field-of-view overlap and conditions generation on the selected context. COVRAG~\citep{joo2026retrieve} uses a pretrained 3D prior to construct a target-view coverage map and iteratively selects historical frames by marginal coverage gain. MosaicMem~\citep{yu2026mosaicmem} lifts image patches into 3D and retrieves patches aligned with the target viewpoint. These methods estimate relevance through spatial correspondence at different granularities: field-of-view overlap provides an efficient view-level cue, whereas coverage estimation and 3D patch alignment provide finer visibility information. Their limitation is that geometric relevance alone may not capture state changes that occur during occlusion or interaction.

\vspace{-1.2em}
\paragraph{Learned Retrieval.} A second line of work learns query representations, selection functions, or conditional attention mechanisms, allowing memory access to adapt to the current generation context. LongLive-RAG~\citep{hu2026longlive} trains a compact retrieval encoder offline while keeping the generator frozen, using a window-level temporal-difference objective to reduce redundancy across adjacent windows. MemLearner~\citep{yu2026memlearner} inserts query tokens between historical and target tokens to learn target-relevant history extraction, while LayerRecall~\citep{ding2026layerrecall} trains a current-conditioned, layer-selective router through prediction matching with a privileged long-context reference. CaR~\citep{peng2026compression} retrieves information implicitly from compressed history through viewpoint-aware attention, whereas HyDRA~\citep{chen2026out} selects motion cues for occluded dynamic subjects according to spatiotemporal relevance. Ring Forcing~\citep{xue2026ring} constructs training sequences in which target evidence appears in distant history, making long-range retrieval necessary for accurate generation. These methods capture semantic, geometric, and dynamic relevance beyond predefined access rules, but their reliability still depends on sufficient training coverage of revisitation, occlusion, and reappearance.

Across both routes, retrieval quality depends on more than query similarity. Retrieved evidence should complement information already available in active memory, remain compatible with the current world state, and avoid reintroducing stale or corrupted history. Retrieval learning is therefore closely coupled with memory management and integration: selection determines what becomes accessible, while subsequent reliability assessment and integration determine whether that evidence should influence generation.

\subsubsection{Learning to Adapt}
\label{sec:learning_to_adapt}

The preceding pathways manage historical information through contextual or external memory states. Learning to adapt instead encodes prior rollout experience in an updatable parameter state that persists across autoregressive steps and modifies subsequent generation. Let $\theta$ denote the base generator parameters, $\Delta\theta_u$ the parameter-memory state after the $u$-th adaptation event, and $\mathbf{e}_u$ the update evidence derived from previously generated or observed history, optionally together with task feedback. At generation step $n$, the generator uses the most recent parameter-memory state $\Delta\theta_{u(n)}$, where $u(n)$ indexes the latest adaptation event completed before that step. Its update and subsequent use can be written as:
\begin{equation}
\begin{aligned}
\Delta\theta_{u+1}
&=
\mathcal{U}^{\mathrm{par}}_\phi
\left(
    \Delta\theta_u,
    \mathbf{e}_u
\right),
\\
\hat{\mathbf{y}}_n
&=
G_{\theta\oplus\Delta\theta_{u(n)}}
\left(
    \boldsymbol{\epsilon}_n;
    \mathbf{h}_n,
    \mathbf{a}_n
\right).
\end{aligned}
\end{equation}
Here, $\oplus$ denotes the model-specific composition of the base parameters with the adaptable parameter state. Unlike KV caches, memory tokens, or retrieved entries, $\Delta\theta_{u(n)}$ is not supplied to the generator as an addressable context input, but instead influences generation through its effective parameters. Adaptation constitutes parametric memory when its updates are derived from prior rollout experience and the resulting parameter state is reused across subsequent autoregressive steps. Transient optimization that affects only the current prediction is therefore not treated as persistent memory.

\vspace{-1.2em}
\paragraph{Recurrent Fast-Weight Adaptation.}
One line of work treats an updatable fast-weight state as parametric memory that evolves during autoregressive generation. Evidence from previously processed tokens or video chunks is written into the fast weights through online updates, and the resulting parameter state influences subsequent predictions. Test-Time Training Done Right~\citep{zhang2026test} introduces Large-Chunk TTT (LaCT), which performs fast-weight updates over substantially larger token chunks, improving hardware utilization and enabling higher-capacity nonlinear memory states for autoregressive video diffusion. Recurrent Autoregressive Diffusion (RAD)~\citep{chen2025rad} studies TTT as one of several recurrent memory instantiations, in which evolving fast weights carry long-range information across generation steps while full attention over overlapping local windows preserves fine-grained recent context. HippoCampus~\citep{peng2026hippocampus} similarly maintains a fixed-capacity fast-weight state that is updated online through a self-supervised latent prediction objective, while combining it with an explicit recent-context window and hierarchically compressed older history. Together, these methods illustrate how parameter adaptation can provide bounded persistent memory, with its effectiveness depending on the update objective, update granularity, parameter-state capacity, computational efficiency, and stability of repeated online updates.

\vspace{-1.2em}
\paragraph{Localized Parameter Adaptation.}
A complementary approach confines adaptation to a designated subset of parameters, reducing the cost of updating and storing parametric memory while leaving most of the generator unchanged. SlowFast-VGen~\citep{hong2025slowfast} updates a temporal LoRA module at inference using the input-output chunks accumulated along the generated trajectory, thereby encoding sequence-specific episodic information in a compact parameter state. ISPA~\citep{fu2026towards} instead derives instance-specific modulation for selected attention layers from the discrepancy between full-context and local-attention outputs observed during a warm-up phase. It computes the modulation through a closed-form least-squares solution, allowing the adapted local-attention layers to compensate for evicted historical KV context without iterative gradient-based optimization. These methods illustrate two complementary forms of localized parameter writing: historical evidence can be accumulated iteratively in lightweight parameter modules or absorbed analytically into selected model weights.

Parameter adaptation provides persistent memory without increasing the number of explicitly retained context tokens, but shifts memory costs and failure modes to the parameter-update process. Gradient-based updates to fast weights, including LoRA modules, introduce sequential computation and online optimization overhead. Closed-form parameter writing avoids iterative optimization, but its fidelity depends on whether the selected parameter subspace and approximation objective can preserve the historical information relevant to future generation. Across both forms, repeated updates may interfere with previously stored information, while updates derived from self-generated trajectories can consolidate prediction errors and propagate their effects to subsequent outputs. Reliable parametric memory must therefore balance the rapid incorporation of new evidence against the preservation of useful prior information, while controlling update cost, interference, forgetting, and error accumulation over long rollouts.

\section{Evaluation: How to Measure Memory?}
\label{sec:evaluation}

Video quality alone does not establish memory. A model can produce a coherent result without relying on its rollout history: the relevant subject may remain within the current context window, the prompt may restate a later event, or a learned prior may make the outcome plausible without preserving the episode state. Thus, although visual fidelity, text alignment, motion quality, and short-range consistency remain necessary quality checks, they do not by themselves provide evidence of memory.

We therefore adopt a stricter criterion. An evaluation is \emph{memory-revealing} only when the correct output depends on information acquired earlier in the rollout that is no longer available from the current condition alone. The protocol must therefore create an information gap, for example through disappearance, revisitation, hidden evolution, delayed interaction, or a controlled change in prior history. Mere duration is insufficient: a long rollout does not test memory if the current condition still reveals the information needed for the output.

\begin{figure*}[!t]
\centering
\vspace{-0.4em}
\includegraphics[width=0.98\linewidth]{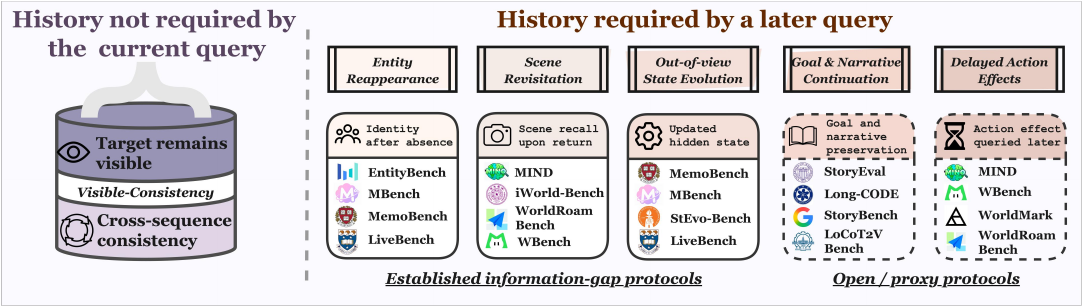}
\vspace{-0.6em}
\caption{Evaluation protocols organized by whether the current query requires rollout history. Visible consistency serves as a baseline because the target remains available. Established information-gap protocols include entity reappearance, scene revisitation, and out-of-view state evolution. Goal/narrative continuation and delayed action effects remain open or proxy directions; the listed benchmarks provide representative proxies or protocol components rather than fully standardized memory-revealing evaluations.}
\label{fig:memory_protocols}
\vspace{-0.6em}
\end{figure*}

This criterion also separates state retention from successful expression. A model may retain the relevant state yet fail to execute the requested action, return the camera to the target view, or render the queried content. Conversely, a correct output need not imply retention if the answer remains available in the prompt or can be inferred from a learned prior. A valid evaluation must therefore control both directions: execution controls verify that the queried behavior can be produced, whereas paired histories test whether the output actually depends on the relevant prior history.
\obsbox{
\textbf{Evaluation principle.}
A memory-revealing evaluation must make prior history necessary, verify the queried behavior, and rule out non-memory explanations.}
Following this principle, we first distinguish existing benchmarks by the strength of memory evidence they provide in Table~\ref{tab:benchmark_landscape}. We then organize evaluation protocols by the information they hide and the states they later query, as illustrated in Figure~\ref{fig:memory_protocols}. Finally, we match these protocols to target-specific measurements
(Table~\ref{tab:target_metric_map}) and examine the controls required for valid
attribution.

\subsection{Benchmark Landscape: Levels of Memory Evidence}
\label{sec:eval_landscape}

Because not every benchmark makes rollout history necessary, existing benchmarks provide different strengths of evidence for memory. We therefore group them by their \emph{evidential role}: whether and how the later query depends on information from prior rollout history, rather than by dataset size, task duration, or model family. Table~\ref{tab:benchmark_landscape} summarizes this landscape.

\vspace{-1.2em}
\paragraph{Memory-oriented benchmarks.}
These benchmarks create an explicit information gap between rollout history and a later query, for example by removing an entity from view, leaving and revisiting a scene, or hiding a state transition. MBench~\citep{zhang2026mbench}, MemoBench~\citep{MemoBench}, StEvo-Bench~\citep{ma2026out}, LiveBench~\citep{duan2026liveworld}, MIND~\citep{ye2026mind}, WBench~\citep{WBench}, and WorldRoamBench~\citep{xu2026worldroambench} instantiate such designs. Their evidence is direct only when the queried state is unavailable from the current prompt and observation; otherwise, local cues can explain a correct output without memory. This qualification can apply at the protocol or subset level: for example, only the gated camera-return subset of WBench~\citep{WBench} satisfies this criterion, rather than every reported metric.

\vspace{-1.2em}
\paragraph{Adjacent understanding-memory benchmarks.}
A second group evaluates delayed use of previously observed video in streaming or long-form understanding rather than generation. StreamMemBench~\citep{StreamMemBench}, EGOSTREAM~\citep{EGOSTREAM}, RIVER~\citep{RIVER}, EgoMemReason~\citep{EgoMemReason}, M3Eval~\citep{huang2026m}, S-EMBER~\citep{S-EMBER}, and LongSpace-Bench~\citep{LongSpace-Bench} provide useful protocol references for retention, retrieval, and past-dependent reasoning, but their outputs are answers, retrieval results, or auxiliary memory responses rather than generated video. We therefore discuss them as adjacent evidence without including them in Table~\ref{tab:benchmark_landscape}. Likewise, Video-MME~\citep{Video-MME}, LongVideoBench~\citep{LongVideoBench}, MLVU~\citep{MLVU}, EgoSchema~\citep{EgoSchema}, or LVBench~\citep{LVBench} primarily stress long-context understanding or evaluator capacity because the source video remains available at query time.

\providecommand{\toprule}{\hline}
\providecommand{\midrule}{\hline}
\providecommand{\bottomrule}{\hline}
\providecommand{\addlinespace}[1][0pt]{}
\providecommand{\rowcolor}[2][]{}
\providecommand{\evaltablehead}[1]{\makebox[\linewidth][c]{\textbf{#1}}}
\providecommand{\repolink}[1]{%
  \if\relax\detokenize{#1}\relax
  \else
    \href{#1}{\IfFileExists{figures/github-logo.pdf}%
      {\raisebox{-0.15ex}{\includegraphics[height=1.9ex]{figures/github-logo.pdf}}}%
      {\textsf{GH}}}%
  \fi}

\begin{table*}[!t]
  \centering
  \scriptsize
  \setlength{\tabcolsep}{4.2pt}
  \renewcommand{\arraystretch}{1.2}

\caption{Generation-side benchmarks for memory evaluation in video generation and world models.  
\textbf{Settings:} WM (world model), MS (multi-shot), Stream (streaming), AV (audio-video), Pred. (prediction), AR (autoregressive), I2V (image-to-video), and T2V (text-to-video).
%
\textbf{Memory targets:} indicate the memory dimensions probed by
the benchmark protocol: \textbf{E} (entity), \textbf{S} (spatial), \textbf{St} (state), and \textbf{A} (action).
\textbf{MD (Memory Dependence):} \protect\gapyes = prior rollout history is required by the evaluated protocol; \protect\gapsubset = this holds only for a designated subset, track, or protocol variant; \protect\gapno = prior rollout history is not directly required by the
protocol.}
  \vspace{-0.6em}
  \label{tab:benchmark_landscape}
  \begin{tabular}{
    l c|c cccc
    c
    l|c
  }
  \hlineB{2.5}
  \rowcolor{CadetBlue!20}
  \textbf{Benchmark} &
  \textbf{Link} &
  \textbf{Setting} &
  \textbf{E} &
  \textbf{S} &
  \textbf{St} &
  \textbf{A} &
  \textbf{MD} &
  \textbf{Primary feature} &
  \textbf{Temporal regime}
  \\

  \hlineB{1.5}

  \multicolumn{10}{c}{
    \cellcolor{gray!10}
    \textit{Memory-oriented generation and world-model benchmarks}
  }
  \\

  \hline

  MBench \citep{zhang2026mbench}
  & \repolink{https://github.com/study-overflow/MBench}
  & WM
  & \greenmark & \greenmark & \greenmark & \greenmark
  & \gapsubset
  & Triggered memory tests
  & 5 s--15 min sources
  \\
  LiveBench \citep{duan2026liveworld}
  & \repolink{https://github.com/ZichengDuan/LiveWorld}
  & WM
  & \greenmark & \greenmark & \greenmark & \greenmark
  & \gapyes
  & Same-/different-pose revisit
  & 100 scenes / 400 seq.; 260 frames
  \\
  WRBench \citep{WRBench}
  & \repolink{https://github.com/JinPLu/WRBench}
  & WM
  & \greenmark & \greenmark & \greenmark & \greenmark
  & \gapsubset
  & Hidden-and-returned state
  & 9,600 videos; 2,073 gated
  \\
  WBench \citep{WBench}
  & \repolink{https://github.com/meituan-longcat/WBench}
  & WM
  & \greenmark & \greenmark & \greenmark & \greenmark
  & \gapsubset
  & Gated round-trip return
  & 289 cases / 1,058 turns
  \\

  MemoBench \citep{MemoBench}
  & \repolink{https://github.com/MemoBench-Team/MemoBench}
  & I2V/WM
  & \greenmark & \greenmark & \greenmark & \redmark
  & \gapyes
  & Disappear--reappear
  & 4.1--5.4 s outputs
  \\
  R2M-Bench \citep{gu2026r2m}
  & \repolink{https://github.com/AMAP-ML/R2MBench}
  & WM
  & \greenmark & \greenmark & \greenmark & \redmark
  & \gapyes
  & Relative revisit (MG/NMR)
  & 300 instances; ~481 frames
  \\

  WorldRoamBench \citep{xu2026worldroambench}
  & \repolink{https://worldroam.amap.com/}
  & WM
  & \greenmark & \greenmark & \redmark & \greenmark
  & \gapsubset
  & Executed-path revisit
  & 10--60 s / 300--1,100 frames
  \\

  EntityBench \citep{he2026entitybench}
  & \repolink{https://github.com/Catherine-R-He/EntityBench}
  & MS
  & \greenmark & \redmark & \redmark & \redmark
  & \gapyes
  & Name-only reappearance
  & 140 episodes / 2,491 shots
  \\

  MIND \citep{ye2026mind}
  & \repolink{https://github.com/CSU-JPG/MIND}
  & WM
  & \redmark & \greenmark & \redmark & \greenmark
  & \gapsubset
  & Revisit / symmetric paths
  & 250 videos / 40+ environments
  \\

  iWorld-Bench \citep{iWorld-Bench}
  & \repolink{https://github.com/EmbodiedCity/iWorld-Bench}
  & WM
  & \redmark & \greenmark & \redmark & \greenmark
  & \gapsubset
  & Reciprocal camera return
  & 200 / 4,900 cases; 3.2--7.6 s
  \\

  UCM \citep{xu2026ucm}
  & \repolink{https://github.com/HumanAIGC/UCM}
  & WM
  & \redmark & \greenmark & \redmark & \greenmark
  & \gapsubset
  & Reverse-cycle / memory init.
  & 480 hist. $\rightarrow$ 321 pred.
  \\

  MAG-Bench \citep{zhu2025memorize}
  & \repolink{https://github.com/Xilluill/MAG}
  & AR
  & \redmark & \greenmark & \redmark & \redmark
  & \gapyes
  & Symmetric leave--return
  & 176 videos; $\sim$10 s protocol
  \\

  LoopNav \citep{lian2025loopnav}
  & \repolink{https://github.com/Kevin-lkw/LoopNav}
  & WM
  & \redmark & \greenmark & \redmark & \redmark
  & \gapyes
  & Loop-based spatial consistency
  & ABA/ABCA loops; 250 h / 20M frames
  \\

  StEvo-Bench \citep{ma2026out}
  & \repolink{https://github.com/jhanliufu-personal/STEVO-Bench}
  & I2V/T2V
  & \redmark & \redmark & \greenmark & \greenmark
  & \gapyes
  & Hidden process evolution
  & Local hidden-state protocol
  \\

  \hline

  \multicolumn{10}{c}{
    \cellcolor{gray!10}
    \textit{Generation and world-model sequence stress tests}
  }
  \\

  \hline

  VBench-Long \citep{VBench++}
  & \repolink{https://github.com/Vchitect/VBench/tree/master/vbench2_beta_long}
  & T2V
  & \greenmark & \greenmark & \greenmark & \redmark
  & \gapno
  & Long-range visible consistency
  & Long-video / arbitrary duration
  \\
  StoryBench \citep{StoryBench}
  & \repolink{https://github.com/google/storybench}
  & T2V
  & \greenmark & \greenmark & \greenmark & \redmark
  & \gapno
  & Story-level consistency
  & $\leq$30 s; mostly single-shot
  \\
  LoCoT2V-Bench \citep{LoCoT2V}
  & \repolink{https://github.com/XqZeppelinhead0702/LoCoT2V-Bench}
  & MS
  & \greenmark & \greenmark & \greenmark & \redmark
  & \gapno
  & Local/global consistency
  & 234 prompts; multi-shot
  \\
  InterVBench \citep{InterVBench}
  & \repolink{https://github.com/alibaba-damo-academy/BIFE}
  & T2V
  & \greenmark & \greenmark & \greenmark & \redmark
  & \gapno
  & Block-wise drift (VDE)
  & 1,000 videos; minute-long
  \\
  MSVBench \citep{MSVBench}
  & \repolink{https://github.com/HITsz-TMG/MSVBench}
  & MS
  & \greenmark & \greenmark & \greenmark & \redmark
  & \gapno
  & Hierarchical multi-shot eval.
  & 20 stories
  \\
  MSVE-Bench \citep{MSVE-Bench}
  & \repolink{https://reca.vmv.re/}
  & MS
  & \greenmark & \greenmark & \greenmark & \redmark
  & \gapno
  & Source-grounded extrapolation
  & 3--5 min
  \\
  DirectorBench \citep{DirectorBench}
  & \repolink{https://github.com/jiaminchen-1031/DirectorBench}
  & MS
  & \greenmark & \greenmark & \greenmark & \redmark
  & \gapno
  & Workflow checkpoints
  & 40 checkpoints; minute-long
  \\
  NarraStream-Bench \citep{NarraStream-Bench}
  & \repolink{https://github.com/Eddie0521/IAMFlow}
  & Stream
  & \greenmark & \redmark & \greenmark & \greenmark
  & \gapsubset
  & Prompt-stream continuity
  & 324 sequences; 60 s
  \\
  GroundBench \citep{lai2026groundshot}
  & \repolink{}
  & MS
  & \greenmark & \redmark & \greenmark & \redmark
  & \gapno
  & Reference-grounded recurrence
  & 54 scripts / 309 shots
  \\
  StoryEval \citep{wang2025your}
  & \repolink{https://github.com/ypwang61/StoryEval}
  & T2V
  & \greenmark & \redmark & \greenmark & \redmark
  & \gapno
  & Ordered event realization
  & 423 prompts; 2.0--10.6 s
  \\

  Long-CODE \citep{Long-CODE}
  & \repolink{https://github.com/ZhijiangTang/Long-CODE}
  & MS
  & \greenmark & \redmark & \greenmark & \redmark
  & \gapno
  & Structural corruption
  & 60--120 s / 12--24 shots
  \\

  SeqBench \citep{tang2025seqbench}
  & \repolink{https://github.com/TangZhengxu/SeqBench-Benchmarking-Sequential-Narrative-Generation-in-Text-to-Video-Models}
  & T2V
  & \greenmark & \redmark & \greenmark & \redmark
  & \gapno
  & Sequential narratives
  & $\leq$10 s
  \\
  NarrLV \citep{NarrLV}
  & \repolink{https://github.com/AMAP-ML/NarrLV}
  & T2V
  & \greenmark & \redmark & \greenmark & \redmark
  & \gapno
  & Narrative-atom evaluation
  & 5--12 s main; 60 s appendix
  \\

  LongAV-Compass \citep{LongAV-Compass}
  & \repolink{https://github.com/pkucs-Ltf/LongAV-Compass}
  & AV
  & \greenmark & \redmark & \greenmark & \redmark
  & \gapno
  & Multimodal continuity
  & 284 cases; minute-scale
  \\
  UnityShots \citep{UnityShots}
  & \repolink{https://github.com/JIA-Lab-research/UnityShots}
  & MS/AV
  & \greenmark & \redmark & \redmark & \redmark
  & \gapsubset
  & Audio-visual shot continuity
  & 200 sequences; 3--9 shots
  \\

  WorldOlympiad \citep{WorldOlympiad}
  & \repolink{https://github.com/alibaba-damo-academy/WorldOlympiad}
  & WM
  & \redmark & \greenmark & \greenmark & \greenmark
  & \gapno
  & Stitched world rollouts
  & $>$60 s / up to 6 chunks
  \\
  TECO \citep{TECO}
  & \repolink{https://github.com/wilson1yan/teco}
  & Pred.
  & \redmark & \greenmark & \greenmark & \redmark
  & \gapsubset
  & Horizon-wise fidelity
  & 300-frame prediction
  \\

  WorldMark \citep{WorldMark}
  & \repolink{https://github.com/AlayaLab/WorldMark}
  & WM
  & \redmark & \greenmark & \redmark & \greenmark
  & \gapsubset
  & Action and revisit tests
  & 500 cases; 20/40/60 s
  \\
  ChronoMagic-Bench \citep{ChronoMagicBench}
  & \repolink{https://github.com/PKU-YuanGroup/ChronoMagic-Bench}
  & T2V
  & \redmark & \redmark & \greenmark & \redmark
  & \gapno
  & Visible process progression
  & 1,649 prompts; single clip
  \\

  \hlineB{2.5}

  \end{tabular}
\end{table*}

\vspace{-1.2em}
\paragraph{Generation and world-model sequence stress tests.}
These benchmarks expose long-range degradation without making prior rollout history strictly necessary for the query. FVD~\citep{FVD} and suites such as VBench~\citep{huang2024vbench} measure fidelity, motion, alignment, and visible temporal consistency, while long-video and multi-shot evaluations extend these checks across chunks, shots, and narrative events \citep{VBench++,wang2025your,Long-CODE}. They are valuable for revealing drift and error accumulation, but success does not directly establish memory when the target remains visible or is specified by the current script. GroundBench~\citep{lai2026groundshot} and NarraStream-Bench~\citep{NarraStream-Bench}, for example, stress recurring identities while retaining scripts, descriptions, or references at recurrence. WorldModelBench~\citep{WorldModelBench} and WorldMark~\citep{WorldMark} similarly expose physical, control, action, and revisit failures without isolating memory under their default protocols, whereas UniVBench~\citep{UniVBench} serves as a broader evaluator reference. We therefore treat these suites as complementary sequence-level stress tests rather than direct memory benchmarks.

\vspace{-1.2em}
\paragraph{Component-level ablations.}
Benchmark evidence can be complemented by architecture-specific interventions. Disabling, deleting, replacing, or corrupting a memory bank, recurrent state, retrieval module, or KV cache can test whether a behavior depends on that component under matched conditions. Such ablations strengthen mechanistic attribution within a model, but their conclusions remain tied to the intervened architecture and therefore cannot substitute for a model-independent memory-revealing protocol.

Table~\ref{tab:benchmark_landscape} separates benchmark target coverage from memory dependence. \textbf{MD} indicates whether prior rollout history is required by the evaluated protocol, whereas \textbf{E}, \textbf{S}, \textbf{St}, and \textbf{A} denote non-exclusive entity, spatial, state, and action targets. These target labels describe what a benchmark evaluates; they do not imply that every target or metric independently requires memory. This distinction separates \emph{how directly} a benchmark reveals memory from \emph{what} behavior it measures, and motivates the protocol-centered analysis in Section~\ref{sec:eval_protocols}.

\providecommand{\FloatBarrier}{\clearpage}
\FloatBarrier

\subsection{Memory-Revealing Evaluation Setups}
\label{sec:eval_protocols}

Protocol design determines whether prior history is necessary for a later query. We distinguish a visible-consistency baseline from three established information-gap setups, entity reappearance, scene revisitation, and out-of-view state evolution, and two less standardized but important designs for goal/narrative continuation and delayed action effects. Across these setups, history-query separation provides a common stress variable. Figure~\ref{fig:memory_protocols} summarizes the corresponding queries and representative benchmarks.

\vspace{-1.2em}
\paragraph{Visible-Consistency Baseline.}
Visible and cross-sequence consistency provides a prerequisite baseline because the queried content remains available rather than being hidden from the current condition. VBench~\citep{huang2024vbench}, VBench++~\citep{VBench++}, and StoryEval~\citep{wang2025your} assess identity, background, motion, event structure, and long-range consistency across frames, chunks, or shots, with broader evaluator coverage provided by \citep{EvalCrafter,FETV,TC-Bench,FVMD,DirectMotion,ling2025vmbench,LoCoT2V,Long-CODE, InterVBench}. Method-specific studies further report scene-cut, motion-warp, drift, and duration-dependent behavior \citep{henschel2025streamingt2v,EgoLCD,li2026stable,yu2025videossm,yu2025malt}. These evaluations expose error accumulation and verify that the queried content can be generated, but they do not directly establish memory when the target remains visible or recoverable from the current script.

\vspace{-1.2em}
\paragraph{Entity Reappearance.}
Entity reappearance creates an explicit information gap by removing an entity from view and querying it again after occlusion, camera motion, or intervening shots. EntityBench~\citep{he2026entitybench}, MBench~\citep{zhang2026mbench}, MemoBench~\citep{MemoBench}, and LiveBench~\citep{duan2026liveworld} provide representative instances of this design, while recurrence-oriented suites and method-specific evaluations offer complementary stress cases \citep{lai2026groundshot,NarraStream-Bench,VideoMemory,SlotMem,zhang2025storymem,SlotMemory, Memento,ARGUS,CoTriSyGen}. Evaluation should first verify successful reappearance and only then assess identity and attributes. Two controls are therefore essential: the entity must actually leave the observable context, and the later prompt or reference must not restate the appearance to be recalled; otherwise, visible continuity or prompt leakage can explain success without memory.

\vspace{-1.2em}
\paragraph{Scene Revisitation.}
Scene revisitation asks whether a previously observed region remains recoverable after the camera leaves and later returns to a comparable view. MIND~\citep{ye2026mind}, iWorld-Bench~\citep{iWorld-Bench}, WBench~\citep{WBench}, WorldRoamBench~\citep{xu2026worldroambench}, and R2M-Bench~\citep{gu2026r2m} instantiate reciprocal, return, loop, and executed-path revisitation protocols, with related variants explored in \citep{zhu2025memorize,xu2026ucm}. Retrieval- and explicit-state methods provide additional geometric stress tests through pose-indexed frames, maps, and spatial memory \citep{yu2026memlearner,joo2026retrieve,yu2025context,li2025vmem,xiao2025worldmem, huang2025memory,chou2026captain,wang2025evoworld,Closing-the-Loop,gao2026memcam,li2026i3dm, WorldStereo,fu2026plenoptic,OmniRoam}. Because the query jointly depends on appearance, geometry, and camera closure, valid attribution requires the camera to leave, the target to become unavailable, and the rollout to return to a sufficiently comparable pose; otherwise, trajectory or control failure is confounded with memory failure.

\vspace{-1.2em}
\paragraph{Out-of-View State Evolution.}
A stronger information gap queries not the past observation itself but the state that should result from an unobserved transition. MemoBench~\citep{MemoBench}, StEvo-Bench~\citep{ma2026out}, and LiveBench~\citep{duan2026liveworld} provide representative protocols in which a visible process continues while hidden and its updated state is queried after reappearance, with related settings covered by \citep{zhang2026mbench,WRBench}. Such tests probe state, progress, event order, and delayed consequences rather than image reconstruction alone. Before an incorrect later state is attributed to memory, however, the model must demonstrate that it can generate the corresponding transition when it remains observable. Process-aware evaluations \citep{OSCBench,WorldReasonBench,Beyond-Last-Frame,Event-Conditioned-Fields} and physics datasets \citep{CATER,CLEVRER,IntPhys,IntPhys-2} provide complementary controls for this prerequisite, separating failure to model the transition from failure to retain or update its result.

\vspace{-1.2em}
\paragraph{Goal and Narrative Continuation.}
A semantic information gap arises when an early goal, narrative plan, role assignment, or unresolved commitment must constrain later generation after it has been omitted from the local condition. StoryEval~\citep{wang2025your}, Long-CODE~\citep{Long-CODE}, StoryBench~\citep{StoryBench}, and LoCoT2V-Bench~\citep{LoCoT2V} provide useful event-, structure-, and story-level proxies for goal satisfaction, ordering, omission, repetition, and contradiction. However, their complete scripts or structured prompts generally remain available, so these evaluations do not by themselves create a clean memory information gap. A stronger protocol would establish the relevant commitment early, remove it from later local conditions, and require a locally underspecified continuation whose correct resolution depends on rollout history.

\vspace{-1.2em}
\paragraph{Delayed Action Effects.}
Interactive generation motivates a setup in which an action changes the world state and its consequence is queried only after a delay, distraction, further interaction, or revisit. MIND~\citep{ye2026mind}, WBench~\citep{WBench}, WorldRoamBench~\citep{xu2026worldroambench}, and WorldMark~\citep{WorldMark} provide representative action, trajectory, and revisit components, but current protocols do not yet cleanly standardize a delayed query of an action-induced state; WBench~\citep{WBench}, for example, scores state-changing interactions within the current turn. Interactive, embodied, and counterfactual suites provide complementary execution and attribution controls \citep{KineBench,WorldScore,VP2,Omni-WorldBench,WorldArena,WorldGym,huang2026live, SANA-WM,WhatIfWorld,CRONOS,CounterScene}. These controls establish whether the action succeeds and whether alternative actions produce different outcomes, but memory is revealed only when the earlier action continues to constrain a later output after its immediate evidence is no longer available.

\vspace{-1.2em}
\paragraph{History-Query Separation.}
Across information-gap protocols, separation controls how far the relevant history is removed from the later query rather than defining a distinct memory target. TECO~\citep{TECO}, EntityBench~\citep{he2026entitybench}, WBench~\citep{WBench}, and InterVBench~\citep{InterVBench} illustrate horizon- or depth-stratified reporting. Depending on the protocol, suitable axes include recurrence gap for identity, path length or viewpoint displacement for spatial recall, hidden duration for evolving state, intervening events for semantic commitments, and interaction depth for delayed action effects. Such curves should not automatically be interpreted as memory decay: for example, the turn-wise curve in WBench~\citep{WBench} primarily measures rollout stability rather than controlled retention. Meaningful separation analysis therefore requires matched generation conditions and controlled resource budgets so that generic quality degradation or changing compute constraints are not mistaken for memory failure.

Together, these setups make memory observable by creating an information gap, verifying the prerequisite behavior, and querying how performance changes as the relevant history becomes more remote. This protocol view provides the basis for selecting target-specific metrics in the next subsection.

\subsection{Evaluation Targets and Metrics}
\label{sec:eval_targets_metrics}

A protocol makes prior history necessary; a target-specific metric then tests whether the delayed query is answered correctly. No score is memory-specific in isolation. For example, DINO \citep{caron2021emerging} similarity between adjacent frames measures visible consistency, whereas the same score after disappearance and reappearance can measure delayed identity preservation. The protocol therefore determines why a score is memory-relevant, while the queried state determines what should be measured.

\vspace{-1.2em}
\paragraph{Generation Quality Gates.}
Before interpreting target-specific scores, an evaluation should establish that the generated output is sufficiently valid for those scores to be meaningful. FVD~\citep{FVD}, VBench~\citep{huang2024vbench}, VBench++~\citep{VBench++}, and EvalCrafter~\citep{EvalCrafter} provide representative distributional and suite-based measures of visual quality, motion, alignment, and temporal consistency, with coverage from \citep{FETV,AIGCBench,STREAM}. Learned and human-aligned evaluators further provide prompt-, preference-, and fine-grained quality judgments \citep{T2VEval,LOVE,FingER,VideoScore,VideoScore2,VideoGen-Eval,Video-Bench}. These measures help identify general generation failures that would make a memory-specific score difficult to interpret, but they do not establish memory on their own.

\vspace{-1.2em}
\paragraph{Entity and Identity Metrics.}
For entity reappearance, evaluation should proceed from basic presence verification to identity and attribute consistency. Detection or segmentation first verifies that the queried entity actually reappears; DINO \citep{caron2021emerging}, CLIP \citep{radford2021learning}, ArcFace \citep{deng2019arcface}, or VLM-based judgments \citep{fu2024gptscore} can then compare its identity and attributes with the earlier observation. This ordering prevents similarity computed over selected frames from hiding omission, substitution, or duplication. T2V-CompBench~\citep{T2V-CompBench} and KeyFrame-Compass~\citep{KeyFrame-Compass} provide representative compositional and keyframe-level checks, with multi-reference, identity, and artifact diagnostics provided by \citep{MultiRef-Compass,ARGUS,DynamicEval,VC-Bench,Artifact-Bench, BrokenVideos}.

\vspace{-1.2em}
\paragraph{Spatial and Geometric Metrics.}
Scene revisitation requires complementary measurements of appearance and geometry. PSNR \citep{hore2010image}, SSIM \citep{wang2004image}, LPIPS \citep{zhang2018unreasonable}, and DreamSim \citep{fu2023dreamsim} quantify aligned visual reconstruction, whereas relative pose error, depth and reprojection error, registration accuracy, Chamfer distance, and point-cloud precision/recall measure camera closure and scene structure. Neither family is sufficient alone: perceptual agreement need not imply geometric consistency, while geometric alignment can miss texture, appearance, or entity changes. GeCo~\citep{GeCo}, PDI-Bench~\citep{PDI-Bench}, and WorldScore~\citep{WorldScore} provide representative diagnostics, with related spatial and 3D evaluation tools explored in \citep{SGC,L3DE,3DSPA,MVTrack4Gen}.

\vspace{-1.2em}
\paragraph{State and Process Metrics.}
For out-of-view state evolution, the target shifts from reconstructing an old image to identifying the latest valid process or world state. State classifiers, progress labels, event predicates, causal-order checks, and VQA/VLM rubrics can evaluate whether the hidden transition reached an appropriate outcome. Such state-level measurements are particularly important when multiple visual realizations are valid and pixel-wise comparison to a single reference would be overly restrictive. TC-Bench~\citep{TC-Bench} and OSCBench~\citep{OSCBench} provide representative state- and process-oriented evaluation designs, with additional physical, grounding, and mechanism-level diagnostics in \citep{VideoPhy-2,PhyGenBench,PhyCoBench,T2VPhysBench,PhyWorldBench,VAMP, PQSG,PhyGround,jain2026mechverse}.

\vspace{-1.2em}
\paragraph{Goal and Narrative Metrics.}
Goal and narrative continuation requires measurements beyond local frame quality. Goal completion and contradiction checks test whether an earlier commitment remains satisfied, while event coverage, ordering, causal consistency, and omission or repetition measure how the continuation realizes that commitment. StoryEval~\citep{wang2025your} and Long-CODE~\citep{Long-CODE} provide representative event- and structure-level measures, while StoryBench~\citep{StoryBench} and LoCoT2V-Bench~\citep{LoCoT2V} assess broader story consistency and alignment. These scores become evidence of memory only under an information-gap protocol in which the relevant goal or script is no longer available from the local condition; otherwise, they primarily measure visible sequence consistency.

\vspace{-1.2em}
\paragraph{Action-Effect Metrics.}
Delayed action effects likewise require staged evaluation. Action success, response accuracy, and trajectory error first establish that the intervention was executed correctly; state predicates or delayed-effect accuracy can then test whether the induced change persists. Action-conditioned suites such as MIND \citep{ye2026mind}, WBench \citep{WBench}, WorldRoamBench \citep{xu2026worldroambench}, and WorldMark \citep{WorldMark} provide execution and trajectory controls, while VACT \citep{VACT} and What-If World \citep{WhatIfWorld} offer complementary counterfactual or intervention-based causal checks. Physics-oriented continuation benchmarks such as WorldBench \citep{WorldBench} can further establish whether the underlying state transition is generatively plausible.
These components do not yet form a standardized delayed-action memory metric. A delayed-state score is therefore interpretable only when the original action succeeds and its specification is no longer sufficient to answer the later query.

\vspace{-1.2em}
\paragraph{History-Query Separation Curves.}
Once a target-specific score is defined, it should be examined as a function of the separation between the relevant history and the later query. Suitable axes include recurrence gap, path length or viewpoint displacement, hidden duration, intervening events, and interaction depth. Derived summaries such as degradation slope, first-failure time, retention half-life, or area under a retention curve may aid comparison, but should not replace the underlying curves, since gradual drift and abrupt failure can yield similar endpoint scores. Memory budget is a complementary resource axis rather than a measure of separation and should therefore be controlled or reported separately.

\vspace{-1.2em}
\paragraph{Evaluator Reliability and Calibration.}
Target-specific measurements are only as reliable as the evaluators that produce them. Automatic judges may miss brief failures, depend strongly on temporal sampling, or disagree with human assessments, making both temporal coverage and calibration important. UVE-Bench~\citep{UVE} and SLVMEval~\citep{SLVMEval} directly examine evaluator reliability, while EvalVerse~\citep{EvalVerse} and related human-calibrated or artifact-focused resources provide complementary evidence \citep{WorldJen,MSAVBench,Q-Bench-Video,MVQA-68K,GeneVA}. Tool-oriented frameworks such as AIGVE-Tool~\citep{AIGVE-Tool} and Q-Router~\citep{Q-Router} further illustrate tool-oriented or agentic composition of multiple evaluators.

Table~\ref{tab:target_metric_map} summarizes the resulting protocol-target-metric mapping. Target-specific scoring, however, does not by itself establish that a measured outcome was caused by memory use; the validity and attribution controls required for that conclusion are discussed next.

\begin{table*}[!t]
\centering
\scriptsize
\setlength{\tabcolsep}{5.2pt}
\renewcommand{\arraystretch}{1.2}
\caption{Target-specific evaluation for memory-related video generation.
Each protocol pairs an evaluation target with representative benchmarks and
the measurements needed to test it. Diagnostic measurements are synthesized at the survey level and are not necessarily source-defined metrics of every benchmark listed in the same row.}
\vspace{-0.6em}
\label{tab:target_metric_map}
\begin{tabular}{l|c|l}
\hlineB{2.5}
\rowcolor{CadetBlue!20}
\textbf{Evaluation target} &
\textbf{Representative benchmarks} &
\textbf{Recommended diagnostic} \\
\hlineB{1.5}
\multicolumn{3}{c}{\cellcolor{gray!10}\textit{Visible and cross-sequence consistency}} \\
Subject identity
& VBench-Long~\citep{VBench++}, LoCoT2V-Bench~\citep{LoCoT2V}
& DINO/CLIP similarity; ReID \\
\addlinespace[1.5pt]
Background and scene
& VBench-Long~\citep{VBench++}, InterVBench~\citep{InterVBench}
& Background consistency; LPIPS; drift \\
\addlinespace[1.5pt]
Motion and event structure
& StoryEval~\citep{wang2025your}, Long-CODE~\citep{Long-CODE}
& Flow/flicker; event accuracy; order score \\

\hline
\multicolumn{3}{c}{\cellcolor{gray!10}\textit{Entity leaves and returns}} \\
Presence and permanence
& MBench~\citep{zhang2026mbench}, MemoBench~\citep{MemoBench},
  LiveBench~\citep{duan2026liveworld}
& Detection rate; segmentation; reappearance success \\
\addlinespace[1.5pt]
Identity and attributes
& EntityBench~\citep{he2026entitybench}, GroundBench~\citep{lai2026groundshot}
& DINO/CLIP; ArcFace/ReID; attribute fidelity \\

\hline
\multicolumn{3}{c}{\cellcolor{gray!10}\textit{Camera leaves and returns}} \\
Appearance reconstruction
& MIND~\citep{ye2026mind}, MAG-Bench~\citep{zhu2025memorize}, UCM~\citep{xu2026ucm}
& PSNR; SSIM; LPIPS; perceptual similarity \\
\addlinespace[1.5pt]
Camera and pose closure
& iWorld-Bench~\citep{iWorld-Bench}, WBench~\citep{WBench},
  UCM~\citep{xu2026ucm}
& Pose closure; memory symmetry; trajectory error \\
\addlinespace[1.5pt]
Geometry and scene relations
& LiveBench~\citep{duan2026liveworld}, WorldRoamBench~\citep{xu2026worldroambench}
& Depth/reprojection; Chamfer distance; 3D P/R/F1 \\

\hline
\multicolumn{3}{c}{\cellcolor{gray!10}\textit{State changes out of view}} \\
State and progress
& MemoBench~\citep{MemoBench}, StEvo-Bench~\citep{ma2026out}
& State/progress accuracy; phase completion \\
\addlinespace[1.5pt]
Event and causal consistency
& MBench~\citep{zhang2026mbench}, LiveBench~\citep{duan2026liveworld}
& Event predicates; causal/order accuracy; VQA/VLM \\

\hline
\multicolumn{3}{c}{\cellcolor{gray!10}\textit{Semantic preservation}} \\
Goal and narrative commitments
& StoryBench~\citep{StoryBench}, LoCoT2V-Bench~\citep{LoCoT2V}
& Goal completion; contradiction; global alignment \\
\addlinespace[1.5pt]
Event and relational structure
& StoryEval~\citep{wang2025your}, Long-CODE~\citep{Long-CODE}
& Event coverage; ordering; omission/repetition \\

\hline
\multicolumn{3}{c}{\cellcolor{gray!10}\textit{Action effect queried later}} \\
Action execution
& MIND~\citep{ye2026mind}, WBench~\citep{WBench}, WorldMark~\citep{WorldMark}
& Action success; response accuracy; trajectory error \\
\addlinespace[1.5pt]
Delayed effect retention
& VACT~\citep{VACT}, What-If World~\citep{WhatIfWorld}
& Delayed-state correctness after successful intervention \\

\hline
\multicolumn{3}{c}{\cellcolor{gray!10}\textit{Cross-cutting reporting over separation}} \\
Retention over separation
& EntityBench~\citep{he2026entitybench}, InterVBench~\citep{InterVBench}
& Normalized degradation; first-failure time \\
\addlinespace[1.5pt]
Performance over interaction
& TECO~\citep{TECO}, WBench~\citep{WBench}
& Horizon curves; turn-wise success and degradation \\
\hlineB{2.5}
\end{tabular}
\end{table*}

\subsection{Validity Controls and Attribution}
\label{sec:eval_validity}

A target-specific score establishes whether the queried outcome is correct, but not why. A high score may arise from prompt leakage, visible cues, or a learned prior, whereas a low score may reflect failed task execution, general generation error, or evaluator failure rather than deficient memory. Valid attribution therefore requires controls that isolate dependence on prior history and rule out these alternative explanations.

\vspace{-1.2em}
\paragraph{Paired-History Controls.}
Paired histories provide the most direct behavioral test of whether an output depends on rollout history. The current observation and query are held fixed while the relevant earlier history is changed; the output should respond to changes in relevant history while remaining stable to matched changes that are irrelevant to the query. 
Existing counterfactual and paired-condition benchmarks provide partial ingredients for this design, although few generation benchmarks currently implement the strict paired-history intervention in which the current condition is held fixed while only relevant earlier history is changed \citep{WhatIfWorld, CounterScene, CRONOS}.
To preserve the intended information gap, prompts must not restate the hidden state, and paired trajectories should be matched in length, current visible content, and other non-causal cues. Otherwise, output differences may be explained without retrieving the relevant history.

\vspace{-1.2em}
\paragraph{Protocol Prerequisite Checks.}
History dependence alone does not distinguish memory failure from failure to execute the protocol. Each setup therefore requires prerequisite conditions to be verified before its memory score is interpreted: entity reappearance requires confirmed absence and return; scene revisitation requires successful departure and pose-compatible closure; out-of-view evolution requires a valid state transition; goal or narrative continuation requires an established earlier commitment that is absent from the later local condition; and delayed action effects require successful execution of the original intervention. Reporting these prerequisite success rates separates memory failure from rendering, control, navigation, dynamics, or narrative-realization failure.

\vspace{-1.2em}
\paragraph{Component-Level Interventions.}
Paired-history controls establish behavioral dependence on prior history, but do not identify which internal mechanism carries or exposes that information. Architecture-specific interventions can provide this complementary evidence by disabling memory banks, retrieval, recurrence, compression, or retained context \citep{VideoMemory,duan2026liveworld,king2026echo}. More targeted tests delete, replace, corrupt, or block specific memory content while matching capacity, computation, and other conditioning signals. A selective performance change under such controls supports component-level attribution and can reveal whether the relevant evidence was functionally accessible to generation. Because these interventions depend on a particular architecture, however, they provide mechanistic evidence within a system rather than a model-independent memory benchmark.

\vspace{-1.2em}
\paragraph{Evaluator Reliability Controls.}
Attribution is also limited by errors in the evaluator itself. Identity judges may overlook omission or duplication, geometric metrics may miss appearance changes, and sparsely sampled VLM judges may fail to observe brief temporal errors. Evaluations should therefore report temporal sampling and failure coverage, and calibrate automatic scores against human judgments or controlled corruptions that instantiate the targeted failure modes. Metric reliability is consequently part of evaluation validity rather than a post-processing concern.

Together, these controls answer complementary attribution questions: paired histories test whether prior history matters, prerequisite checks establish that the queried behavior is executable, component interventions localize dependence within a particular memory mechanism, and evaluator controls verify that the resulting failure is measured reliably. No single control is sufficient on its own; credible memory evaluation requires these sources of evidence to agree.

\section{Future: What Remains Open?}
\label{sec:future}

\begin{figure*}[!t]
\centering
\vspace{-0.4em}
\includegraphics[width=0.9\linewidth]{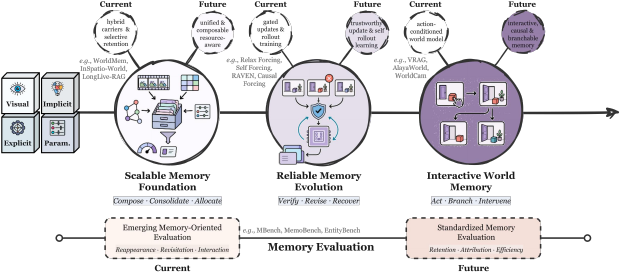}
\vspace{-0.6em}
\caption{Roadmap of current progress and future directions for memory-augmented autoregressive video generation. The upper pathway traces three transitions: from heterogeneous carriers to unified and resource-aware memory, from gated updates to trustworthy self-rollout learning, and from action-conditioned world models to interactive, causal, and branchable memory. The lower pathway highlights the parallel evolution from emerging memory-oriented benchmarks toward standardized evaluation of retention, attribution, and efficiency.}
\label{fig:figure_future}
\vspace{-0.6em}
\end{figure*}

Despite rapid progress, memory mechanisms for autoregressive video generation remain fragmented across representation, operation, learning, and evaluation. Existing methods often optimize individual carriers or local objectives, leaving unresolved how heterogeneous memory states can be coordinated, selectively maintained, reliably revised, and effectively utilized throughout long-horizon self-rollouts \citep{huang2026self, yang2025longlive, zhu2026causal}. The remaining challenge is therefore not simply to retain more history, but to maintain useful, reliable, accessible, and consequential state under bounded computation and recursive generation. These challenges are coupled: composable and resource-aware memory provides the substrate for persistent state, trustworthy updating preserves its validity, self-rollout learning adapts the model to induced states, and interactive settings further require action-induced changes to remain consequential over time. Figure~\ref{fig:figure_future} summarizes this progression and the cross-cutting role of memory evaluation. This section examines the resulting research directions toward scalable, reliable, and diagnostically evaluable memory-augmented video generation.

\subsection{Unified and Composable Memory}

\paragraph{Current Progress and Limitations.}
Long-horizon video generation requires memory to preserve heterogeneous information across temporal and representational scales. Visual and implicit states retain fine-grained appearance and motion, explicit representations support spatial, entity, and causal structure, and parametric memory provides an additional channel for persistent adaptation across rollout steps \citep{dalal2025one, hong2025slowfast}. Recent systems increasingly combine visual anchors, compressed features, geometric maps, semantic states, or adaptive parameters within a shared generation pipeline \citep{duan2026liveworld, long20262, inspatio2026world, xiao2025worldmem, yu2026mosaicmem}. Such combinations exploit complementary carrier properties, since perceptual fidelity, structured representation, efficient access, editability, and adaptation impose different requirements.

Current combinations, however, are often coordinated only loosely. Carriers may use incompatible entity identifiers, temporal references, coordinate systems, access mechanisms, or update rules, while cross-carrier verification and revision remain limited. Visual memories preserve detailed evidence at substantial storage and access cost \citep{henschel2025streamingt2v, joo2026retrieve, li2025vmem, yu2025context}; structured memories improve compactness and interpretability but may omit rendering detail \citep{garcin2026beyond, long20262, yu2026mosaicmem}; and parametric memory introduces interference and limited reversibility \citep{dalal2025one, hong2025slowfast}. The open problem is therefore not to collapse these carriers into a single representation, but to coordinate historical evidence and evolving world-state estimates so that they remain consistent and jointly usable by the generator.

\vspace{-1.2em}
\paragraph{Future Directions.}
Future systems should develop shared interfaces for composable memory without requiring a unified representation. Common entity identifiers, timestamps, coordinate frames, provenance, confidence, and validity intervals could align visual evidence, geometric structure, semantic relations, and parametric state across carriers. Such interfaces should support selective access, cross-carrier consistency checking, revision, replacement, invalidation, and reset without corrupting unrelated memory content.

Hierarchical composition may further coordinate information across temporal and representational scales. Detailed evidence can remain in visual or implicit memory when future reconstruction may require it, while stable history is selectively consolidated into compact entity, scene, event, or causal state. Rollout-specific regularities may also be absorbed into modular parametric memory when updates can be isolated and reversed. The goal is a coordinated memory system that preserves the representation best suited to each type of information while exposing only what later generation requires; how to achieve this under bounded storage and access motivates resource-aware memory design.

\subsection{Resource-Aware Memory}

\paragraph{Current Progress and Limitations.}
Long-horizon video generation cannot preserve its entire history at full fidelity or expose it to every generation step. Practical memory systems must therefore decide what to write, retain, compress, retrieve, and discard under bounded storage and computation. Existing methods control memory growth through rolling windows, persistent anchors \citep{yang2025longlive, yang2026anchorforcing, zhao2026relax}, cache eviction \citep{chen2026past, xu2026sparse}, hierarchical compression \citep{bian2026echo, li2026attend, yesiltepe2026videomla, zhang2026frame}, and selective retrieval \citep{hu2026longlive, joo2026retrieve, peng2026compression, yu2026memlearner}. Adaptive strategies further estimate importance from attention, novelty, reconstruction error, temporal coverage, or geometric overlap to preserve or reactivate selected historical evidence.

A central limitation is that future utility is only partially observable when these decisions are made. Policies based on current salience or similarity may discard rare but consequential events, dormant causal commitments, or temporarily absent entities \citep{chen2026past, joo2026retrieve, xu2026sparse}, while redundant evidence continues to consume resources. Moreover, memory cost extends beyond storage to active access, retrieval latency, and integration into the generator. Because writing, compression, retention, and retrieval are often designed separately, errors introduced at one stage constrain what later stages can recover. Capacity alone therefore does not determine memory quality; retained information must also remain reliable, accessible, and useful.

\vspace{-1.2em}
\paragraph{Future Directions.}
Future systems should treat memory allocation as a prospective utility problem rather than a retrospective salience decision. Retention and access policies should estimate the downstream value of historical evidence from recurrence likelihood, state reliability, temporal coverage, and the consequence of forgetting. This would favor low-frequency but high-impact information, such as an early identity cue or an intervention whose effect must persist after its local evidence disappears. Counterfactual removal or replacement of memory entries may further help identify which evidence is consequential for later generation.

Resource allocation should also be coordinated across memory operations and resource types. High-fidelity storage can be reserved for evidence that is difficult to reconstruct and likely to become relevant again, while stable long-term information is selectively consolidated into compact states. Rather than increasing memory uniformly with rollout length, systems could reallocate a bounded active-access budget around revisitations, interactions, or state changes. Storage, access cost, retrieval latency, and computation should then be reported together with memory performance. The goal is not merely to retain more history, but to preserve useful information at the fidelity and access cost required for future generation, which in turn depends on whether the retained state can be trusted.

\subsection{Trustworthy Memory Updating}

\paragraph{Current Progress and Limitations.}
Persistent memory must preserve commitments that remain valid while adapting when the generated world legitimately changes. Existing systems address this stability--plasticity problem through recurrent aggregation \citep{yu2025malt, yu2025videossm}, cache replacement \citep{mao2026packforcing, zhao2026relax}, geometric fusion \citep{wang2025evoworld, WorldStereo}, and parameter adaptation \citep{dalal2025one, hong2025slowfast}. Gated mechanisms regulate how strongly new evidence modifies previous states, while structured memories revise entity, spatial, or event states more explicitly. Selective forgetting further removes information judged stale or invalid, allowing memory to follow legitimate world changes rather than simply preserve the past.

A central limitation is that new evidence is not necessarily trustworthy. Self-generated observations may be consolidated without adequately distinguishing genuine state changes from accumulated generation error, allowing local artifacts to affect subsequent rollout \citep{huang2026self, lu2026raven}. Conversely, overly persistent anchors may suppress legitimate transitions and restore outdated states. Such failures are hard to diagnose when memory does not expose provenance, confidence, or validity, and are especially difficult to reverse in parametric memory. Trustworthy updating must therefore determine not only \emph{what} has changed, but whether the evidence justifies revising an existing commitment.

\vspace{-1.2em}
\paragraph{Future Directions.}
Future systems should distinguish stored evidence from the persistent state currently treated as valid. Evidence could carry provenance, source type, timestamps, confidence, and validity intervals, allowing conflicting observations to be assessed by source reliability, temporal order, and cross-carrier agreement before a revision is committed. Uncertain evidence could then remain available for later verification without immediately overwriting a more reliable state.

Revision should also be selective, localized, and reversible. Rather than overwriting an entire memory state, future systems should update only the entities, relations, regions, or commitments invalidated by new evidence while preserving unrelated state. Versioning, rollback, and conflict-aware consolidation could limit persistent contamination; modular adaptation or isolated parameter patches may provide analogous control for parametric memory.

Trustworthy updating should also be learned under closed-loop rollout. Training can expose models to contradictory evidence, corrupted histories, delayed corrections, stale states, and invalid transitions, requiring them to identify and revise the responsible memory content. Counterfactual rollouts can test whether a correction produces the intended downstream change without disturbing unrelated commitments. The goal is confidence-aware maintenance of persistent state under the noisy histories produced by AR generation.

\subsection{Self-Rollout Learning}

\paragraph{Current Progress and Limitations.}
Memory is ultimately deployed on histories generated by the model itself, so prediction errors can enter persistent state and affect subsequent generation. Teacher-forced training remains widely used because it provides stable optimization and clean histories, while existing objectives supervise future outputs, intermediate memory states, or state transitions \citep{bardhan2026persistent, huang2026live, jiang2026panoworld}. History augmentation broadens the training distribution through masked, perturbed, or corrupted histories \citep{chen2025skyreels, ji2026videoar, sun2025ar}, whereas self-rollout constructs memory from the model's own recursive predictions \citep{cui2026self, huang2026self, liu2026opsd, lu2026raven}.

A central gap remains between the memory states seen during training and those induced during deployment. Teacher forcing emphasizes idealized histories, while hand-designed augmentation may not reproduce structured errors such as identity drift, geometric distortion, stale state, or incorrect updates \citep{chen2024diffusion, guo2025end, ruhe2024rolling, valevski2025diffusion}. Self-rollout provides more realistic states but remains costly and difficult to stabilize over long horizons. Delayed failures are also hard to assign to a particular write, compression, update, retrieval, or integration decision, so immediate prediction losses may improve local quality without teaching the memory behaviors required for distant generation.

\vspace{-1.2em}
\paragraph{Future Directions.}
Future self-rollout learning should improve coverage of deployment-induced memory states, especially states that are both likely and consequential. Curricula over rollout depth or memory difficulty \citep{huang2026live, lu2026raven, po2026bagger}, mixtures of clean, augmented, and self-generated histories, and replay of failed trajectories may expose recurring errors and recovery states without repeatedly sampling full rollouts. The objective is not simply longer trajectories, but better coverage of the states that govern long-horizon failure.

Training must also provide useful supervision on states actually visited by the model. Distant-state prediction, trajectory-level consistency, verifier-guided feedback, and teacher or reference supervision can test whether retained information remains correct, accessible, and effective after long gaps \citep{bardhan2026persistent, xu2026next, yin2026vigor, zhu2026video}. Interventions on memory entries or operations may further localize whether a later failure arose from missing retention, incorrect state, failed retrieval, stale updating, or weak integration.

Scalable training should reduce the cost of long-range credit assignment without removing the closed-loop conditions that create memory errors. Segment-level rollout, reuse of generated trajectories, selective backpropagation, and training around critical memory events can reduce serial cost \citep{cui2026self, huang2026self, liu2026opsd}. The resulting objectives should optimize not only predictions along generated trajectories, but also the persistent states and transitions recursively created by those trajectories.

\subsection{Interactive and Causal Memory}

\paragraph{Current Progress and Limitations.}
Interactive generation requires memory to preserve not only prior observations but also state changes induced by action. Recent world models increasingly support action-conditioned rollout, embodied navigation, object manipulation, and multi-turn interaction \citep{alonso2024diffusion, duan2026liveworld, garcin2026beyond, valevski2025diffusion}. Visual histories and recurrent states maintain short-range continuity, while geometric maps, entity states, and event summaries provide more persistent representations \citep{long20262, pondaven2026actionparty, inspatio2026world}. Some systems further model state transitions or action effects \citep{duan2026liveworld, team2026advancing, SANA-WM}, extending memory beyond appearance preservation toward persistent physical and semantic state.

Action conditioning alone, however, does not establish causal memory. A model may execute an intervention correctly yet fail to preserve its consequences once the action and its visual evidence leave local context. A door may revert to an earlier state, a displaced object may return after revisitation, or an acknowledged event may cease to affect later rendering \citep{xu2026worldroambench, WorldMark, ye2026mind, WBench}. Such failures can arise when intervention-induced state changes are not persistently represented, correctly updated, or later reactivated, although control, dynamics, and rendering errors remain alternative explanations. The open challenge is therefore to maintain valid action consequences as persistent state that continues to constrain subsequent generation.

\vspace{-1.2em}
\paragraph{Future Directions.}
Future systems should preserve intervention-linked transitions as first-class memory, regardless of carrier. Such state can record agents, affected entities, preconditions, outcomes, and scope \citep{chen2025vrag, team2026advancing, xiao2025worldmem}, while remaining consistent with visual, spatial, dynamic, and semantic evidence. Code-as-memory could couple executable rules to persistent runtime state, keeping action effects revisable as generated evidence arrives \citep{chen2026code, huang2026programmable, fang2026codeplansdiffusionrenders}. Consequence-aware access should reactivate interventions whose effects remain relevant after the originating action becomes distant.

Interactive memory should further support controlled revision and branching. Corrections, resets, and rollbacks should modify the affected persistent state without corrupting unrelated history. Counterfactual generation additionally requires shared history and branch-specific updates to remain separated so that alternative futures can evolve without cross-branch contamination. Each branch should maintain persistent state that conditions subsequent visual generation, rather than serving only as trajectory bookkeeping.

Progress also requires learning and evaluation protocols that connect actions to delayed outcomes. Act-wait-query trajectories, counterfactual action pairs, and interventions followed by scene revisitation can test whether an induced state change remains consequential after local evidence disappears \citep{CRONOS, WhatIfWorld, CounterScene, VACT}. Training should encourage successful action effects to persist, invalid transitions to be rejected or revised, and delayed consequences to remain accessible when later generation depends on them. The broader goal is persistent interactive state whose action-induced changes remain accessible, revisable, and consequential over long-horizon rollout.

\subsection{Comparable Memory Evaluation Protocols}

\paragraph{Current Progress and Limitations.}
Memory evaluation increasingly moves beyond general video quality toward protocols that make prior rollout history necessary. Existing benchmarks test entity reappearance \citep{MemoBench, duan2026liveworld, he2026entitybench, zhang2026mbench}, scene revisitation \citep{iWorld-Bench, xu2026ucm, xu2026worldroambench, WBench}, and out-of-view state evolution \citep{MemoBench, ma2026out, zhang2026mbench}. Interactive suites further provide action and multi-turn stress tests \citep{xu2026worldroambench, WBench}, while long-horizon quality benchmarks expose complementary drift and error accumulation. Together, these developments provide increasingly direct evidence for identity, spatial, and state persistence, although controlled delayed-action and goal-dependent memory protocols remain less mature.

Evaluation nevertheless remains incomplete and difficult to compare across systems. Coverage is strongest for identity recurrence and spatial revisitation, whereas correction, contradiction resolution, stale-state invalidation, selective forgetting, rollback, and delayed intervention effects remain underexplored. Moreover, endpoint failures do not reveal where the memory pipeline failed: relevant information may have been discarded, incorrectly updated, unavailable to retrieval, or retrieved but ineffective during generation. Method-specific interventions can localize some of these failures within an architecture \citep{duan2026liveworld, king2026echo, VideoMemory}, but offer limited cross-model comparability. Differences in history--query separation, resource budgets, prerequisite success, temporal sampling, and evaluator reliability further complicate comparison.

\vspace{-1.2em}
\paragraph{Future Directions.}
Future evaluation should move from isolated endpoint scores toward diagnostic memory capability profiles. Compatible protocols should specify the relevant history, delayed query, history-query separation, prerequisite conditions, and controls for local cues or prompt leakage. Beyond measuring whether an output is correct, evaluation should distinguish whether relevant information was retained, whether the persistent state remained correct, whether that information was accessible when needed, and whether accessible memory actually influenced generation. This stage-wise diagnosis would separate capacity from state correctness, retrieval from utilization, and architectural storage from effective memory behavior.

Evaluation should also cover how memory changes, not only how long it lasts. Protocols for contradiction, delayed correction, selective forgetting, stale-state invalidation, reset, and rollback should test whether a system preserves commitments that remain valid while revising those invalidated by new evidence. Interactive settings further require controlled act-wait-query protocols in which action execution is verified before the persistence of its delayed consequence is scored. Such tests would expose trustworthy updating and causal persistence that conventional recurrence or revisitation benchmarks do not capture.

Comparable reporting should jointly characterize memory performance, history-query separation, and resource use. Recurrence gap, path length, hidden duration, and interaction depth should be varied independently from storage capacity, active-access budget, retrieval latency, and computation \citep{he2026entitybench, TECO, InterVBench}. Paired histories, prerequisite success rates, and calibrated evaluators should accompany these measurements so that changes in performance can be attributed to memory rather than execution failure, alternative cues, or evaluator error. The resulting profile should reveal what information remains valid, accessible, and effective, over what horizon, and at what resource cost.

\section{Conclusion}
\label{sec:conclusion}

This survey treats memory in autoregressive (AR) video generation as a persistent dependence on history after the relevant evidence leaves the active context. Organizing prior work by forms, functions, operations, learning, and evaluation reveals a sequence of distinct bottlenecks. A system may discard a crucial observation, maintain an incorrect or stale state, leave a correct state inaccessible, or retrieve relevant content without using it. These failures can produce similar visual symptoms but require different remedies. They also explain why longer context, larger stores, and aggregate video-quality scores do not, on their own, establish durable memory. The central question is whether the right historical information remains available and consequential for later generation.

We argue that the immediate priority is to make state maintenance and memory use verifiable. Historical evidence should be kept distinct from revisable estimates of the current world; updates should be justified by observations or actions, localized to the affected state, and robust to errors in self-generated histories. Evaluation should create a genuine history-query gap, verify that the queried action or revisit occurred, and test whether changing relevant past content changes later output. Read-time interventions can further determine whether the retrieved content itself contributes to generation. Without these controls, improvements in long-range consistency remain difficult to attribute to memory. The longer-term objective is a generator whose remembered state remains correct, accessible, and consequential through extended rollout, interaction, and revision.

\clearpage
\bibliography{main}

\end{document}